\documentclass[pdflatex,sn-nature]{sn-jnl}

\usepackage{graphicx}%
\usepackage{multirow}%
\usepackage{amsmath,amssymb,amsfonts}%
\usepackage{amsthm}%
\usepackage[title]{appendix}%
\usepackage{xcolor}%
\usepackage{url}
\usepackage{textcomp}%
\usepackage{manyfoot}%
\usepackage{booktabs}%
\usepackage{float}
\usepackage{algorithm}%
\usepackage{algorithmicx}%
\usepackage{algpseudocode}%
\usepackage{listings}%
\usepackage{gensymb}
\usepackage{threeparttable}
\usepackage{makecell}
\theoremstyle{thmstyleone}%
\newtheorem{theorem}{Theorem}%  meant for continuous numbers
\theoremstyle{thmstyletwo}%

\theoremstyle{thmstylethree}%

\begin{document}

\title{
    Resolving sources of uncertainty in AI weather forecasting
}


\author[1]{\fnm{Wenbo} \sur{Hu}}
\equalcont{These authors contributed equally to this work.}
\author[1]{\fnm{Xinlei} \sur{Xiong}}
\equalcont{These authors contributed equally to this work.}
\author[1]{\fnm{Shuxun} \sur{Zhou}}
\author[2]{\fnm{Kaifeng} \sur{Bi}}
\author[2]{\fnm{Lingxi} \sur{Xie}}
\author[3]{\fnm{Jun} \sur{Zhu}}
\author[1]{\fnm{Richang} \sur{Hong}}
\author*[2]{\fnm{Qi} \sur{Tian}}\email{wywqtian@gmail.com}

\affil[1]{Hefei University of Technology, Hefei, China}
\affil[2]{Huawei Inc., Shenzhen, China}
\affil[3]{Tsinghua University, Beijing, China}

\abstract{
Weather forecast uncertainty arises from imperfect analyses and forecast models, but ensemble spread alone does not reveal how distinct sources relate to downstream targets. We introduce Pangu-Bayes, a probabilistic forecasting hierarchy that treats atmospheric-state and learned-model uncertainty as distinct stochastic variables, crossing flow-dependent perturbations of the evolving state with Bayesian parameter samples. This construction yields model-defined source-resolved variance components and matched pathway evaluation. Across 90 held-out 2023 tropical cyclones, Pangu-Bayes reduces track, pressure and wind errors by 54.2\%, 17.2\% and 24.9\%, respectively, while improving rapid-intensification detection. Among 88 cyclones supporting pathway comparison, atmospheric-state variability is more consistently associated with improved track prediction, whereas learned-model variability is more often associated with improved intensity prediction. During Mawar and Khanun, state perturbations sample alternative steering-flow evolutions associated with recurvature, whereas parameter sampling broadens intensity evolution. Pangu-Bayes thus connects model-defined uncertainty resolution to target-dependent value and dynamical interpretation while retaining competitive global probabilistic skill.
}

\keywords{
    Probabilistic weather forecasting,
    uncertainty decomposition,
    Bayesian deep learning,
    ensemble forecasting,
    interpretable AI,
    tropical cyclones
}


\maketitle

Weather forecasting is inherently probabilistic because atmospheric evolution is chaotic, the analysed state is imperfect and forecast models are incomplete~\cite{lorenz1963deterministic,lorenz1996essence,palmer2002economic}.
Medium-range forecasts must therefore represent both expected evolution and plausible alternatives~\cite{gneiting2007strictly,leutbecher2008ensemble}.
Ensemble numerical weather prediction has long distinguished initial-condition from model uncertainty, representing them through perturbations of the analysed state and forecast-model dynamics, such as bred or singular vectors and stochastic parameterizations~\cite{toth1993ensemble,buizza1999stochastic,leutbecher2008ensemble,palmer2009stochastic,palmer2019ecmwf}.
Although these sources can be isolated experimentally, their effects propagate together in operational ensemble trajectories and are further coupled by nonlinear dynamics and ensemble data assimilation~\cite{leutbecher2008ensemble,palmer2009stochastic,palmer2019ecmwf}.
For probabilistic AI, the more specific question is whether state and model uncertainty can be defined as separately addressable stochastic variables of the learned predictive distribution itself.

AI weather models now achieve competitive medium-range skill at substantially lower inference cost than traditional numerical systems~\cite{pathak2022fourcastnet,bi2023accurate,lam2023learning,chen2025operational,chen2023fuxi,bodnar2025foundation}, and probabilistic systems extend this progress to predictive distributions.
GenCast generates stochastic trajectories through conditional diffusion~\cite{price2024probabilistic}; WeatherNext Cyclones (WN-C), built on functional generative networks, combines independently trained models with learned within-model functional perturbations while conditioning on an unperturbed initial analysis~\cite{alet2026operational}; FuXi-ENS applies learned flow-dependent perturbations at initialization and repeatedly during autoregressive forecasting~\cite{fuxiens2025}; and SFNO-BVMC crosses bred-vector initial-condition perturbations with independently trained model checkpoints~\cite{mahesh2025huge1,mahesh2025huge2}.
These systems show that structured variability, evolving perturbations and crossed initial-condition--model ensembles are already possible.
The remaining question is whether atmospheric-state and learned-model uncertainty can be defined as stochastic components of the predictive model itself, rather than identified only through ensemble organization or grouping.

\begin{figure}
    \centering
    \includegraphics[width=0.8\linewidth]{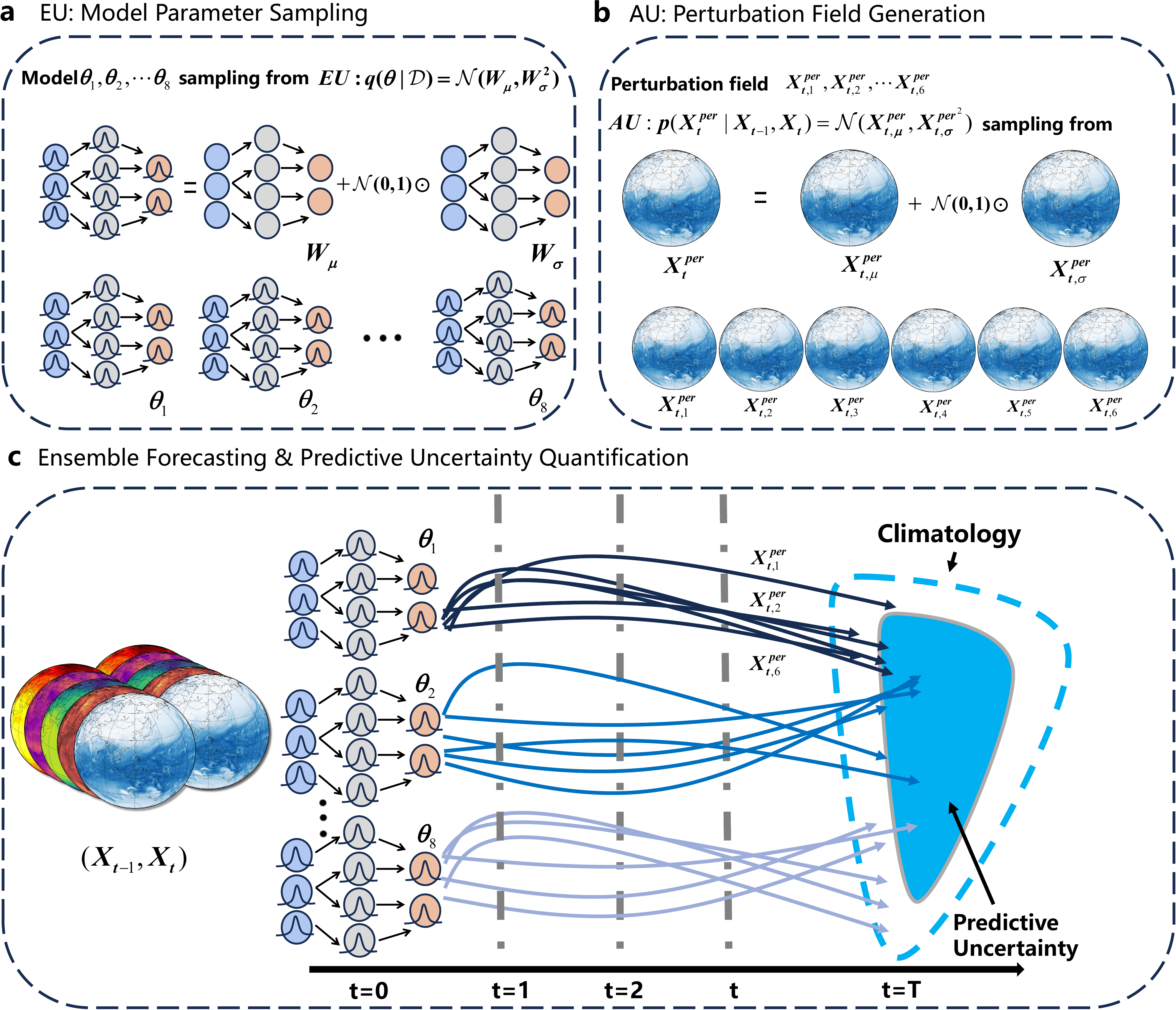}
    \caption{
        \textbf{Source-resolved uncertainty pathways in Pangu-Bayes.}
        \textbf{a,} Learned-model uncertainty is represented by sampling forecast-model parameters \(\theta_i\) from a variational posterior \(q(\theta\mid\mathcal{D})\).
        \textbf{b,} Atmospheric-state uncertainty is represented by a perturbation model that, conditioned on two consecutive states \((x_{t-1},x_t)\), samples flow-dependent additive perturbations \(\eta_{t,j}\), yielding perturbed states \(\tilde{x}_{t,j}=x_t+\eta_{t,j}\).
        \textbf{c,} Autoregressive ensemble forecasting crosses the two pathways: \(8\) model realizations and \(6\) atmospheric-state perturbations per realization form a \(48\)-member ensemble up to lead time \(T\).
        The probabilistic hierarchy defines atmospheric-state and learned-model uncertainty as distinct stochastic axes whose crossing generates the \(48\)-member ensemble, enabling model-defined variance decomposition and matched pathway evaluation.
    }
    \label{fig:model_overview}
\end{figure}

This leaves a specific diagnostic gap: \emph{a skilful predictive distribution does not reveal how differently sourced uncertainty is structured or which sources carry useful information for particular targets}.
If the sources are stochastic variables of the predictive hierarchy, their conditional contributions become intrinsic to the distribution, allowing one pathway to be controlled or marginalized while the other is evaluated.
The key question is therefore not only how predictive variance is partitioned, but whether model-defined uncertainty pathways acquire different forecast value after nonlinear propagation and whether those differences have recognizable dynamical expressions.

Here we introduce Pangu-Bayes, which defines atmospheric-state and learned-model uncertainty within a shared probabilistic forecasting hierarchy.
Flow-dependent perturbations of the analysed and autoregressively evolving state form one conditional stochastic variable, while parameter samples from a variational posterior over a shared forecast model form the other (Fig.~\ref{fig:model_overview}).
Their crossing generates the ensemble; the corresponding state and model variance components are therefore properties of the specified predictive distribution, with finite members providing Monte Carlo estimates.
This construction enables matched pathway comparisons of source-specific forecast value and dynamical manifestation.

We evaluate Pangu-Bayes on 90 held-out tropical cyclones from 2023.
Under a common-case protocol, it yields lower track errors than WN-C and lower probabilistic MSLP and MSW errors on the metrics evaluated here, with mean relative reductions of 54.2\%, 17.2\% and 24.9\% for track, pressure and wind, respectively, while improving rapid-intensification detection.
Among 88 cyclones supporting pathway comparison, atmospheric-state variability is more consistently associated with improved track prediction, whereas learned-model variability is more often associated with improved intensity prediction.
During Mawar and Khanun recurvature, state perturbations sample alternative steering-flow evolutions linked to the western Pacific subtropical high, whereas parameter sampling broadens intensity evolution.
Pangu-Bayes also retains competitive global probabilistic skill for selected variables and lead times on the 2022 WeatherBench2 benchmark under matched operational-analysis initialization.
Together, these results connect model-defined source resolution to target-dependent forecast value and dynamical interpretation while retaining credible global probabilistic skill.

\section{Results}\label{sec:results}
We first establish the source-resolved construction of Pangu-Bayes, then evaluate tropical-cyclone forecasts and source-specific pathway behaviour, and finally assess global probabilistic skill and calibration.

\subsection{Source-resolved uncertainty construction in Pangu-Bayes}
\label{sec:source_resolved_construction}

Pangu-Bayes defines atmospheric-state and learned-model uncertainty as distinct stochastic variables of its predictive hierarchy: a conditional flow-dependent perturbation distribution represents the evolving atmospheric state, and a variational posterior over the shared forecast backbone represents learned-model uncertainty.
At inference, eight parameter realizations are crossed with six state perturbations to form a \(48\)-member Monte Carlo ensemble (Fig.~\ref{fig:model_overview}).
Because the two axes are defined before sampling, one can be varied while the other is controlled or averaged over, enabling matched pathway evaluation.

For \(\theta\sim q(\theta\mid\mathcal D)\) and
\(\eta_t\sim p(\eta_t\mid x_{t-1},x_t)\), the law of total variance gives
\begin{equation}
\mathrm{Var}
\left(
x_{t+1}\mid x_{t-1},x_t,\mathcal D
\right)
=
\underbrace{
\mathbb{E}_{\theta}
\left[
\mathrm{Var}_{\eta_t}
\left(
x_{t+1}\mid \theta
\right)
\right]
}_{V_{\mathrm{state}}}
+
\underbrace{
\mathrm{Var}_{\theta}
\left[
\mathbb{E}_{\eta_t}
\left(
x_{t+1}\mid \theta
\right)
\right]
}_{V_{\mathrm{model}}}.
\label{eq:tc_uncertainty_decomposition}
\end{equation}
Here, \(V_{\mathrm{state}}\) is the expected conditional variance induced by atmospheric-state perturbations within a model realization, whereas \(V_{\mathrm{model}}\) is the variance of model-realization conditional means after marginalizing over state perturbations.
These are population-level quantities of the specified predictive model; the decomposition is exact under this hierarchy but is not a unique causal partition of physical atmospheric uncertainty.
The \(8\times6\) ensemble provides Monte Carlo estimates, as detailed in Methods~\ref{sec:source_resolved_formulation}; local propagation is analysed in Methods~\ref{sec:local_uncertainty_propagation} and Supplementary Section~\ref{app:local_uncertainty_propagation}.

\subsection{Tropical cyclone track, intensity and rapid intensification}
\label{sec:tc_forecast}

\begin{figure}[htbp]
    \centering
    \includegraphics[width=0.97\linewidth]{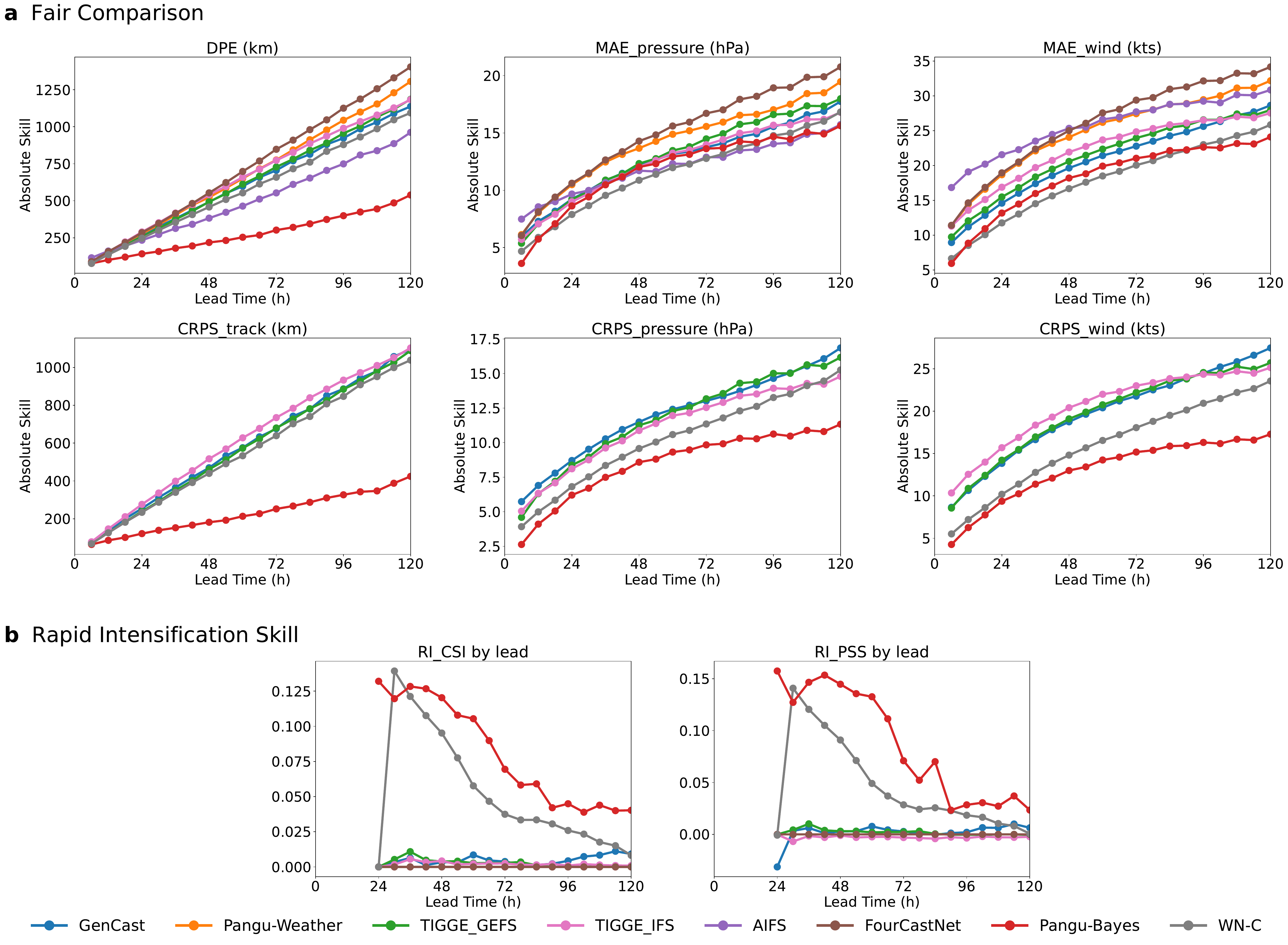}
    \caption{
        \textbf{Tropical cyclone forecast performance across track, intensity and rapid intensification.}
        \textbf{a,} Fair-protocol comparison of deterministic and probabilistic track and intensity forecast errors from 6 to 120~h for Pangu-Bayes and leading operational and AI baselines, including WeatherNext Cyclones.
        Missing cyclone forecasts are replaced by persistence from the initial cyclone state so that all systems are evaluated on the same verifying cases.
        Lower values indicate better performance for DPE, MAE and CRPS.
        \textbf{b,} Rapid-intensification forecast skill from 24 to 120~h, measured using the critical success index (CSI) and Peirce skill score (PSS) against IBTrACS observations.
        Higher CSI and PSS indicate better RI detection.
        Axis ranges are truncated for readability without excluding values from the reported statistics.
    }
    \label{fig:cyclone_evaluation}
\end{figure}

We evaluate the held-out 2023 cohort of 90 tropical cyclones against IBTrACS best-track records~\cite{knapp2010international}, comparing Pangu-Bayes with TIGGE IFS and GEFS, Pangu-Weather, FourCastNet, AIFS, GenCast and WeatherNext Cyclones (WN-C)~\cite{bougeault2010thorpex,pathak2022fourcastnet,bi2023accurate,lang2024aifs,price2024probabilistic,alet2026operational}.
The main comparison uses a fair protocol in which missing forecasts are replaced by persistence from the initial cyclone state, ensuring common verifying cases while penalizing failures to maintain a detectable cyclone.
Figure~\ref{fig:cyclone_evaluation} summarizes track and intensity forecasts from 6 to 120~h and rapid intensification (RI) from 24 to 120~h; evaluation details are given in Methods~\ref{sec:tc_benchmark_protocol} and the Supplementary Information.

\noindent\textbf{Track prediction against leading cyclone baselines.}
Track errors increase with lead time for all systems, but Pangu-Bayes maintains substantially lower errors in the fair-protocol comparison (Fig.~\ref{fig:cyclone_evaluation}a).
On the track metrics evaluated here, Pangu-Bayes yields lower DPE and track CRPS than WN-C when aggregated over the fair-protocol verification cases.
Across 6--120~h, the mean relative reductions, averaged over lead times and competing models, are 51.6\% for DPE and 56.9\% for track CRPS, corresponding to an average reduction of 54.2\% across the two track metrics.
The complementary raw and detection-coverage analyses show that this advantage is not obtained simply by excluding difficult cases in which a system fails to maintain a detectable cyclone (Supplementary Figs.~\ref{fig:raw_comparison} and \ref{cnt}).

\noindent\textbf{Cyclone intensity prediction.}
Intensity is more difficult to recover directly from global fields because MSLP and MSW depend on compact inner-core structures that are only partially represented at global-model resolution.
Pangu-Bayes therefore uses the probabilistic intensity module described in Methods, operating on cyclone-level predictors derived from the forecast fields.
On the probabilistic metrics evaluated here, Pangu-Bayes achieves lower MSLP and MSW CRPS than WN-C.
Across the full benchmark, the corresponding mean relative reductions are 9.4\% for MSLP MAE and 25.0\% for MSLP CRPS, giving an average pressure reduction of 17.2\%; for MSW, the reductions are 18.9\% for MAE and 30.9\% for CRPS, giving an average wind reduction of 24.9\% (Fig.~\ref{fig:cyclone_evaluation}a).
Because raw-field intensity estimates are substantially less skilful, these gains should be attributed to the dedicated intensity capability rather than direct raw-field intensity prediction.

\noindent\textbf{Rapid-intensification detection.}
RI remains challenging for the global forecasting systems considered here (Fig.~\ref{fig:cyclone_evaluation}b).
Using the same probabilistic intensity module, Pangu-Bayes increases CSI from 0.0087 for the competing-model mean to 0.0804 and PSS from 0.0066 to 0.0866 over 24--120~h.
We do not interpret these results as uniform superiority over WN-C for RI; rather, they show that the intensity module substantially improves the representation of abrupt intensity changes.
Together, the track, intensity and RI results establish the forecast-performance basis for testing whether the two uncertainty pathways carry systematically different information about cyclone evolution.

\subsection{Source-resolved uncertainty in tropical cyclone forecasts}
\label{sec:tc_source_decomposition}
Having defined the two uncertainty sources within the predictive model, we next ask whether their distinct probabilistic roles translate into systematically different forecast value for tropical-cyclone targets after nonlinear propagation.
Pathway-level comparison is available for 88 of the 90 cyclones; two do not yield valid pathway-level results.
For the remaining 88 cyclones, both pathways use the same ERA5 initial states, cyclone matching and IBTrACS verification, excluding initialization or verification differences as explanations for pathway-level contrasts.
We examine the cyclone-level prevalence of source--target contrasts, their lead-time structure and relation to component spread, and their dynamical expression in representative recurvature cases.

\noindent\textbf{Cohort-level source--target contrasts.}
Among the 88 cyclones with valid pathway-level comparisons, 69 cyclones (78.41\%; 95\% Wilson confidence interval, 68.72--85.72\%) contain at least one initialization for which the atmospheric-state pathway yields lower track error than the learned-model pathway.
Similarly, 58 cyclones (65.91\%; 95\% Wilson confidence interval, 55.53--74.96\%) contain at least one initialization for which the learned-model pathway yields lower intensity error than the atmospheric-state pathway.
At the cyclone level, both directional patterns occur somewhere among the available initializations for 49 of 88 cyclones (55.68\%; 95\% Wilson confidence interval, 45.28--65.61\%).
A stricter same-initialization criterion is satisfied in 37 cyclones (42.05\%; 95\% Wilson confidence interval, 32.28--52.48\%), for which at least one initialization simultaneously exhibits both the atmospheric-state--track and learned-model--intensity directional contrasts.

These results show that the atmospheric-state--track association is more prevalent than the learned-model--intensity association and that both can vary with forecast initialization.
They quantify source-specific forecast value rather than variance magnitude: both pathways can affect both targets, but atmospheric-state variability is more consistently informative for track, whereas learned-model variability is more often informative for intensity, with the latter relationship weaker.
Cohort summaries are given in Supplementary Table~\ref{tab:source_target_summary}, with additional track visualizations in Supplementary Figs.~\ref{All_Typhoons_4col6row_Part1}--\ref{All_Typhoons_4col6row_Part4}.

\begin{figure}[htbp]
    \centering
    \includegraphics[width=1.0\linewidth]{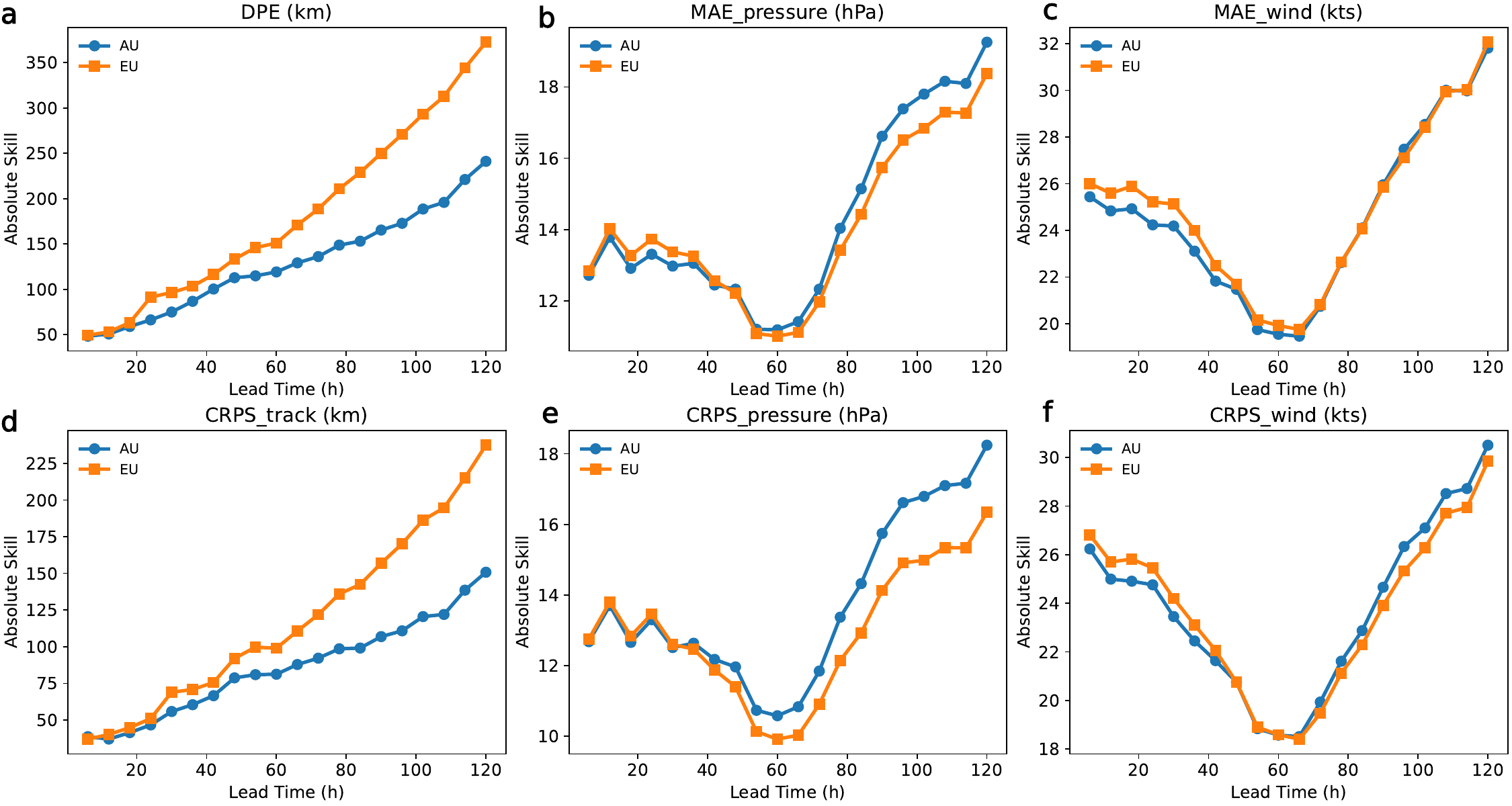}
    \caption{
        \textbf{Lead-time evolution of source-specific tropical cyclone track and intensity forecast errors.}
        \textbf{a--f,} Mean direct positional error (DPE), pressure mean absolute error (MAE), wind-speed MAE, track continuous ranked probability score (CRPS), pressure CRPS and wind-speed CRPS, respectively, from 6 to 120~h.
        Results are shown for the descriptive subset of cyclone initializations in which the atmospheric-state--track and learned-model--intensity directional criteria are satisfied simultaneously at the same initialization.
        AU and EU in the figure denote the atmospheric-state and learned-model component ensembles, respectively.
        Lower values indicate better forecast performance.
        The atmospheric-state pathway shows a clearer and more persistent advantage for track prediction, whereas the learned-model advantage for intensity is smaller and most evident in pressure-based probabilistic forecasts.
        Because subset membership is defined using the same directional contrasts, these curves characterize lead-time structure rather than unbiased cohort-wide effect sizes.
    }
    \label{fig:qualified_mean_lead_curves}
\end{figure}

\noindent\textbf{Lead-time evolution and source-specific forecast value.}
We next examine the descriptive subset satisfying both directional criteria at the same initialization.
Because subset membership is defined using the same contrasts being visualized, the resulting effect sizes characterize lead-time structure rather than unbiased cohort-wide effects.
Within this subset, the atmospheric-state pathway reduces DPE by 28.64\% and track CRPS by 25.73\% relative to the learned-model pathway, corresponding to a mean reduction of 27.19\% across the two track metrics (Fig.~\ref{fig:qualified_mean_lead_curves}a,d).
The learned-model pathway reduces pressure MAE by 2.88\% and pressure CRPS by 9.23\% (Fig.~\ref{fig:qualified_mean_lead_curves}b,e).
For wind, MAE is 2.66\% higher under the learned-model pathway, whereas CRPS is lower by only 0.55\% (Fig.~\ref{fig:qualified_mean_lead_curves}c,f).
Averaged over the four intensity metrics, the learned-model pathway yields a modest 2.50\% relative reduction.
Thus, the atmospheric-state advantage for track is clearer and more persistent across lead times, whereas the learned-model advantage for intensity is weaker and concentrated mainly in probabilistic pressure prediction.

\begin{figure}[htbp]
    \centering
    \includegraphics[width=1.0\linewidth]{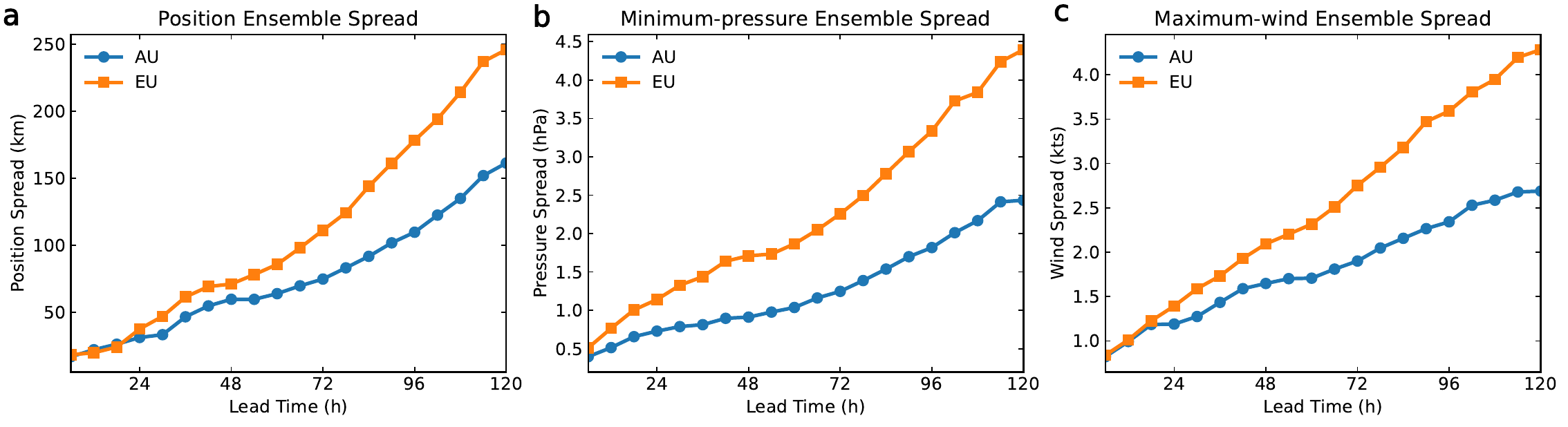}
    \caption{
        \textbf{Lead-time evolution of atmospheric-state and learned-model component-ensemble dispersion.}
        \textbf{a,} Positional spread, calculated as
        $\sqrt{\mathrm{Var}(x)+\mathrm{Var}(y)}$
        after converting ensemble-member locations to local Cartesian coordinates, where \(x\) and \(y\) denote east--west and north--south displacements in kilometres.
        \textbf{b,} Ensemble standard deviation of minimum sea-level pressure.
        \textbf{c,} Ensemble standard deviation of maximum sustained wind.
        Spread is calculated separately for the atmospheric-state and learned-model component ensembles for each cyclone-initialization sample in the descriptive joint-contrast subset and then averaged at each lead time from 6 to 120~h.
        AU and EU in the figure denote the atmospheric-state and learned-model pathways, respectively.
        The learned-model component generally produces broader distributions, demonstrating that component spread magnitude and source-specific forecast value are not equivalent.
    }
    \label{fig:au_eu_tc_spread}
\end{figure}

Component spread does not explain these forecast-value differences.
Position, pressure and wind spread generally increase with lead time, and the learned-model component is broader over most lead times (Fig.~\ref{fig:au_eu_tc_spread}).
Yet lower atmospheric-state track errors coexist with narrower positional spread, while the modest learned-model pressure advantage coexists with broader pressure and wind distributions.
Greater source-specific dispersion therefore need not imply greater source-specific forecast value; value depends on how variability is centred and dynamically organized relative to the verifying evolution.

\noindent\textbf{Dynamical interpretation during cyclone recurvature.}
The cohort-level statistics establish recurrent source--target organization but do not explain how the pathways are expressed in individual atmospheric evolutions.
We therefore examine Typhoons Mawar and Khanun as illustrative western North Pacific recurvature cases; they provide dynamical interpretation rather than independent population-level evidence.

\begin{figure}[htbp]
    \centering
    \includegraphics[width=1.0\linewidth]{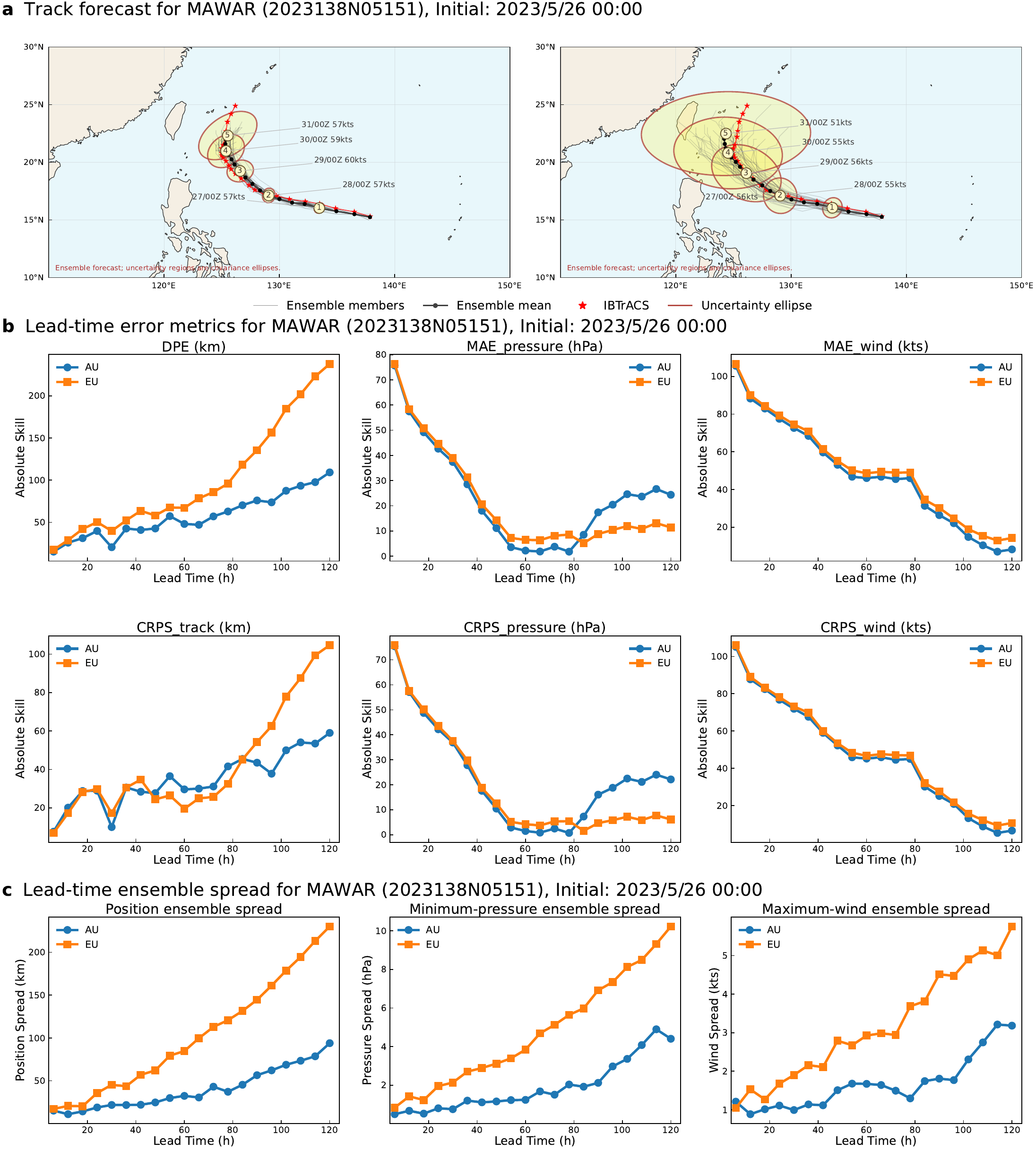}
    \caption{
        \textbf{Source-specific track and intensity evolution for Typhoon Mawar.}
        Forecasts are initialized at 00:00 UTC on 26 May 2023.
        \textbf{a,} Atmospheric-state (AU) and learned-model (EU) track ensembles compared with the IBTrACS best-track trajectory.
        \textbf{b,} Lead-time-dependent deterministic and probabilistic forecast errors for track, minimum sea-level pressure (MSLP) and maximum sustained wind (MSW).
        \textbf{c,} Lead-time evolution of component-ensemble dispersion for cyclone position, MSLP and MSW.
        Position spread is calculated as
        $\sqrt{\mathrm{Var}(x)+\mathrm{Var}(y)}$
        after converting ensemble-member locations to local Cartesian coordinates.
        AU and EU denote the atmospheric-state and learned-model pathways, respectively.
    }
    \label{fig:mawar_track_intensity_distribution}
\end{figure}

For Mawar, initialized at 00:00 UTC on 26 May 2023, atmospheric-state members follow the observed trajectory more closely and indicate the impending eastward recurvature earlier than learned-model members (Fig.~\ref{fig:mawar_track_intensity_distribution}a).
Differences between the pathways are more pronounced for track than intensity, whereas the learned-model pathway spans a broader range of intensity evolution and shows a clearer advantage for probabilistic MSLP at selected lead times (Fig.~\ref{fig:mawar_track_intensity_distribution}b,c).

Khanun provides an independent illustration of track sensitivity to the atmospheric-state pathway.
For the earlier initialization, atmospheric-state members indicate the forthcoming north-easterly turn, whereas learned-model members favour a more north-westerly trajectory (Supplementary Fig.~\ref{730_track_intensity_distribution}).
For the later initialization, atmospheric-state members again show a stronger northward component and follow the observed evolution more closely, whereas learned-model members remain concentrated along a more north-easterly trajectory (Supplementary Fig.~\ref{803_track_intensity_distribution}).

At 120~h, atmospheric-state members also show substantial variation in the 5,880-gpm boundary of the western Pacific subtropical high (WPSH), including differences in its western extension and northern-edge position (Supplementary Fig.~\ref{WPSH}).
These variations are consistent with distinct large-scale steering environments sampled by atmospheric-state perturbations, providing a meteorologically interpretable link between uncertainty in the evolving flow and cyclone-track diversity.
They should be interpreted as dynamical evidence consistent with the source--track association rather than causal attribution of cyclone motion to a single circulation feature.
Additional pathway-specific track visualizations and U10 EOF diagnostics for Mawar and Khanun are shown in Supplementary Fig.~\ref{fig:typhoon_kn}.

Across Mawar and Khanun, atmospheric-state perturbations sample alternative surrounding-flow and recurvature evolutions, whereas learned-model members span a broader range of plausible MSLP--MSW evolution.
Both pathways still under-represent parts of the strongest early intensification, and the learned-model intensity advantage remains weaker and more metric dependent than the atmospheric-state track advantage.
Together with the cohort and lead-time results, these cases connect model-defined source separation to target-dependent forecast value and dynamical interpretation.

\subsection{Global probabilistic forecast skill and calibration}
\label{sec:global_probabilistic_forecast_skill}

Finally, we assess whether the source-resolved behaviour identified in tropical cyclone forecasts is embedded in a credible global probabilistic forecasting system.
Global forecasts are evaluated on the 2022 WeatherBench2 test year and compared with IFS-ENS, GenCast and FGN under matched operational-analysis initialization, with ERA5 verification~\cite{rasp2024weatherbench,price2024probabilistic,alet2026operational}.
We assess ensemble-mean RMSE, CRPS, ensemble spread and spread--skill ratio (SSR); definitions are given in Methods~\ref{sec:global_dynamical_evaluation}.
Figure~\ref{fig:global_evaluation} shows T850, U850, T2M and U10, with additional variables in Supplementary Fig.~\ref{z500_v850}.

\begin{figure}[htbp]
    \centering
    \includegraphics[width=1.0\linewidth]{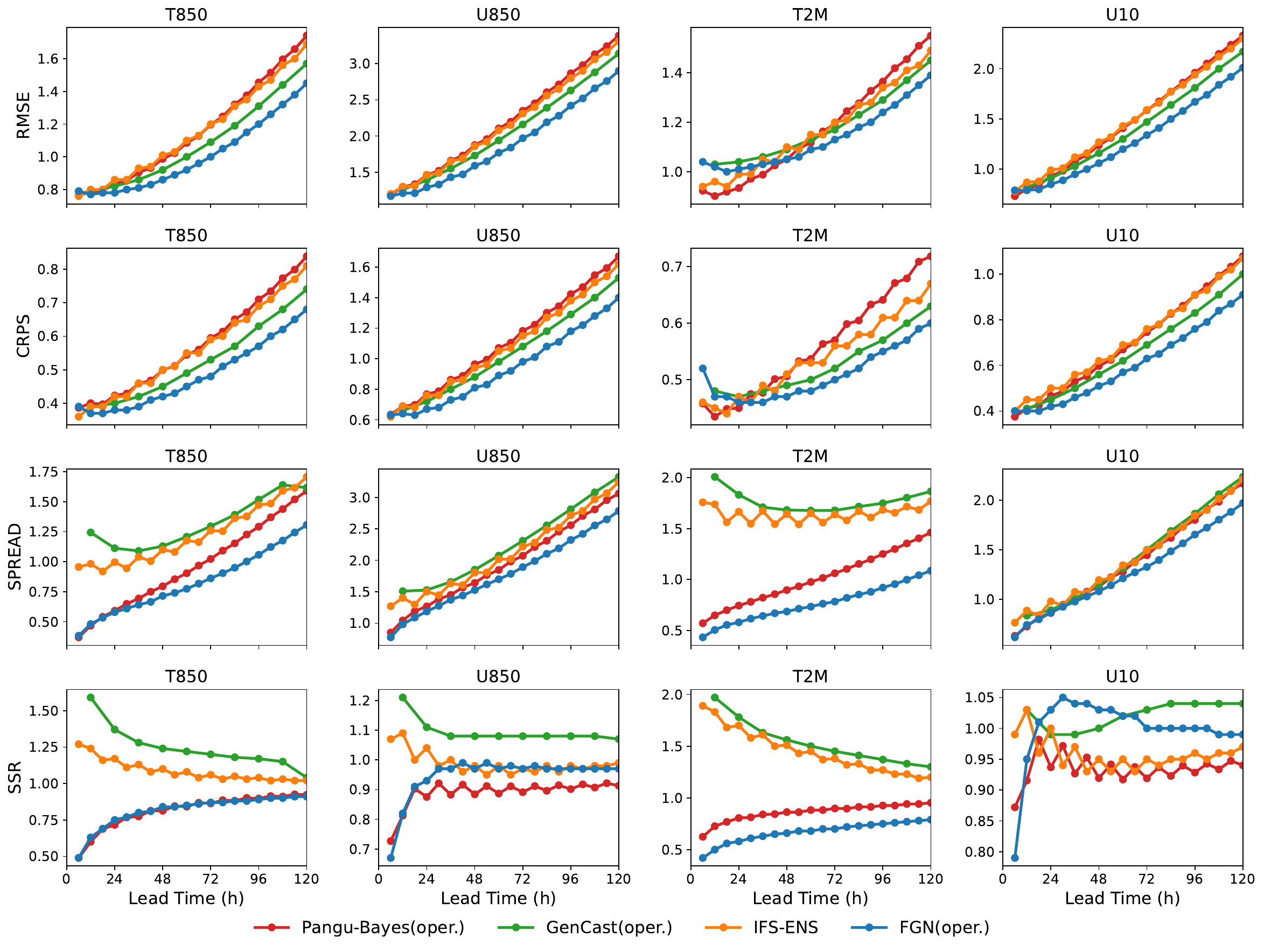}
    \caption{
        \textbf{Global probabilistic forecast skill and calibration.}
        Globally averaged and latitude-weighted RMSE, CRPS, ensemble spread and spread--skill ratio (SSR) for Pangu-Bayes, IFS-ENS, GenCast and FGN over the 2022 WeatherBench2 test year.
        Pangu-Bayes, GenCast and FGN are evaluated under matched operational-analysis initialization, with ERA5 used for verification.
        Columns show 850-hPa temperature (T850), 850-hPa zonal wind (U850), 2-m temperature (T2M) and 10-m zonal wind (U10); rows show RMSE, CRPS, ensemble spread and SSR, respectively.
        Lower RMSE and CRPS indicate better forecast skill, whereas SSR values closer to one indicate better agreement between ensemble dispersion and ensemble-mean forecast error.
        Spread is reported on the standard-deviation scale and therefore has the same physical units as the corresponding forecast variable.
    }
    \label{fig:global_evaluation}
\end{figure}

\noindent\textbf{Global ensemble skill.}
Pangu-Bayes achieves competitive global probabilistic skill for selected variables and lead-time ranges rather than uniformly outperforming the strongest baselines (Fig.~\ref{fig:global_evaluation}).
Its clearest advantage occurs for T2M at short-to-medium lead times, where RMSE and CRPS are lower than those of the comparison systems shown.
Performance is more mixed for upper-air and wind variables: Pangu-Bayes remains competitive with IFS-ENS in several regimes, whereas GenCast and particularly FGN generally retain lower errors at medium-to-long lead times.
The same pattern extends to Z500, V850, V10 and MSL (Supplementary Fig.~\ref{z500_v850}), with FGN providing the strongest overall performance across many longer-range metrics.

\noindent\textbf{Calibration and dispersion.}
Pangu-Bayes generally produces narrower predictive distributions than IFS-ENS and GenCast, but sharpness alone does not imply calibration.
For T2M, SSR remains comparatively close to one over short-to-medium lead times, consistent with the regime of strongest RMSE and CRPS performance.
U850 and U10 show lower SSR at several lead times, indicating residual underdispersion.
Calibration is therefore credible but variable- and lead-time-dependent rather than uniformly optimal.

Taken together, WeatherBench2 shows that Pangu-Bayes retains useful global probabilistic skill and credible calibration while introducing explicit source-resolved uncertainty structure.
The cyclone source--target relationships are therefore embedded in a global medium-range forecasting system rather than arising from a cyclone-specific diagnostic model.

\section{Discussion}\label{sec:discussion}

Probabilistic forecasting is commonly judged by accuracy and
calibration~\cite{murphy1993good,gneiting2007strictly}, but these characterize
the predictive distribution as a whole.
Pangu-Bayes instead defines atmospheric-state and learned-model uncertainty as
two source-specific stochastic variables of the predictive model itself.
The ensemble is therefore a Monte Carlo realization of a source-resolved
probability model, rather than the structure from which the sources are defined
retrospectively, allowing their population-level contributions and downstream
forecast value to be examined separately.
This turns the ensemble from a generator of plausible futures into a controlled
diagnostic of how differently sourced uncertainty propagates through learned
dynamics.

The source categories themselves are not new; the key distinction is where
uncertainty is defined.
Traditional numerical ensembles separate atmospheric-state and model
uncertainty through perturbation design and controlled
experiments~\cite{buizza1999stochastic,leutbecher2008ensemble,palmer2019ecmwf,pickl2022sppt}.
Recent machine-learning systems instantiate different stochastic structures.
WeatherNext Cyclones, through FGN, represents between-model variability with
independently trained models and within-model functional variability with
learned perturbations, while the atmospheric analysis is not an independent
stochastic coordinate~\cite{alet2026operational}.
FuXi-ENS learns flow-dependent state perturbations sampled at initialization
and throughout autoregressive forecasting, but does not separately represent
forecast-model parameters as a second stochastic variable~\cite{fuxiens2025}.
SFNO-BVMC/HENS comes closest to a two-source construction by crossing
bred-vector initial-condition perturbations with independently trained
checkpoints, with checkpoint identity indexing model variability and
bred-vector members indexing initial-condition
variability~\cite{mahesh2025huge1,mahesh2025huge2}.
Part~2 then uses a hierarchical statistical model over these grouped outputs to
compare within- and between-model variability; this analysis characterizes the
realized ensemble rather than constituting the predictive forecasting
distribution that generated it.

Pangu-Bayes differs at the level of probabilistic construction.
Atmospheric-state uncertainty is represented by
\(\eta_t\sim p_{\phi}(\eta_t\mid x_{t-1},x_t)\) and learned-model uncertainty
by \(\theta\sim q(\theta\mid\mathcal D)\) within one predictive hierarchy.
These variables exist before finite sampling, and their crossing generates the
ensemble.
The corresponding variance components are therefore model-defined population
quantities, with the \(8\times6\) ensemble providing Monte Carlo estimates,
while both stochastic mechanisms are refined within the same autoregressive
probabilistic training framework.
The novelty lies neither in the classical state--model distinction, evolving
perturbations, crossed ensembles nor the law of total variance, but in making
source identity part of the learned predictive distribution and linking
model-defined source resolution to source-specific forecast value.
This creates a progression from probabilistic uncertainty construction,
through model-defined source resolution and target-dependent predictive value,
to meteorologically interpretable dynamical manifestations.

The tropical-cyclone experiments show why this distinction matters beyond
variance accounting.
Atmospheric-state variability is more consistently associated with improved
track prediction, whereas learned-model variability is more often associated
with improved intensity prediction, although the latter relationship is weaker
and metric dependent.
These relationships vary across initializations, indicating flow dependence
rather than a fixed storm property.
This asymmetry is dynamically plausible because track is strongly constrained
by large-scale steering and cyclone--environment interactions, whereas
intensity depends more strongly on compact inner-core and thermodynamic
processes incompletely represented by global
models~\cite{emanuel2016predictability,green2013impacts,alet2026operational}.
During Mawar and Khanun, atmospheric-state perturbations sample alternative
surrounding-flow and recurvature evolutions associated with variations in the
western Pacific subtropical high, whereas learned-model sampling more strongly
broadens plausible intensity evolution.
These are preferential source--target associations, not exclusive assignments
or causal partitions.

The results also show that source-resolved variance is only the first step:
uncertainty magnitude is not equivalent to predictive value.
Learned-model component ensembles are generally broader, yet the
atmospheric-state pathway often provides greater value for track.
Thus, larger source-specific spread need not imply stronger alignment with
target-relevant errors.
Spread and calibration remain essential ensemble
diagnostics~\cite{leutbecher2008ensemble,gneiting2007strictly}, but even a
calibrated ensemble can contain uncertainty components with substantially
different downstream usefulness.

The interpretation has clear boundaries.
The variance decomposition is exact for the specified predictive hierarchy but
is not a unique causal decomposition of physical atmospheric uncertainty.
The pathways are operational rather than exhaustive: the atmospheric-state
pathway is learned rather than sampled from a complete data-assimilation
posterior~\cite{carrassi2018data}, while the variational posterior captures
only one tractable aspect of learned-model uncertainty, excluding architecture,
training-data, unresolved-process and distribution-shift
uncertainty~\cite{kendall2017uncertainties,zhu2017big}.
The source--target relationships are established only for the cyclone cohort
and forecasting configuration studied here; whether they persist across years,
basins, phenomena, architectures and forecasting regimes remains open.
The recurvature cases provide dynamical consistency rather than causal
attribution to a single circulation feature.

More broadly, in systems \(x_{t+1}=F_{\theta}(x_t)\), uncertainty in an
imperfectly known evolving state and an imperfect learned transition model can
propagate differently and affect different targets~\cite{cheng2023machine}.
Source-resolved learning therefore asks whether these components remain
separately addressable after nonlinear propagation, acquire different
predictive value, and reveal where forecast improvement should come from.
State-dominated uncertainty may motivate better observations or state
estimation, whereas uncertainty in learned dynamics may motivate changes to
data, model structure or training.
Establishing such actionability will require coupling learned perturbations to
ensemble data assimilation, broadening model uncertainty beyond parameter
variability and testing source resolution under real-time cycling and
distribution shift.
Trustworthy probabilistic prediction may ultimately require not only accurate
and calibrated distributions, but uncertainty representations whose components
can be independently controlled, evaluated and used for scientific diagnosis.

\section{Methods}
Pangu-Bayes combines a global autoregressive forecasting model with explicit atmospheric-state and learned-model uncertainty pathways.
The following sections describe the datasets and forecast initialization, the source-resolved probabilistic formulation and its local dynamical interpretation, model architecture and training, and the evaluation protocols used for tropical-cyclone and global probabilistic forecasts.

\subsection{Datasets and forecast initialization}
\label{sec:datasets_initialization}

Atmospheric fields were represented on a \(0.25^\circ\times0.25^\circ\) latitude--longitude grid at 6-h intervals.
Pangu-Bayes predicts 69 prognostic variables: geopotential (Z), temperature (T), zonal wind (U), meridional wind (V) and specific humidity (Q) on 13 pressure levels (50, 100, 150, 200, 250, 300, 400, 500, 600, 700, 850, 925 and 1000~hPa), together with 2-m temperature (T2M), 10-m zonal and meridional winds (U10 and V10) and mean sea-level pressure (MSL) at the surface.
ERA5 reanalysis~\cite{hersbach2020era5} provides the principal atmospheric dataset for model development and the initialization of the tropical-cyclone experiments, whereas ECMWF IFS HRES-fc0 operational analyses are used for the operational-analysis fine-tuning and matched global evaluation described below.
Additional dataset definitions, preprocessing details and the complete list of atmospheric and static variables are provided in Supplementary Section~\ref{app:dataset} and Supplementary Table~\ref{ec_variable}.

Global probabilistic forecasts are evaluated over the 2022 WeatherBench2 test year~\cite{rasp2024weatherbench}.
Pangu-Bayes is initialized from HRES-fc0 operational-analysis fields at 00:00 and 12:00 UTC and verified against ERA5, matching the initialization convention used for the operational-analysis configurations of the AI baselines.
Forecasts are produced at 6-h intervals to a maximum lead time of 120~h.

Tropical-cyclone observations are obtained from the International Best Track Archive for Climate Stewardship (IBTrACS)~\cite{knapp2010international}.
Best-track records are represented at 6-h intervals, with maximum sustained wind (MSW) and minimum sea-level pressure (MSLP) used to characterize cyclone intensity.
Tropical-cyclone forecasts are evaluated on the held-out 2023 cohort and are initialized from ERA5 at 00:00 and 12:00 UTC, with 6-h forecasts to 120~h.
For source-specific pathway comparisons, the atmospheric-state and learned-model component ensembles use identical ERA5 initial states, cyclone matching and IBTrACS verification, so their differences cannot arise from mismatched initialization or verification datasets.

\subsection{Source-resolved probabilistic formulation}
\label{sec:source_resolved_formulation}

\noindent\textbf{Autoregressive forecasting formulation.}
Pangu-Bayes uses a second-order autoregressive forecast model.
Let \(x_t\) denote the atmospheric state at time \(t\), and let one forecast step correspond to 6~h.
The multi-step predictive distribution over a horizon of \(T\) steps is factorized as
\begin{equation}
p(x_{t+1:t+T}\mid x_t,x_{t-1})
=
\prod_{k=1}^{T}
p(x_{t+k}\mid x_{t+k-1},x_{t+k-2}),
\end{equation}
where \(T=20\) corresponds to a 120-h forecast.
For fixed model parameters \(\theta\), the deterministic forecasting backbone predicts
\begin{equation}
x_{t+1}
=
\mathcal{M}_{\theta}(x_t,x_{t-1}).
\end{equation}

\noindent\textbf{Crossed atmospheric-state and learned-model sampling.}
Pangu-Bayes introduces stochasticity along two explicit sampling axes.
Learned-model uncertainty is represented by sampling forecast-model parameters
\begin{equation}
\theta_i\sim q(\theta\mid\mathcal D),
\qquad i=1,\ldots,I,
\end{equation}
where \(q(\theta\mid\mathcal D)\) is a variational posterior over the shared forecast model.
Atmospheric-state uncertainty is represented by flow-dependent additive perturbations
\begin{equation}
\eta_{t,i,j}
\sim
p_{\phi}
\left(
\eta_t\mid x_{t-1}^{(i,j)},x_t^{(i,j)}
\right),
\qquad
\tilde{x}_{t}^{(i,j)}
=
x_t^{(i,j)}+\eta_{t,i,j},
\qquad
j=1,\ldots,J ,
\end{equation}
where \(p_{\phi}\) is the learned perturbation distribution.
Each crossed ensemble member is propagated autoregressively as
\begin{equation}
x_{t+1}^{(i,j)}
=
\mathcal{M}_{\theta_i}
\left(
\tilde{x}_{t}^{(i,j)},
x_{t-1}^{(i,j)}
\right).
\end{equation}
During inference, \(I=8\) model realizations are crossed with \(J=6\) atmospheric-state realizations, yielding \(48\) ensemble trajectories.
The model realization indexes the learned-model axis, whereas variation within each model realization indexes the atmospheric-state axis.
This hierarchy allows either source to be examined while retaining matched initialization, forecast dynamics and verification, enabling source-specific pathway comparisons within the same forecasting system.

The corresponding one-step predictive distribution marginalizes over both stochastic pathways:
\begin{equation}
p(x_{t+1}\mid x_t,x_{t-1},\mathcal D)
=
\iint
p
\left(
x_{t+1}\mid x_t+\eta_t,x_{t-1},\theta
\right)
p_{\phi}
\left(
\eta_t\mid x_{t-1},x_t
\right)
q(\theta\mid\mathcal D)
\,\mathrm d\eta_t\,\mathrm d\theta .
\end{equation}

\noindent\textbf{Exact source-resolved variance decomposition.}
For any scalar forecast quantity \(Y\) induced by this predictive distribution, the law of total variance gives
\begin{equation}
\mathrm{Var}(Y)
=
\underbrace{
\mathbb{E}_{\theta}
\left[
\mathrm{Var}_{\eta}
\left(
Y\mid\theta
\right)
\right]
}_{V_{\mathrm{state}}}
+
\underbrace{
\mathrm{Var}_{\theta}
\left[
\mathbb{E}_{\eta}
\left(
Y\mid\theta
\right)
\right]
}_{V_{\mathrm{model}}}.
\label{eq:source_resolved_variance}
\end{equation}
The first term measures variability induced by atmospheric-state perturbations within a model realization, whereas the second measures variability among model-realization means after averaging over atmospheric-state perturbations.
This identity is exact for the predictive distribution defined by the specified sampling hierarchy, but the two terms are model-defined and do not constitute a unique causal partition of all physical atmospheric uncertainty.

In practice, the conditional moments are approximated from the \(I\times J\) nested ensemble.
For a scalar ensemble quantity \(Y_{i,j}\), we compute
\begin{equation}
\bar{Y}_{i}
=
\frac{1}{J}\sum_{j=1}^{J}Y_{i,j},
\qquad
\bar{Y}
=
\frac{1}{I}\sum_{i=1}^{I}\bar{Y}_{i},
\end{equation}
and use the plug-in Monte Carlo estimates
\begin{equation}
\widehat{V}_{\mathrm{state}}
=
\frac{1}{I}
\sum_{i=1}^{I}
\frac{1}{J-1}
\sum_{j=1}^{J}
\left(
Y_{i,j}-\bar{Y}_{i}
\right)^2,
\end{equation}
\begin{equation}
\widehat{V}_{\mathrm{model}}
=
\frac{1}{I-1}
\sum_{i=1}^{I}
\left(
\bar{Y}_{i}-\bar{Y}
\right)^2.
\end{equation}
These finite-ensemble quantities are Monte Carlo approximations to the two population terms above; sampling error from finite \(I\) and \(J\) is distinct from the exact population-level variance identity.

\subsection{Local interpretation of uncertainty propagation}
\label{sec:local_uncertainty_propagation}

To interpret how the two uncertainty pathways propagate through the autoregressive forecast, we additionally consider a local tangent-linear approximation around a reference trajectory generated with the posterior-mean parameters \(\bar{\theta}\).
Let \(e_t\) denote the forecast-state perturbation, \(\eta_t\) the atmospheric-state perturbation and \(\Delta\theta=\theta-\bar{\theta}\) the learned-model parameter perturbation.
To first order,
\begin{equation}
e_{t+1}
\approx
A_t(e_t+\eta_t)
+
B_t e_{t-1}
+
G_t\Delta\theta,
\label{eq:local_error_dynamics}
\end{equation}
where \(A_t\) and \(B_t\) are Jacobians with respect to the two atmospheric-state inputs and \(G_t\) is the Jacobian with respect to model parameters.
This expression shows that atmospheric-state and learned-model perturbations enter through distinct local sensitivities but are subsequently amplified, damped or reorganized by the evolving forecast dynamics.
This tangent-linear analysis provides only a local first-order interpretation of uncertainty propagation; it is not required for the exact law-of-total-variance decomposition in Section~\ref{sec:source_resolved_formulation}.
The full augmented-state derivation, covariance theorem and proof are provided in Supplementary Section~\ref{app:local_uncertainty_propagation}.

\subsection{Model architecture}
\label{sec:model_architecture}

\noindent\textbf{Forecast backbone.}
The forecasting backbone follows the three-dimensional Earth-specific Transformer architecture of Pangu-Weather~\cite{bi2023accurate}.
At each forecast step, two consecutive atmospheric states, \(x_{t-1}\) and \(x_t\), are used to predict the 6-h-ahead state \(x_{t+1}\).
Each state contains the 69 prognostic variables described in Section~\ref{sec:datasets_initialization} on the \(0.25^\circ\) global grid.
Upper-air and surface variables are embedded separately and then combined into a common three-dimensional representation.
The encoder--decoder contains two resolution stages with feature dimensions of 192 and 384 and Transformer depths of \(8\) and \(24\), respectively.
Three-dimensional windowed self-attention is used to model interactions across the vertical and horizontal atmospheric dimensions, and skip connections link encoder and decoder representations at corresponding resolutions.
The output layer reconstructs the 69 prognostic fields on the native grid.
Detailed patch embeddings, attention-window configuration, resolution transformations and reconstruction operations are provided in Supplementary Section~\ref{app:network_details}.

\noindent\textbf{Atmospheric-state perturbation model.}
Atmospheric-state uncertainty is represented by a lightweight conditional perturbation network using the same atmospheric input representation as the forecasting backbone.
The perturbation network adopts a shallower encoder--decoder with Transformer depths of \(2\) and \(6\) and a base feature dimension of 192.
Conditioned on the two most recent atmospheric states, it predicts spatially varying mean and standard-deviation fields for additive perturbations,
\begin{equation}
\eta_t
\sim
\mathcal{N}
\left(
\mu_{\eta,t},
\sigma_{\eta,t}^{2}
\right),
\qquad
\tilde{x}_t=x_t+\eta_t .
\end{equation}
The resulting perturbations are therefore flow-dependent rather than fixed perturbations of the initial analysis.
During autoregressive forecasting, the perturbation distribution is recomputed from the evolving atmospheric states, allowing atmospheric-state uncertainty to change with the forecast flow.
Each perturbed state is subsequently propagated by a sampled realization of the forecasting backbone as defined in Section~\ref{sec:source_resolved_formulation}.
Architectural and sampling details of the atmospheric-state perturbation model are provided in Supplementary Section~\ref{app:perturbation_model}.

\noindent\textbf{Probabilistic intensity module.}
Tropical-cyclone intensity is estimated using a lightweight probabilistic intensity module operating on cyclone-level predictors extracted from the global forecast fields.
The input features include the maximum and range of 10-m wind speed, the minimum and range of mean sea-level pressure, the minimum Z500, the range of T850, the current MSW and MSLP, and forecast lead time.
A multilayer perceptron predicts the means and positive standard deviations of future changes in maximum sustained wind and minimum sea-level pressure,
\begin{equation}
\Delta\mathrm{MSW}_{\tau}
\sim
\mathcal{N}
\left(
\mu_{\mathrm{MSW},\tau},
\sigma_{\mathrm{MSW},\tau}^{2}
\right),
\qquad
\Delta\mathrm{MSLP}_{\tau}
\sim
\mathcal{N}
\left(
\mu_{\mathrm{MSLP},\tau},
\sigma_{\mathrm{MSLP},\tau}^{2}
\right).
\end{equation}
This module provides heteroscedastic probabilistic intensity predictions and is used only for the tropical-cyclone intensity and rapid-intensification evaluations.
Global field forecasts, cyclone-track forecasts and the source-specific pathway analyses are obtained directly from the Pangu-Bayes ensemble and do not depend on this intensity module.
Details of the cyclone-centred predictor extraction and probabilistic intensity regression architecture are provided in Appendices~\ref{app:intensity_predictor_extraction} and~\ref{app:intensity_module_details}.

\subsection{Training strategy}
\label{sec:training_strategy}

\noindent\textbf{Deterministic pretraining.}
We first train the forecasting backbone for deterministic 6-h prediction.
Given two consecutive atmospheric states \(x_{t-1}\) and \(x_t\), the model predicts \(x_{t+1}\) by minimizing the area-weighted multi-variable \(L_1\) loss
\begin{equation}
\mathcal{L}_{\mathrm{pre}}
=
\mathcal{L}_1
\left(
\mathcal{M}_{\theta}(x_t,x_{t-1}),
x_{t+1}
\right).
\end{equation}
The weights for the upper-air variables Z, Q, T, U and V are 3.00, 0.60, 1.50, 0.77 and 0.54, respectively, and those for MSL, U10, V10 and T2M are 1.50, 0.77, 0.66 and 3.00.
Upper-air and surface losses are combined with weights 1.0 and 0.25.
Training uses \(6\times10^{4}\) steps, a peak learning rate of \(3\times10^{-4}\), batch size 8, weight decay 0.1 and drop-path rate 0.2.
Further details of the deterministic pretraining objective and the complete hyperparameter configuration are provided in Supplementary Section~\ref{app:pretraining_details}.

\noindent\textbf{Bayesian parameter initialization.}
Learned-model uncertainty is introduced by replacing the deterministic weights with a mean-field Gaussian variational posterior,
\begin{equation}
q(\theta\mid\mathcal D)
=
\prod_i
\mathcal N
\left(
\theta_i;
W_{\mu_i},
W_{\sigma_i}^{2}
\right),
\end{equation}
whose mean is initialized from the pretrained deterministic model.
We use a Gaussian prior centred at the pretrained weights,
\begin{equation}
p(\theta)
=
\prod_i
\mathcal N
\left(
\theta_i;
\theta_i^{\mathrm{pre}},
1
\right),
\end{equation}
and optimize
\begin{equation}
\mathcal L_{\mathrm{Bayes}}
=
\mathbb E_{q(\theta\mid\mathcal D)}
\left[
\mathcal L_1
\left(
\mathcal M_{\theta}(x_t,x_{t-1}),
x_{t+1}
\right)
\right]
+
\beta
\mathrm{KL}
\left(
q(\theta\mid\mathcal D)\,\|\,p(\theta)
\right),
\end{equation}
with \(\beta=10^{-4}\).
This stage is trained for \(3\times10^{4}\) steps with a peak learning rate of \(2\times10^{-5}\), batch size 8 and weight decay 0.1.
The variational parameterization, analytical KL term and full optimization configuration are detailed in Supplementary Section~\ref{app:eu_training}.

\noindent\textbf{Joint probabilistic refinement.}
After Bayesian initialization, the atmospheric-state perturbation parameters \(\phi\) and the variational parameters \(W_{\mu}\) and \(W_{\sigma}\) are jointly refined using an autoregressive fair-CRPS objective.
During distributed training, one posterior-sampled forecast-model realization is assigned to each of eight GPUs, and the perturbation model generates one atmospheric-state perturbation for the corresponding realization.
The resulting eight-member training ensemble is optimized using
\begin{equation}
\mathcal L_{\mathrm{joint}}
=
\mathrm{fCRPS}
\left(
\left\{
\mathcal M_{\theta_m}
\left(
x_{t,m}+\eta_{t,m},
x_{t-1,m}
\right)
\right\}_{m=1}^{8},
x_{t+1}
\right),
\end{equation}
where \(\theta_m\sim q(\theta\mid\mathcal D)\) and
\(\eta_{t,m}\sim p_{\phi}(\eta_t\mid x_{t-1,m},x_{t,m})\).
Training uses a fixed learning rate of \(3\times10^{-6}\), batch size 1 and weight decay 0.1.
The autoregressive rollout is progressively extended: one-step forecasts are trained for 5,000 steps, followed by 3,000 steps each at rollout lengths of 2, 3 and 4, giving 14,000 refinement steps in total.
At inference, six atmospheric-state perturbations are sampled for each of eight posterior model realizations, producing the \(48\)-member crossed ensemble described in Section~\ref{sec:source_resolved_formulation}.
The full autoregressive refinement objective, rollout schedule and optimization settings are provided in Supplementary Section~\ref{app:au_training}.

\noindent\textbf{Operational-analysis fine-tuning.}
For the matched operational-analysis WeatherBench2 evaluation, the jointly refined Pangu-Bayes model is further fine-tuned on HRES-fc0 fields.
The perturbation parameters and variational posterior parameters continue to be optimized jointly with the same autoregressive fCRPS objective, learning rate \(3\times10^{-6}\), batch size 1, weight decay 0.1 and progressive rollout schedule.
This fine-tuning defines the operational-analysis-initialized configuration used for the global comparison, rather than a separate forecasting system.

\noindent\textbf{Probabilistic intensity-module training.}
The probabilistic intensity module is trained independently on cyclone-level predictors and IBTrACS intensity targets using the sum of Gaussian CRPS losses for future changes in MSW and MSLP:
\begin{equation}
\mathcal L_{\mathrm{intensity}}
=
\mathrm{CRPS}_{\mathrm{Gaussian}}
\left(
\mu_{\mathrm{MSW}},
\sigma_{\mathrm{MSW}};
\Delta\mathrm{MSW}_{\tau}
\right)
+
\mathrm{CRPS}_{\mathrm{Gaussian}}
\left(
\mu_{\mathrm{MSLP}},
\sigma_{\mathrm{MSLP}};
\Delta\mathrm{MSLP}_{\tau}
\right).
\end{equation}
The multilayer perceptron contains six hidden layers of width 44 with dropout 0.2.
It is trained for 100 epochs using AdamW with an initial learning rate of \(10^{-4}\), weight decay \(10^{-4}\), batch size 32 and exponential learning-rate decay.
The 2023 tropical-cyclone test cohort is excluded from training and model selection.
Cyclone-centred predictor extraction, the Gaussian CRPS objective and the complete post-processing configuration are provided in Supplementary Section~\ref{app:intensity_predictor_extraction}.

\subsection{Tropical cyclone tracking and benchmark protocol}
\label{sec:tc_benchmark_protocol}

\noindent\textbf{Benchmark systems.}
Tropical-cyclone forecasts are compared with both operational numerical ensembles and leading AI weather forecasting systems.
The numerical baselines are TIGGE IFS and TIGGE GEFS from the TIGGE archive~\cite{bougeault2010thorpex}.
The AI baselines include FourCastNetv2~\cite{bonev2023spherical}, Pangu-Weather~\cite{bi2023accurate}, AIFS~\cite{lang2024aifs}, GenCast~\cite{price2024probabilistic} and WeatherNext Cyclones (WN-C)~\cite{alet2026operational}.
Cyclone-level forecasts for GenCast and WN-C are obtained from Google WeatherLab, whereas trajectories for the remaining global-field forecasting systems are extracted from their forecast fields using the common tracking pipeline described below.

\noindent\textbf{Cyclone tracking and matching.}
For models evaluated from global forecast fields, tropical cyclones are identified using TempestExtremes~\cite{ullrich2021tempestextremes}, based on mean-sea-level-pressure minima collocated with coherent upper-level warm-core structure.
Detected nodes are linked into trajectories using great-circle-distance criteria.
Forecast trajectories are then matched to IBTrACS cyclones using the \texttt{assess.match} routine in HuracanPy.
Forecast and observed nodes are first paired at identical valid times, and candidate pairs are retained when their Haversine separation is no greater than 300~km.
Node-level matches are subsequently aggregated into forecast--observed track pairs.
Intensity information is not used for matching; MSLP and MSW are evaluated only after a forecast trajectory has been associated with an IBTrACS cyclone.
The complete TempestExtremes configuration and trajectory-stitching criteria are provided in Supplementary Section~\ref{app:tc_tracking}.

\noindent\textbf{Fair and raw verification protocols.}
Because forecasting systems differ in their ability to maintain a detectable cyclone, we use two complementary verification protocols.
In the main \emph{fair protocol}, a missing cyclone forecast at a given lead time is replaced by persistence from the initial cyclone state, with cyclone position and intensity held fixed at their initial values.
This ensures that all systems are evaluated against the same IBTrACS verification cases while penalizing failures to sustain a detectable cyclone.
In the complementary \emph{raw protocol}, no persistence replacement is applied and scores are computed only from valid model-generated cyclone forecasts.
Headline relative error reductions are computed from the fair-protocol scores at each lead time and for each competing model. For metric \(q\), competing model \(b\) and lead time \(l\), the relative error reduction of Pangu-Bayes is defined as \begin{equation} R_{q,b,l} = 100 \frac{ E_{q,b,l}-E_{q,\mathrm{PB},l} }{ E_{q,b,l} }, \end{equation} where \(E_{q,b,l}\) and \(E_{q,\mathrm{PB},l}\) denote the corresponding errors of competing model \(b\) and Pangu-Bayes, respectively. For each metric, \(R_{q,b,l}\) is averaged arithmetically over all evaluated lead times and competing models. This gives mean relative reductions of 51.5991\% for DPE, 56.8720\% for track CRPS, 9.3572\% for MSLP MAE, 24.9743\% for MSLP CRPS, 18.9295\% for MSW MAE and 30.8636\% for MSW CRPS. The headline track, pressure and wind reductions are then obtained by averaging the two corresponding metrics, yielding 54.24\%, 17.17\% and 24.90\%, respectively.

\noindent\textbf{Track, intensity and rapid-intensification metrics.}
Track accuracy is evaluated using direct positional error (DPE) and a spherical ensemble CRPS.
For ensemble-member distances \(d_m\) from the observed cyclone centre and pairwise member distances \(d_{m,n}\),
\begin{equation}
\mathrm{DPE}
=
\frac{1}{M}\sum_{m=1}^{M}d_m,
\qquad
\mathrm{CRPS}_{\mathrm{track}}
=
\frac{1}{M}\sum_{m=1}^{M}d_m
-
\frac{1}{2M(M-1)}
\sum_{m=1}^{M}\sum_{n=1}^{M}d_{m,n}.
\end{equation}
Cyclone intensity is evaluated separately for MSLP and MSW using ensemble mean absolute error and the corresponding finite-ensemble CRPS.
Rapid intensification (RI) is defined as an increase in MSW of at least 30~kt within 24~h.
Binary RI skill is quantified using the critical success index,
\begin{equation}
\mathrm{CSI}
=
\frac{\mathrm{TP}}
{\mathrm{TP}+\mathrm{FN}+\mathrm{FP}},
\end{equation}
and the Peirce skill score,
\begin{equation}
\mathrm{PSS}
=
\frac{\mathrm{TP}}
{\mathrm{TP}+\mathrm{FN}}
-
\frac{\mathrm{FP}}
{\mathrm{FP}+\mathrm{TN}}.
\end{equation}

\subsection{Global probabilistic evaluation and dynamical diagnostics}
\label{sec:global_dynamical_evaluation}

\noindent\textbf{Global probabilistic metrics.}
Global probabilistic forecasts are evaluated over the 2022 WeatherBench2 test year~\cite{rasp2024weatherbench}.
Pangu-Bayes is compared with IFS-ENS, GenCast~\cite{price2024probabilistic} and FGN~\cite{alet2026operational}.
For GenCast and FGN, we use the operational-analysis-initialized configurations reported by WeatherBench2; Pangu-Bayes is evaluated under the matched HRES-fc0 initialization described in Section~\ref{sec:datasets_initialization}, with ERA5 used as the common verification reference.
IFS-ENS is included as the operational numerical ensemble baseline.

Global scores are computed on the regular latitude--longitude grid using latitude-dependent grid-cell-area weights \(w_i\), where \(i\) indexes latitude and the same weight is applied to all longitude grid points at that latitude.
For an \(M\)-member ensemble with ensemble mean \(\bar{x}_{i,j}\) and verification \(y_{i,j}\), deterministic skill is measured using latitude-weighted RMSE,
\begin{equation}
\mathrm{RMSE}
=
\left[
\frac{
\sum_{i,j}w_i
\left(
\bar{x}_{i,j}-y_{i,j}
\right)^2
}{
\sum_{i,j}w_i
}
\right]^{1/2}.
\end{equation}
Probabilistic skill is evaluated using the finite-ensemble CRPS,
\begin{equation}
\mathrm{CRPS}
=
\frac{
\sum_{i,j}w_i
\left[
\frac{1}{M}\sum_{m=1}^{M}|x^m_{i,j}-y_{i,j}|
-
\frac{1}{2M(M-1)}
\sum_{m=1}^{M}\sum_{n=1}^{M}
|x^m_{i,j}-x^n_{i,j}|
\right]
}{
\sum_{i,j}w_i
}.
\end{equation}
Ensemble spread is reported on the standard-deviation scale as the area-weighted root mean ensemble variance,
\begin{equation}
\mathrm{Spread}
=
\left[
\frac{
\sum_{i,j}w_i
\mathrm{Var}_{m}
\left(x^m_{i,j}\right)
}{
\sum_{i,j}w_i
}
\right]^{1/2},
\end{equation}
and calibration is summarized by the spread--skill ratio
\begin{equation}
\mathrm{SSR}
=
\sqrt{\frac{M+1}{M}}
\frac{\mathrm{Spread}}{\mathrm{RMSE}}.
\end{equation}
Values of SSR close to one indicate agreement between ensemble dispersion and ensemble-mean error.
Upper-air variables are evaluated independently at each pressure level.
Additional implementation details for area weighting, finite-ensemble CRPS, ensemble spread and the spread--skill ratio are provided in Supplementary Section~\ref{app:global_metrics}.

\noindent\textbf{Cyclone dynamical diagnostics.}
Mawar and Khanun are used as illustrative recurvature cases to examine the dynamical organization of the pathway-specific forecast differences identified statistically across the cyclone cohort.
For the large-scale steering-flow analysis, member-wise 500-hPa geopotential-height fields are examined and the western Pacific subtropical high (WPSH) is represented by the \(5{,}880\)-gpm Z500 contour.
Differences in the western extension and northern boundary of this contour at 120-h lead time are compared across atmospheric-state ensemble members and related qualitatively to the corresponding track trajectories.
These diagnostics are used to assess dynamical consistency with the source--track association, rather than to establish causal attribution of cyclone motion to the WPSH.
Pathway-specific empirical orthogonal function analyses of U10 provide an additional spatial diagnostic of forecast variability, with representative results shown in Supplementary Fig.~\ref{fig:typhoon_kn}.

A large language model (ChatGPT, OpenAI) was used to assist with language editing and manuscript organization; all scientific content, analyses, interpretations and conclusions were reviewed and verified by the authors.

\backmatter
\bmhead*{Data availability}
ERA5 atmospheric reanalysis fields used for model development, tropical-cyclone forecast initialization and global forecast verification were obtained from the Copernicus Climate Data Store (CDS), including ERA5 hourly data on single levels
(\url{https://cds.climate.copernicus.eu/datasets/reanalysis-era5-single-levels})
and pressure levels
(\url{https://cds.climate.copernicus.eu/datasets/reanalysis-era5-pressure-levels}).
ECMWF IFS HRES-fc0 operational-analysis fields used for operational-analysis fine-tuning and initialization of the matched global evaluation were obtained through the WeatherBench2 data resources~\cite{rasp2024weatherbench}.
Tropical-cyclone best-track observations were obtained from the International Best Track Archive for Climate Stewardship (IBTrACS), maintained by the NOAA National Centers for Environmental Information
(\url{https://www.ncei.noaa.gov/products/international-best-track-archive}).
Global benchmark scores for IFS-ENS, GenCast and FGN were obtained from WeatherBench2, and cyclone-level forecasts for GenCast and WeatherNext Cyclones used in the tropical-cyclone comparison were obtained from Google WeatherLab.
These datasets and benchmark resources are publicly available subject to their respective access and licensing terms.

\bmhead*{Code and model availability}
The training and inference code, tropical-cyclone tracking and evaluation scripts are publicly available at
\url{https://github.com/hfutml/pangu-bayes}.
The Pangu-Bayes model checkpoints are publicly available at
\url{https://huggingface.co/xionge/pangu-bayes}.

\bibliography{sn-bibliography}

\clearpage
\appendix

\begin{center}

{\LARGE\bfseries Supplementary Information}

\vspace{1.2em}

{\Large\bfseries
Resolving sources of uncertainty in AI weather forecasting
}

\vspace{1.2em}

Wenbo Hu$^{1\dagger}$,
Xinlei Xiong$^{1\dagger}$,
Shuxun Zhou$^{1}$,
Kaifeng Bi$^{2}$,
Lingxi Xie$^{2}$,
Jun Zhu$^{3}$,
Richang Hong$^{1}$,
Qi Tian$^{2*}$

\vspace{0.8em}

$^{1}$Hefei University of Technology, Hefei, China\\
$^{2}$Huawei Inc., Shenzhen, China\\
$^{3}$Tsinghua University, Beijing, China

\vspace{0.8em}

$\dagger$These authors contributed equally to this work.\\
$^{*}$Correspondence: wywqtian@gmail.com

\vspace{1.2em}

\textbf{This Supplementary Information contains:}\\
Supplementary Sections A--H,\\
Supplementary Tables S1--S8, and\\
Supplementary Figures S1--S11.

\end{center}

\vspace{1.5em}

\setcounter{figure}{0}
\setcounter{table}{0}
\setcounter{theorem}{0}
\renewcommand{\thefigure}{S\arabic{figure}}
\renewcommand{\thetable}{S\arabic{table}}

\makeatletter
\renewcommand{\theHfigure}{S\arabic{figure}}
\renewcommand{\theHtable}{S\arabic{table}}
\makeatother
\section{Datasets}\label{app:dataset}
\textbf{ERA5.} As the fifth generation of ECMWF atmospheric reanalyses, the ERA5 archive~\cite{hersbach2020era5} represents a state-of-the-art global dataset describing the evolution of the atmosphere, land surface, and ocean waves from 1959 to the present. It provides a dense record of hundreds of meteorological variables at an hourly temporal resolution. The production of ERA5 relies on the Integrated Forecast System (IFS) Cy42r1-a configuration that was operational at ECMWF throughout most of 2016-utilizing a sophisticated ensemble 4-dimensional variational (4D-Var) data assimilation system. This system optimally combines vast amounts of historical observations-organized into 12-hour assimilation windows (21:00-09:00 UTC and 09:00-21:00 UTC)-with background model forecasts to produce a physically coherent estimate of the atmospheric state. Although ERA5 is natively generated on a reduced Gaussian grid (T639) with a horizontal spacing of approximately 31~km, we employed the version accessible via the Copernicus Climate Change Service (C3S) Climate Data Store (CDS), which is interpolated to a regular $0.25\degree \times 0.25\degree$ equiangular latitude/longitude grid. Our specific dataset comprises a curated subset of variables (detailed in Supplementary Table~\ref{ec_variable}) sampled on 13 vertical pressure levels: 50, 100, 150, 200, 250, 300, 400, 500, 600, 700, 850, 925, and 1000~hPa. These levels were chosen to ensure compatibility with the standard WeatherBench2 benchmark~\cite{rasp2024weatherbench}. The data covers a 45-year period from January 1, 1979, to December 31, 2023, and was temporally subsampled to 6-hour intervals (00:00, 06:00, 12:00, and 18:00 UTC) for the purpose of this study.\\
\textbf{IBTrACS.} The International Best Track Archive for Climate Stewardship (IBTrACS)~\cite{knapp2010international} currently stands as the most comprehensive and centralized observational archive of global tropical cyclones (TCs), consolidating best-track data from numerous agencies worldwide. This dataset offers unrestricted global coverage of historical TC trajectories, encapsulating critical physical parameters including the geographic location of the cyclone center, intensity metrics, and size estimates. To maintain temporal consistency with standard meteorological analysis, we filtered the dataset to retain only the standard synoptic time steps at 6-hour intervals (00:00, 06:00, 12:00, and 18:00 UTC). A significant challenge in global TC analysis arises from the heterogeneity in intensity definitions across different Regional Specialized Meteorological Centers (RSMCs). To address this, we adopted the standards typically employed by U.S. agencies (consistent with the National Hurricane Center and Joint Typhoon Warning Center protocols): specifically, we utilized the Minimum Sea-Level Pressure (MSLP) and the Maximum Sustained Wind (MSW) speed averaged over a 1-minute period at a standard 10~m altitude, variables that are systematically harmonized within the IBTrACS database. Furthermore, we acknowledge inherent uncertainties related to track initialization across different agencies and observational eras. Since there is no universally applied intensity threshold that triggers the recording of a new track, the first recorded data point for a given TC varies significantly depending on the specific operational procedures and detection capabilities of the reporting agency at the time.\\
\textbf{HRES-fc0.} To evaluate ECMWF HRES operational forecasts, we use “HRES-fc0”~\cite{rasp2024weatherbench}, from the initial time step of each HRES forecast at 00Z, 06Z, 12Z, and 18Z. HRES-fc0 is similar to ERA5, but is generated by the ECMWF NWP system operational at the forecast time, with observations assimilated up to +3 h from each initialization time. It is distinct from the ECMWF “HRES Analysis” archive, which can differ from HRES-fc0 because of land surface analysis and data assimilation differences.\\
\textbf{HRES NaN handling.} A very small fraction of HRES geopotential values at 850 hPa (Z850) and 925 hPa (Z925) are NaNs, distributed approximately uniformly across 2016–2021 and forecast times. They account for about 0.00001\% of Z850 pixels and 0.00000001\% of Z925 pixels, with no measurable impact on performance. For consistency, these rare missing values were filled using the weighted average of neighboring pixels, with weights of 1 for side-to-side neighbors and 0.5 for diagonal neighbors~\cite{lam2023learning}.
\begin{table*}[htbp]
% \small
\centering
\caption{ECMWF variables we used in datasets. The “Type” column indicates whether the variable represents a static property, a time-varying single-level property (e.g., surface variables are included), or a time-varying atmospheric property. The “Long name” and “Short name” columns are ECMWF’s labels. The "level" column represents whether the variable is a surface variable or a upper variable, and indicates how many pressure levels were selected for the upper variables. The “Role” column indicates whether the variable is something Pangu-Bayes takes as input and predicts, or only uses as input context.}
\begin{tabular}{lcccc}
\hline
\multicolumn{1}{c}{Type} & Long name                   & Short name & Levels       & Role            \\ \hline
Atmospheric              & Geopotential                & Z          & 13 levels    & Input/Predicted \\
Atmospheric              & Temperature                 & T          & 13 levels    & Input/Predicted \\
Atmospheric              & Specific humidity          & Q          & 13 levels    & Input/Predicted \\
Atmospheric              & U component of wind      & U          & 13 levels    & Input/Predicted \\
Atmospheric              & V component of wind      & V          & 13 levels    & Input/Predicted \\
Single                   & 2m temperature             & T2M        & Singe level  & Input/Predicted \\
Single                   & 10m u component of wind & U10        & Singe level  & Input/Predicted \\
Single                   & 10m v component of wind & V10        & Singe level  & Input/Predicted \\
Single                   & Mean sea level pressure  & MSL        & Singe level  & Input/Predicted \\ \hline
Static                   & Land binary mask          & LSM        & Single level & Input           \\
Static                   & Soil type                  & SLT        & Single level & Input           \\
Static                   & Orography                   & OR  & Single level & Input           \\ \hline
\end{tabular}
\label{ec_variable}
\end{table*}
\section{TempestExtremes tracker and track matching details}
\label{app:tc_tracking}
The TempestExtremes~\cite{ullrich2021tempestextremes} tracker works in two stages. An overview of these two stages is provided below, using default tracker hyperparameter values where mentioned. The first stage, DetectNodes, finds candidate tropical cyclones where minima in mean sea level pressure (MSL) are co-located with coherent upper-level warm cores:
\begin{itemize}
    \item Initial candidate locations are determined by local minima in MSL. Candidates within 6\degree of another stronger MSL minimum are eliminated.
    \item Each MSL minimum must be surrounded by a closed contour of MSL that is at least 200~Pa (2~hPa) greater than the minimum, with this closed contour occurring within a 5.5\degree great-circle distance of the MSL minimum.
    \item A warm-core criterion requires candidate cyclones to be co-located with a local maximum in the geopotential-thickness field between 300 and 500~hPa, defined as \(Z300-Z500\). This field must decrease by at least \(58.8~\mathrm{m^2\,s^{-2}}\) within a 6.5\degree great-circle distance of the reference maximum.
\end{itemize}

The second stage, StitchNodes, detects plausible cyclone trajectories from these candidate cyclone locations by linking them together in time with the following criteria:
\begin{itemize}
    \item Candidates cannot move more than an 8\degree great circle distance between subsequent time slices.
    \item To filter out short-lived cyclones that are not tropical cyclones, trajectories must last for at least 12 hours.
    \item A threshold analysis is performed on candidate trajectories to ensure that cyclones are sufficiently intense and occur within the target latitude band. In our implementation, each trajectory was required to have at least 2 time slices with wind speed greater than 10~m/s, at least 1 time slice with latitude $\leq 50\degree$, and at least 1 time slice with latitude $\geq -50\degree$.
\end{itemize}
The algorithm’s default hyper-parameters were chosen so that when TempestExtremes is applied to 6-hourly analysis datasets the resulting tracks closely match observed tracks, using the IBTrACS dataset as ground truth. All other TempestExtremes hyperparameters were left as their default values, and the same hyperparameters were used for each model and each analysis dataset.

Following trajectory generation, the predicted cyclone tracks were associated with the IBTrACS, which served as the observational reference dataset. This association was performed using the \texttt{assess.match} routine implemented in the \texttt{huracanpy} Python package. For each forecast file, matching was applied pairwise between the IBTrACS track set and the corresponding set of detected model tracks.

In the matching procedure, only the track identifier, geographical position and valid time were used. Specifically, \texttt{assess.match} first extracts the variables \texttt{track\_id}, \texttt{lon}, \texttt{lat} and \texttt{time} from each track dataset, and then aligns the two track sets by exact timestamp equality. For each pair of co-temporal track nodes, the geographical separation is computed as a great-circle distance using the Haversine formula. Longitudes are internally mapped to a consistent \([-180^\circ,180^\circ]\) convention prior to distance calculation. Candidate point pairs are retained only when their separation does not exceed a prescribed distance threshold, which in this study was set to \texttt{max\_dist}=300~km.

For every candidate pair of tracks, the algorithm then aggregates the retained co-temporal point matches to compute two summary quantities: the number of matched time steps and the mean separation distance over those matched time steps. Because the default argument \texttt{min\_overlap}=0 was used in our implementation, any track pair containing at least one co-temporal point within the 300km threshold was returned as a valid candidate association. The resulting matching table therefore identifies predicted-observed track pairs that satisfy this spatiotemporal proximity criterion, rather than solving a globally optimal assignment problem.

Importantly, the association step was based solely on spatiotemporal proximity and did not use cyclone-intensity information. Variables describing cyclone intensity, including maximum 10m wind speed and minimum sea level pressure, were extracted only after a detected track had been associated with an IBTrACS cyclone identifier. In practice, for each matched pair, the IBTrACS cyclone identifier was assigned to the corresponding detected trajectory, and the full time series of predicted position, wind speed and pressure along that detected track was subsequently exported for downstream verification analyses.
\section{Verification Metrics}\label{app:verification_details}
General Notation:
\begin{itemize}
    \item $x_{i,j,k}^{m}$ denotes the value of the mth of $M$ ensemble members in a forecast from initialisation time indexed by $k=1,2,\dots ,K$, at latitude and longitude indexed by $(i,j)\in \mathcal G$.
    \item $\phi_i$ and $\lambda_j$ denote the latitude and longitude coordinates corresponding to the grid indices $i$ and $j$, respectively.
    \item $y_{i,j,k}$ denotes the corresponding verification target.
    \item $\bar{x}_{i,j,k} = \frac{1}{M} \sum_m x^m_{i,j,k}$ denotes the ensemble mean.
    \item $S^2_{i,j,k} = \frac{1}{M-1} \sum_m (x^m_{i,j,k} - \bar{x}_{i,j,k})^2$ denotes the unbiased estimate of ensemble variance.
    \item $\phi_{obs}, \lambda_{obs}$ denote the observed latitude and longitude of the TC center, and let $\phi_{fct}^{m}, \lambda_{fct}^{m}$ denote the forecast position of the $m$-th ensemble member, where $m=1, \dots, M$
\end{itemize}
\subsection{Global weather forecasts}
\label{app:global_metrics}
All metrics are computed using an area-weighting over grid points. This is because on an equiangular latitude-longitude grid, grid cells at the poles have a much smaller area compared to grid cells at the equator. Weighing all cells equally would result in an inordinate bias towards the polar regions. The latitude weights $w(i)$ are computed as:
\begin{equation}
w(i)=\frac{sin\phi_{i}^{u}-sin\phi_{i}^{l}  }{\frac{1}{I} {\textstyle \sum_{i}^{I}(sin\phi_{i}^{u}-sin\phi_{i}^{l})}  }
\end{equation}
where $\phi_{i}^{u}$ and $\phi_{i}^{l}$ indicate upper and lower latitude bounds, respectively, for the grid cell with latitude index $i$. All vertical (pressure) levels are treated separately. For readability, no level index is included in the equations below.\\
\textbf{Ensemble Mean Root Mean Squared Error (RMSE).}
We first assess the deterministic accuracy of the forecast using the EnsembleMeanRMSE. This metric specifically evaluates the error of the ensemble's central tendency (i.e., the ensemble mean, relative to the ground truth. While it is a standard measure of accuracy, it does not evaluate the probabilistic skill of the ensemble, such as its spread or calibration.
\begin{equation}
\text{RMSE} := \sqrt{\frac{1}{K} \sum_{k} \frac{1}{|\mathcal G|} \sum_{i,j} w_i (\bar{x}_{i,j,k} - y_{i,j,k})^2},
\end{equation}\\
\textbf{Continuous Ranked Probability Score (CRPS).}
Our primary metric for probabilistic skill is the CRPS. It is a proper scoring rule that measures how well the marginal distribution of the forecast ensemble represents the ground-truth observation. It generalizes the Mean Absolute Error (MAE) to probabilistic forecasts and is minimized, in expectation, when the forecast distribution matches the true predictive distribution. We use the CRPS~\cite{fricker2013three}, which provides an unbiased estimate of the expected CRPS for a finite ensemble of size $M$.
\begin{equation}
\text{CRPS} := \frac{1}{K} \sum_{k} \frac{1}{|\mathcal G|} \sum_{i,j} w_i \left( \frac{1}{M} \sum_{m} |x_{i,j,k}^m - y_{i,j,k}| - \frac{1}{2M(M-1)} \sum_{m=1}^{M}\sum_{n=1}^{M} |x_{i,j,k}^m - x_{i,j,k}^{n}| \right).
\end{equation}

\noindent\textbf{Ensemble spread and spread--skill ratio.}
For an ensemble of \(M\) forecasts, let \(x_{m,g,k}\) denote the prediction of member \(m\) at grid point \(g\) for forecast initialization \(k\), and let
\[
\bar{x}_{g,k}
=
\frac{1}{M}
\sum_{m=1}^{M}x_{m,g,k}
\]
be the ensemble mean.
The grid-point ensemble variance is
\begin{equation}
s_{g,k}^{2}
=
\frac{1}{M-1}
\sum_{m=1}^{M}
\left(
x_{m,g,k}-\bar{x}_{g,k}
\right)^2 .
\end{equation}
Global ensemble spread is defined on the standard-deviation scale as the square root of the latitude-weighted mean ensemble variance:
\begin{equation}
\mathrm{Spread}
=
\left[
\frac{1}{K}
\sum_{k=1}^{K}
\sum_{g\in\mathcal{G}}
\widetilde{w}_{g}
s_{g,k}^{2}
\right]^{1/2},
\qquad
\widetilde{w}_{g}
=
\frac{w_g}
{\sum_{g'\in\mathcal{G}}w_{g'}},
\end{equation}
where \(w_g=\cos(\phi_g)\) is the latitude weight.
Thus, spread has the same physical units as the forecast variable and the ensemble-mean RMSE.

Following ref.~\cite{fortin2014should}, the spread--skill ratio includes the finite-ensemble correction
\begin{equation}
\mathrm{SSR}
=
\sqrt{\frac{M+1}{M}}\,
\frac{\mathrm{Spread}}
{\mathrm{RMSE}_{\mathrm{ens\mbox{-}mean}}}.
\end{equation}
Under the exchangeability assumptions of a statistically consistent ensemble, SSR is expected to be close to one.
Values below one indicate underdispersion relative to the ensemble-mean error, whereas values above one indicate overdispersion, although forecast bias can also affect this diagnostic.

\subsection{Probabilistic Tropical Cyclone forecast}
\label{app:tc_metrics}
The track uncertainty ellipse is computed from the ensemble forecast positions \(\{(\lambda_{fct}^{m},\phi_{fct}^{m})\}_{m=1}^M\) at a fixed forecast lead time and initialization indexed by \(k\). Using the notation you defined, the ellipse center is given by the ensemble-mean forecast position,
\begin{equation}
    \bar{\lambda}_{fct}=\frac{1}{M}\sum_{m=1}^M \lambda_{fct}^{m},
\bar{\phi}_{fct}=\frac{1}{M}\sum_{m=1}^M \phi_{fct}^{m}.
\end{equation}
The spread and orientation of the ellipse are determined from the sample covariance matrix of longitude and latitude:
\begin{equation}
    \mathbf{\Sigma}_{track}=
\begin{bmatrix}
\mathrm{Var}(\lambda_{fct}) & \mathrm{Cov}(\lambda_{fct},\phi_{fct})\\
\mathrm{Cov}(\lambda_{fct},\phi_{fct}) & \mathrm{Var}(\phi_{fct})
\end{bmatrix},
\end{equation}
where
\begin{equation}
    \mathrm{Var}(\lambda_{fct})=\frac{1}{M-1}\sum_{m=1}^M\left(\lambda_{fct}^{m}-\bar{\lambda}_{fct}\right)^2,\qquad
\mathrm{Var}(\phi_{fct})=\frac{1}{M-1}\sum_{m=1}^M\left(\phi_{fct}^{m}-\bar{\phi}_{fct}\right)^2,
\end{equation}
and
\begin{equation}
    \mathrm{Cov}(\lambda_{fct},\phi_{fct})=\frac{1}{M-1}\sum_{m=1}^M
\left(\lambda_{fct}^{m}-\bar{\lambda}_{fct}\right)
\left(\phi_{fct}^{m}-\bar{\phi}_{fct}\right).
\end{equation}
By eigendecomposing \(\mathbf{\Sigma}_{track}\), we obtain eigenvalues \(\mu_1\ge \mu_2\) and associated eigenvectors. The major-axis direction is determined by the leading eigenvector, and if its components are \((v_1^{(1)},v_1^{(2)})\), the ellipse rotation angle is
\[
\theta=\arctan2\!\left(v_1^{(2)},\,v_1^{(1)}\right).
\]
With the plotting choice \(n_{\mathrm{std}}=2\), the full major- and minor-axis lengths are
\begin{equation}
    W=2\,n_{\mathrm{std}}\sqrt{\mu_1},\qquad
H=2\,n_{\mathrm{std}}\sqrt{\mu_2}.
\end{equation}
Hence, the track uncertainty ellipse summarizes the ensemble dispersion of TC center forecasts in \((\lambda,\phi)\) space: its center represents the ensemble-mean position, its axis lengths quantify positional uncertainty, and its orientation reflects the principal direction of spread relative to the observed TC center \((\lambda_{obs},\phi_{obs})\).

The intensity uncertainty ellipse is constructed in the same way, but in the two-dimensional intensity phase space formed by minimum sea level pressure and maximum wind speed for the \(M\) ensemble members. Denoting the intensity forecasts of member \(m\) by \(p_{fct}^{m}\) and \(w_{fct}^{m}\), the ellipse center is the ensemble mean,
\begin{equation}
    \bar{p}_{fct}=\frac{1}{M}\sum_{m=1}^M p_{fct}^{m},\qquad
\bar{w}_{fct}=\frac{1}{M}\sum_{m=1}^M w_{fct}^{m},
\end{equation}
and the corresponding covariance matrix is
\begin{equation}
    \mathbf{\Sigma}_{intensity}=
\begin{bmatrix}
\mathrm{Var}(p_{fct}) & \mathrm{Cov}(p_{fct},w_{fct})\\
\mathrm{Cov}(p_{fct},w_{fct}) & \mathrm{Var}(w_{fct})
\end{bmatrix},
\end{equation}
with
\begin{equation}
    \mathrm{Var}(p_{fct})=\frac{1}{M-1}\sum_{m=1}^M\left(p_{fct}^{m}-\bar{p}_{fct}\right)^2,\qquad
\mathrm{Var}(w_{fct})=\frac{1}{M-1}\sum_{m=1}^M\left(w_{fct}^{m}-\bar{w}_{fct}\right)^2,
\end{equation}
and
\begin{equation}
    \mathrm{Cov}(p_{fct},w_{fct})=\frac{1}{M-1}\sum_{m=1}^M
\left(p_{fct}^{m}-\bar{p}_{fct}\right)
\left(w_{fct}^{m}-\bar{w}_{fct}\right).
\end{equation}
After eigendecomposition of \(\mathbf{\Sigma}_{intensity}\), the eigenvalues again determine the axis lengths through
\begin{equation}
    W=2\,n_{\mathrm{std}}\sqrt{\mu_1},\qquad
H=2\,n_{\mathrm{std}}\sqrt{\mu_2},
\end{equation}
while the leading eigenvector determines the tilt angle of the ellipse. Therefore, the intensity uncertainty ellipse provides a compact description of the joint ensemble uncertainty in pressure and wind: a larger ellipse indicates greater spread among members, and a more tilted ellipse indicates stronger covariance between the two forecast intensity variables.

We evaluate tropical cyclone (TC) forecasting performance from two complementary perspectives: track prediction and intensity prediction. Track performance measures how accurately the forecast captures the location of the TC center, while intensity performance assesses the prediction of the cyclone strength in terms of maximum sustained wind and minimum sea-level pressure. Since the forecasting system produces an ensemble of \(M\) members, we consider both deterministic-style error metrics based on member-wise deviations from the observation and probabilistic metrics that evaluate the full predictive distribution.

\noindent\textbf{Direct Positional Error (DPE).} Let \((\phi_{obs}, \lambda_{obs})\) denote the observed latitude and longitude of the TC center, and let \((\phi_{fct}^{m}, \lambda_{fct}^{m})\) denote the forecast position of the \(m\)-th ensemble member, where \(m=1,\dots,M\). The orthodromic distance between the \(m\)-th forecast position and the observed TC center is computed using the Haversine formula:
\begin{equation}
d_m = 2r \arcsin\left(
\sqrt{
\sin^2\left(\frac{\phi_{fct}^{m}-\phi_{obs}}{2}\right)
+\cos(\phi_{obs})\cos(\phi_{fct}^{m})
\sin^2\left(\frac{\lambda_{fct}^{m}-\lambda_{obs}}{2}\right)
}
\right),
\end{equation}
where \(r \approx 6371\) km is the Earth's radius. To summarize the expected positional error of the ensemble forecast, we define the Direct Positional Error as the mean member wise great circle distance:
\begin{equation}
\mathrm{DPE} = \frac{1}{M}\sum_{m=1}^{M} d_m.
\end{equation}

For probabilistic evaluation of track forecasts, we use the Track Continuous Ranked Probability Score (\(\mathrm{CRPS}_{\mathrm{track}}\)). For a finite ensemble, this is approximated by
\begin{equation}
\mathrm{CRPS}_{\mathrm{track}}
=
\frac{1}{M}\sum_{m=1}^{M} d_m
-\frac{1}{2M(M-1)}\sum_{m=1}^{M}\sum_{n=1}^{M} d_{m,n},
\end{equation}
where \(d_{m,n}\) is the great-circle distance between the forecast positions of ensemble members \(m\) and \(n\). A smaller \(\mathrm{CRPS}_{\mathrm{track}}\) indicates a better probabilistic track forecast, balancing both reliability and sharpness.

TC intensity is characterized by two scalar quantities: maximum sustained wind (MSW) and minimum sea level pressure (MSLP). Let \(y\) denote the observed value of an intensity variable and let \(x^m\) denote the corresponding prediction from the \(m\)-th ensemble member. We first evaluate deterministic accuracy using the Mean Absolute Error (MAE) averaged over ensemble members:
\begin{equation}
\mathrm{MAE}_{\text{wind/pressure}} = \frac{1}{M}\sum_{m=1}^{M}\left|x^m-y\right|.
\end{equation}
Accordingly, we report \(\mathrm{MAE}_{\mathrm{wind}}\) for maximum sustained wind and \(\mathrm{MAE}_{\mathrm{pressure}}\) for minimum sea-level pressure.

To further assess the probabilistic quality of intensity forecasts, we compute the Continuous Ranked Probability Score (CRPS) for each scalar intensity variable. For an ensemble forecast \(\{x^m\}_{m=1}^{M}\), this admits the empirical form
\begin{equation}
\mathrm{CRPS}_{\text{wind/pressure}}
=
\frac{1}{M}\sum_{m=1}^{M}|x^m-y|
-\frac{1}{2M(M-1)}\sum_{m=1}^{M}\sum_{n=1}^{M}|x^m-x^n|.
\end{equation}
Using this formulation, we compute \(\mathrm{CRPS}_{\mathrm{wind}}\) for maximum sustained wind and \(\mathrm{CRPS}_{\mathrm{pressure}}\) for minimum sea-level pressure. Lower CRPS values indicate better probabilistic forecasts, reflecting both accurate ensemble centering around the observation and appropriate ensemble spread.

Overall, the combination of DPE and MAE provides an intuitive measure of average deterministic forecast error across ensemble members, while \(\mathrm{CRPS}_{\mathrm{track}}\), \(\mathrm{CRPS}_{\mathrm{wind}}\), and \(\mathrm{CRPS}_{\mathrm{pressure}}\) evaluate the full probabilistic forecast quality. This joint evaluation framework allows us to assess not only whether the ensemble forecast is accurate on average, but also whether its spread appropriately represents the uncertainty of TC track and intensity prediction.

\subsection{Capturing Rapid Intensification}
\label{app:ri_metrics}
Rapid intensification (RI) is an infrequent, but not vanishingly rare, event in tropical cyclone intensity forecasting. By construction, RI corresponds to the upper tail of the intensity-change distribution and is commonly defined using a threshold close to the 95th percentile of intensity change. Owing to this class imbalance, standard regression metrics are not the most appropriate diagnostics for RI prediction. We recast RI forecasting as a binary event-prediction task. Specifically, for each forecast instance, the model predicts whether an RI event will occur within the target verification window, yielding a binary ``yes/no'' forecast rather than a continuous intensity-change estimate.

Under this formulation, performance should be assessed using metrics that explicitly characterize rare-event detection skill. We therefore evaluate RI forecasts using the Critical Success Index(CSI) and the Peirce skill score (PSS), both of which are widely used for event-based forecast verification. These metrics are particularly informative for RI because they quantify complementary aspects of forecast performance: CSI emphasizes the successful identification of RI cases in the presence of misses and false alarms, whereas PSS measures the ability of the forecast to discriminate RI from non-RI events.

\noindent\textbf{Critical Success Index (CSI).} For RI forecasting, CSI measures the fraction of correctly predicted RI events among all cases that were either observed as RI or forecast as RI. It is defined as
\begin{equation}
    \mathrm{CSI} = \frac{\mathrm{TP}}{\mathrm{TP + FN + FP}},
\end{equation}
where $TP$ denotes the number of correctly forecast RI events, $FN$ the number of missed RI events, and $FP$ the number of false RI alarms. Unlike overall accuracy, CSI does not include true negatives and is therefore more informative in imbalanced settings where non RI cases dominate. In the present context, a high CSI indicates that the model can identify RI events while keeping both missed events and false alarms under control.

\noindent\textbf{Peirce skill score (PSS).} For RI forecasting, PSS quantifies the degree to which the model separates RI events from non-RI events. It is defined as the difference between the true positive rate and the false positive rate,
\begin{equation}
    \mathrm{PSS = TPR - FPR},
\end{equation}
where
\begin{equation}
    \mathrm{TPR} = \mathrm{\frac{TP}{TP + FN},\qquad FPR = \frac{FP}{FP + TN}}.
\end{equation}
Here, $TPR$ measures the fraction of observed RI events that are correctly detected, whereas $FPR$ measures the fraction of non-RI cases that are incorrectly forecast as RI. PSS ranges from $-1$ to $1$, with $1$ indicating perfect discrimination, $0$ indicating no skill, and negative values indicating performance worse than random discrimination. In the RI setting, PSS is especially useful because it evaluates event detection relative to false alarms, rather than being dominated by the large number of non-events.
For completeness, the contingency table used to compute these quantities is given below.
\begin{table}[!htbp]
    \centering
    \normalsize
    \begin{tabular}{lcc}
        \toprule
         & Forecast RI occurs & Forecast RI does not occur \\
        \cmidrule{2-3}
        Observed RI occurs & $TP$ & $FN$ \\
        Observed RI does not occur & $FP$ & $TN$ \\
        \bottomrule
    \end{tabular}
    \caption{Contingency table for rapid intensification (RI) event prediction.}
\end{table}
\section{Network Details}\label{app:network_details}
\subsection{Deterministic forecast backbone}

The deterministic forecast model is implemented as a 3D Swin Transformer encoder--decoder operating on both vertically resolved upper-air fields and single-level surface fields. The model receives two consecutive atmospheric states, \(x_{t-1}\) and \(x_t\), and predicts the next 6 h state \(x_{t+1}\). For each time step, the upper-air state contains \(C_{\mathrm{upper}}=5\) variables on \(L=13\) pressure levels and has shape
$
C_{\mathrm{upper}}\times L\times H\times W,
$
where \(H=721\) and \(W=1440\) correspond to the \(0.25^\circ\) global latitude--longitude grid. The surface state contains \(C_{\mathrm{surface}}=4\) variables and has shape
$
C_{\mathrm{surface}}\times H\times W.
$
The two consecutive states are concatenated along the channel dimension before entering the model. Therefore, the effective number of upper-air and surface input channels is doubled relative to a single atmospheric state.

\noindent\textbf{Patch embedding.}
The model first maps the physical input fields to a latent token representation using separate upper-air and surface embedding branches. Let the upper-air input be
$
x^{u}\in\mathbb{R}^{B\times C_u\times Z\times H\times W},
$
and the surface input be
$
x^{s}\in\mathbb{R}^{B\times C_s\times H\times W},
$
where \(B\) denotes the batch size. The upper-air field is divided into non-overlapping 3D patches of size
$
(p_z,p_y,p_x)=(2,4,4).
$
Each patch is flattened along the variable and local spatial dimensions, giving a patch vector of dimension
$
C_u \times p_z \times p_y \times p_x,
$
\(1\times1\) convolution is then used to project each flattened upper-air patch to the latent dimension \(192\).
Surface variables are embedded separately using horizontal patches of size
$
(p_y,p_x)=(4,4).
$
Before projection, static geographical information is concatenated to the surface variables. The static inputs consist of a continuous orographic field, a land--sea mask and a soil-type field. The land--sea mask and soil type are first mapped through learnable embedding tables, with two entries for the land--sea mask and eight entries for soil type. These embedded static fields are concatenated with the surface variables and projected to the same latent dimension \(192\) using a separate \(1\times1\) convolution.
After projection, the embedded surface representation is treated as an additional vertical slice. It is concatenated with the embedded upper-air tokens along the vertical dimension. Thus, the latent grid has the form
$
Z_{\mathrm{emb}}\times H_{\mathrm{emb}}\times W_{\mathrm{emb}},
$
where the first vertical slice corresponds to surface information and the remaining slices correspond to upper-air patches. The latent grid is then flattened into a token sequence
$
X\in\mathbb{R}^{B\times N\times 192},
\
N=Z_{\mathrm{emb}} \times H_{\mathrm{emb}} \times W_{\mathrm{emb}}.
$

\noindent\textbf{3D Swin Transformer blocks.}
The embedded tokens are processed by a two-stage encoder--decoder composed of 3D Swin Transformer blocks. In the deterministic backbone, the two stages have depths \((8,24)\) and feature dimensions \(192\) and \(384\), respectively. Each Swin Transformer block reshapes the token sequence back to a 3D latent grid and computes multi-head self-attention within local 3D windows of size
$
(2,6,12),
$
corresponding to the vertical, latitudinal and longitudinal dimensions.
For a local window containing \(N_w=2\times6\times12\) tokens, the input features are linearly projected to queries, keys and values. Scaled dot-product attention is computed within each local window. Learnable relative-position biases are added to the attention logits to encode local spatial offsets within the 3D window. To allow information exchange across neighbouring windows while preserving computational efficiency, consecutive Swin blocks alternate between non-shifted and shifted window partitioning. For shifted windows, an attention mask is applied to avoid attention across invalid window boundaries.
Each Transformer block contains local window attention followed by a two-layer feed-forward network with GELU activation. Layer normalization, residual connections and stochastic depth are used throughout the block. The hidden dimension of the feed-forward network is set to four times the input feature dimension.

\noindent\textbf{Encoder hierarchy and patch merging}
The encoder contains two hierarchical stages. The first stage operates at the patch-embedded resolution with feature dimension 192. Between the two stages, a PatchMerging layer reduces the horizontal resolution while preserving the vertical resolution. Specifically, for an input feature grid of shape
$
Z\times H\times W\times C,
$
PatchMerging groups each \(2\times2\) neighbourhood in the horizontal plane, concatenates the four feature vectors, applies layer normalization and projects the resulting \(4C\)-dimensional vector to \(2C\). This operation changes the feature grid from
$
Z\times H\times W\times C
$
to approximately
$
Z\times \frac{H}{2}\times \frac{W}{2}\times 2C.
$
Thus, the second encoder stage operates at half the horizontal resolution and with doubled feature dimension, increasing the channel dimension from 192 to 384. The vertical dimension is not downsampled.

\noindent\textbf{Decoder hierarchy and skip connections.}
The decoder mirrors the encoder hierarchy. PatchUpsample layers restore the horizontal resolution by expanding the spatial dimensions by a factor of two. The upsampled features are then processed by 3D Swin Transformer blocks at the recovered resolution. At matching resolutions, encoder features are concatenated with decoder features along the channel dimension through lateral skip connections. The skip connections preserve fine-scale spatial information from the encoder and provide it directly to the reconstruction pathway.

\noindent\textbf{Patch recovery.}
The final decoded tokens are converted back to physical variables using a PatchRecover module. The token sequence is first reshaped into a latent 3D grid. The first vertical slice is interpreted as the surface representation, while the remaining slices correspond to upper-air representations. The upper-air branch projects the latent features to
$
C_{\mathrm{upper}}^{\mathrm{out}}\times p_z\times p_y\times  p_x
$
channels and rearranges them through the inverse of the 3D patch embedding operation to recover the upper-air fields on the native grid. The surface branch projects the surface slice to
$
C_{\mathrm{surface}}^{\mathrm{out}}\times p_y\times p_x
$
channels and rearranges it through the inverse horizontal patch operation to recover the surface fields. Any padded regions introduced to make the grid compatible with patch or window sizes are removed at the output stage, so that the reconstructed fields match the original spatial dimensions.

\subsection{Perturbation model for aleatoric uncertainty}
\label{app:perturbation_model}

The perturbation model is used to represent aleatoric uncertainty by learning a conditional distribution of input-state perturbations. It follows the same input representation and patch-embedding design as the deterministic forecast model. Given two consecutive atmospheric states, \(x_{t-1}\) and \(x_t\), together with the same static geographical fields, the perturbation model encodes upper-air and surface variables using the same patch sizes as the deterministic backbone. Upper-air variables are embedded with 3D patches of size \((2,4,4)\), and surface variables are embedded with horizontal patches of size \((4,4)\). As in the deterministic model, the embedded surface representation is treated as an additional vertical slice and concatenated with the embedded upper-air tokens before being processed by the Transformer blocks.

To reduce computational cost, the perturbation model adopts a shallower two-stage encoder--decoder configuration than the deterministic forecast model. Specifically, it uses depths \((2,6)\), attention heads \((6,12)\), local attention windows of size \((2,6,12)\), and an initial latent dimension of 192. The architecture therefore preserves the same spatial representation as the deterministic model while using fewer Transformer blocks.

Unlike the deterministic forecast model, which predicts the next atmospheric state \(x_{t+1}\), the perturbation model outputs the parameters of a spatially varying perturbation distribution on the native grid. For each upper-air and surface variable, it predicts a mean field and a standard-deviation field,
\[
x_{t,\mu}^{\mathrm{per}},\;
x_{t,\sigma}^{\mathrm{per}}
=
\mathcal P_{\psi}(x_t,x_{t-1}),
\]
where \(\mathcal P_{\psi}\) denotes the perturbation model. The standard-deviation field is constrained to be positive using a shifted softplus transformation.

During ensemble generation, a stochastic perturbation field is sampled from the predicted Gaussian distribution,
\begin{equation}
x_t^{\mathrm{per}}
=
x_{t,\mu}^{\mathrm{per}}
+
x_{t,\sigma}^{\mathrm{per}}\odot \xi,
\qquad
\xi\sim\mathcal N(0,I),
\end{equation}
where \(\odot\) denotes element-wise multiplication. The sampled perturbation \(x_t^{\mathrm{per}}\) is then added to the current atmospheric state to form the perturbed model input,
\begin{equation}
\tilde{x}_t
=
x_t + x_t^{\mathrm{per}}.
\end{equation}
\subsection{Probabilistic post-processing network}
\label{app:intensity_module_details}
The tropical cyclone intensity post-processing model is a lightweight multilayer perceptron that operates on cyclone-level scalar predictors extracted from the forecast fields. The input feature vector is
[
$\max\left(\sqrt{V10^2 + U10^2}\right)$,
$\text{range}\left(\sqrt{V10^2 + U10^2}\right)$,
$\min(\text{MSL})$,
$\text{range}(\text{MSL})$,
$\min(\text{Z500})$,
$\text{range}(\text{T850})$,
$\mathrm{MSW}_{t}$,
$\mathrm{MSLP}_{t}$,
$\tau$]
where \(\tau\in\{6,12,\ldots,120\}\) h is the forecast lead time. The network maps input feature vector through fully connected hidden layers with nonlinear activation and dropout, and projects the hidden representation to four output variables:
$
\mu_{\mathrm{MSW}},
\sigma_{\mathrm{MSW}},
\mu_{\mathrm{MSLP}},
\sigma_{\mathrm{MSLP}}.
$
The scale outputs are constrained to be positive using a softplus transformation. The targets are the future changes in maximum sustained wind and minimum sea-level pressure,
$
\Delta \mathrm{MSW}_{\tau}
=
\mathrm{MSW}_{t+\tau}-\mathrm{MSW}_{t},
\Delta \mathrm{MSLP}_{\tau}
=
\mathrm{MSLP}_{t+\tau}-\mathrm{MSLP}_{t}.
$
The post-processing model therefore defines Gaussian predictive distributions,
$
\Delta \mathrm{MSW}_{\tau}
\sim
\mathcal N(\mu_{\mathrm{MSW}},\sigma_{\mathrm{MSW}}^2),
\Delta \mathrm{MSLP}_{\tau}
\sim
\mathcal N(\mu_{\mathrm{MSLP}},\sigma_{\mathrm{MSLP}}^2).
$
This module calibrates cyclone intensity forecasts by learning lead-time-dependent and state-dependent uncertainty in wind and pressure changes.
\section{Training Details}
The overall training procedure is composed of three stages: (1) pretraining, (2) two Stage Uncertainty Quantification, (3) Post-processing on IBTrACS. We describe each of these stages in detail in the following subsections.
\subsection{Pretraining details}
\label{app:pretraining_details}

The deterministic forecast backbone was pretrained as a supervised 6 h forecasting model. Given two consecutive atmospheric states, \(x_{t-1}\) and \(x_t\), the model predicts the next atmospheric state \(x_{t+1}\) as
$
\hat{x}_{t+1}
=
\mathcal M(x_t,x_{t-1};\theta),
$
where \(\theta\) denotes the deterministic model parameters. Here, \(t-1\), \(t\) and \(t+1\) correspond to consecutive 6 h time steps. Although the trained model can be rolled out autoregressively for longer forecasts, the pretraining objective is defined at the single-step 6 h prediction level.

The deterministic model was trained by minimizing an area-weighted multi-variable \(L_1\) loss. Let \(\hat{x}^{c}_{i,j,t+1}\) and \(x^{c}_{i,j,t+1}\) denote the predicted and verifying values of variable \(c\) at latitude index \(i\), longitude index \(j\), and forecast time \(t+1\). The pretraining loss is
\begin{equation}
\mathcal L_{\mathrm{pretrain}}
=
\frac{1}{|\mathcal G|}
\sum_{(i,j)\in\mathcal G}
\omega_i
\sum_{c}
\alpha_c
\left|
\hat{x}^{c}_{i,j,t+1}
-
x^{c}_{i,j,t+1}
\right|,
\end{equation}
where \(\mathcal G\) denotes the horizontal latitude--longitude grid, \(|\mathcal G|=H\times W\), \(\omega_i\) is the latitude-dependent area weight, and \(\alpha_c\) is the variable-specific loss weight. The latitude weight accounts for the decrease in grid-cell area toward the poles on the regular latitude--longitude grid.
Variable weights were introduced to balance fields with different physical units and error magnitudes. The weights for the upper-air variables \((Z,Q,T,U,V)\) were set to
(3.00,\;0.60,\;1.50,\;0.77,\;0.54),
respectively. The effective weights for the surface variables \((\mathrm{MSLP},U10,V10,T2M)\) were set to
(0.375,\;0.1925,\;0.165,\;0.75),
respectively. These surface weights correspond to applying a coefficient of 0.25 to the surface-variable loss relative to the upper-air-variable loss. The complete training configuration is summarized in Table~\ref{tab:deterministic_training_hyperparameters}.

\begin{table}[!htbp]
    \caption{Training hyperparameters for deterministic pretraining.}
    \label{tab:deterministic_training_hyperparameters}
    \centering
    \normalsize
    \begin{tabular}{ll}
        \toprule
        Hyperparameter & Value \\
        \midrule
        Optimizer & AdamW \\
        LR decay schedule & Cosine \\
        Batch size & 8 \\
        Total training steps & \(6 \times 10^{4}\) \\
        Initial learning rate & \(3 \times 10^{-4}\) \\
        Weight decay & 0.1 \\
        Drop-path rate & 0.2 \\
        Attention dropout rate & 0 \\
        \bottomrule
    \end{tabular}
\end{table}
\subsection{Epistemic uncertainty training}
\label{app:eu_training}

Epistemic uncertainty represents uncertainty in the forecast model parameters. After deterministic pretraining, we converted the deterministic forecast backbone into a Bayesian forecast model by placing a mean-field Gaussian variational posterior over the Bayesian weights. Let \(\theta\) denote the collection of Bayesian parameters and let \(\mathcal D\) denote the training dataset. The variational posterior is factorized over individual parameters as
\begin{equation}
q(\theta\mid\mathcal D)
=
\prod_i
\mathcal N
\left(
\theta_i;
W_{\mu_i},
W_{\sigma_i}^2
\right),
\end{equation}
where \(W_{\mu_i}\) and \(W_{\sigma_i}\) are trainable variational parameters corresponding to the posterior mean and standard deviation of the \(i\)-th Bayesian weight.

The prior distribution was centred at the pretrained deterministic weights. Specifically, if \(\theta_i^{\mathrm{pre}}\) denotes the \(i\)-th pretrained weight, the prior is defined as
\begin{equation}
p(\theta)
=
\prod_i
\mathcal N
\left(
\theta_i;
\theta_i^{\mathrm{pre}},
\sigma_p^2
\right),\qquad
\sigma_p=1.
\end{equation}
This prior regularizes the Bayesian model toward the pretrained deterministic solution while allowing data-driven deviations that represent parameter uncertainty.

Parameter samples were drawn using the reparameterization trick,
\begin{equation}
\theta_i
=
W_{\mu_i}
+
W_{\sigma_i}\epsilon_i,\qquad
\epsilon_i\sim\mathcal N(0,1),
\end{equation}
which expresses stochastic weight sampling as a differentiable transformation of Gaussian noise and allows gradients to be propagated to both \(W_{\mu_i}\) and \(W_{\sigma_i}\).

The variational parameters were optimized by minimizing a variational objective consisting of a forecast loss and a Kullback--Leibler regularization term:
\begin{equation}
\mathcal L_{\mathrm{EU}}
=
\mathbb E_{q(\theta\mid\mathcal D)}
\left[
\mathcal L_1
\left(
\mathcal M(x_t,x_{t-1};\theta),
x_{t+1}
\right)
\right]
+
\beta
\,
\mathrm{KL}
\left(
q(\theta\mid\mathcal D)
\,\|\,
p(\theta)
\right),
\end{equation}
where \(\mathcal L_1(\cdot,\cdot)\) is the same area-weighted multi-variable \(L_1\) forecast loss used during deterministic pretraining, and \(\beta=1\times10^{-4}\) controls the strength of the KL regularization.

Because both the variational posterior and the prior are diagonal Gaussian distributions, the KL divergence can be computed analytically:
\begin{equation}
\mathrm{KL}
\left(
q(\theta\mid\mathcal D)
\,\|\,
p(\theta)
\right)
=
\frac{1}{2}
\sum_i
\left[
\frac{
W_{\sigma_i}^2
+
\left(W_{\mu_i}-\theta_i^{\mathrm{pre}}\right)^2
}{\sigma_p^2}
-
1
+
2\log
\frac{\sigma_p}{W_{\sigma_i}}
\right],\qquad
\sigma_p=1.
\end{equation}

In practice, the expectation over \(q(\theta\mid\mathcal D)\) was estimated using Monte Carlo sampling. For each forward pass, one parameter realization \(\hat{\theta}\sim q(\theta\mid\mathcal D)\) was sampled and used to compute
\begin{equation}
\widehat{\mathcal L}_{\mathrm{EU}}
=
\mathcal L_1
\left(
\mathcal M(x_t,x_{t-1};\hat{\theta}),
x_{t+1}
\right)
+
\beta
\,
\mathrm{KL}
\left(
q(\theta\mid\mathcal D)
\,\|\,
p(\theta)
\right).
\end{equation}

The epistemic uncertainty training stage was initialized from the pretrained deterministic model. The detailed training configuration is summarized in Table~\ref{tab:eu_training_hyperparameters}. During inference, epistemic uncertainty was estimated by repeatedly sampling parameter realizations from \(q(\theta\mid\mathcal D)\) and propagating the same input through the corresponding Bayesian forecast models. The spread among these forecasts represents the parameter-induced component of predictive uncertainty.\\
\begin{table}[!htbp]
    \caption{Training hyperparameters for epistemic uncertainty learning.}
    \label{tab:eu_training_hyperparameters}
    \centering
    \normalsize
    \begin{tabular}{ll}
        \toprule
        Hyperparameter & Value \\
        \midrule
        Optimizer & AdamW \\
        LR decay schedule & Cosine \\
        Batch size & 8 \\
        Warm-up steps & 0 \\
        Total training steps & \(3 \times 10^{4}\) \\
        Initial learning rate & \(2 \times 10^{-5}\) \\
        Weight decay & 0.1 \\
        Drop-path rate & 0.2 \\
        Attention dropout rate & 0 \\
        Prior standard deviation & 1.0 \\
        KL weight \(\beta\) & \(1\times10^{-4}\) \\
        Parameter samples per forward pass & 1 \\
        \bottomrule
    \end{tabular}
\end{table}

\subsection{Aleatoric uncertainty training}
\label{app:au_training}
Aleatoric uncertainty was learned through a stochastic perturbation model that generates input-dependent perturbations to the atmospheric state. Given two consecutive states, \(x_{t-1}\) and \(x_t\), the perturbation model predicts the parameters of a spatially varying Gaussian perturbation distribution,
\begin{equation}
x_{t,\mu}^{\mathrm{per}},\;
x_{t,\sigma}^{\mathrm{per}}
=
\mathcal P_{\phi}(x_t,x_{t-1}),
\end{equation}
where \(\mathcal P_{\phi}\) denotes the perturbation model. The standard-deviation field \(x_{t,\sigma}^{\mathrm{per}}\) is constrained to be positive by a shifted softplus transformation. A stochastic perturbation field is sampled as
\begin{equation}
x_{t,n}^{\mathrm{per}}
=
x_{t,\mu}^{\mathrm{per}}
+
x_{t,\sigma}^{\mathrm{per}}\odot \xi_n,
\xi_n\sim\mathcal N(0,I),
\end{equation}
where \(n\) indexes the ensemble member and \(\odot\) denotes element-wise multiplication. The perturbed atmospheric state used by the forecast model is then
\begin{equation}
\tilde{x}_{t,n}
=
x_t+x_{t,n}^{\mathrm{per}}.
\end{equation}

During aleatoric uncertainty training, the perturbation model and the Bayesian forecast model were optimized jointly.
To form an ensemble efficiently, training was distributed across eight NVIDIA A800 GPUs.
On each GPU, one parameter realization was sampled from the variational posterior of the Bayesian forecast model,
\begin{equation}
\theta_m\sim q(\theta\mid\mathcal D),
\qquad m=1,\ldots,M,
\end{equation}
with \(M=8\) during training.
For each training sample, one perturbed atmospheric state \(\tilde{x}_{t,m}\) was generated on each GPU and paired with the corresponding posterior-sampled parameter realization \(\theta_m\).
The resulting one-step ensemble forecast is
\begin{equation}
\hat{x}_{t+1}^{m}
=
\mathcal M
\left(
\tilde{x}_{t,m},x_{t-1};\theta_m
\right),
\qquad
m=1,\ldots,M.
\end{equation}
For autoregressive training with rollout length \(R\), the same construction was applied recursively, with the predictions from each step used as inputs to the subsequent forecast step.

The resulting nested ensemble was jointly refined using the fair continuous ranked probability score (fCRPS). For an ensemble prediction \(\{\hat{x}_{t+r}^{m}\}_{m=1}^{M}\) at autoregressive step \(r\) and the verifying state \(x_{t+r}\), the fCRPS loss is
\begin{equation}
\mathrm{fCRPS}
=
\frac{1}{|\mathcal G|}
\sum_{(i,j)\in\mathcal G}
\omega_i
\sum_c \alpha_c
\left[
\frac{1}{M}
\sum_{m=1}^{M}
\left|
\hat{x}_{i,j,t+r}^{c,m}
-
x_{i,j,t+r}^{c}
\right|
-
\frac{1}{2M(M-1)}
\sum_{m=1}^{M}
\sum_{n=1}^{M}
\left|
\hat{x}_{i,j,t+r}^{c,m}
-
\hat{x}_{i,j,t+r}^{c,n}
\right|
\right],
\end{equation}
where \(\mathcal G\) is the horizontal grid, \(\omega_i\) is the latitude-dependent area weight and \(\alpha_c\) is the variable-specific weight. The first term measures the distance between ensemble members and the verifying state, whereas the second term accounts for ensemble dispersion using the finite-ensemble correction.

For a rollout length \(R\), the joint probabilistic-refinement objective is the average fCRPS over all autoregressive steps,
\begin{equation}
\mathcal L_{\mathrm{joint}}^{R}
=
\frac{1}{R}
\sum_{r=1}^{R}
\mathrm{fCRPS}
\left(
\{\hat{x}_{t+r}^{m}\}_{m=1}^{M},
x_{t+r}
\right).
\end{equation}
This objective jointly refines the state-perturbation distribution and the variational parameter distribution so that the resulting nested ensemble better matches the verifying atmospheric evolution.

To stabilize autoregressive learning, the rollout length was increased progressively. Training began with a one-step autoregressive objective for 5,000 steps. The rollout length was then increased from two to four steps, with 3,000 additional optimization steps at each horizon. The full aleatoric uncertainty training stage therefore used 14,000 steps in total. During inference, six perturbation samples were drawn for each posterior-sampled forecast model on each GPU, yielding \(8\times6=48\) ensemble members in total. The training hyperparameters are summarized in Table~\ref{tab:au_training_hyperparameters}.

\begin{table}[!htbp]
    \caption{Training hyperparameters for aleatoric uncertainty learning.}
    \label{tab:au_training_hyperparameters}
    \centering
    \normalsize
    \begin{tabular}{ll}
        \toprule
        Hyperparameter & Value \\
        \midrule
        Optimizer & AdamW \\
        LR schedule & Constant \\
        Batch size & 1 \\
        Warm-up steps & 0 \\
        Fixed learning rate & \(3 \times 10^{-6}\) \\
        Weight decay & 0.1 \\
        Drop-path rate & 0.0 \\
        Attention dropout rate & 0 \\

        AR1 training steps & 5,000 \\
        AR2 additional steps & 3,000 \\
        AR3 additional steps & 3,000 \\
        AR4 additional steps & 3,000 \\
        Total training steps & 14,000 \\
        \bottomrule
    \end{tabular}
\end{table}
\subsection{Post-processing on IBTrACS}
\label{app:intensity_predictor_extraction}
To refine the raw forecast into calibrated probabilistic intensity predictions, we trained a lightweight probabilistic artificial neural network (ProbANN) on cyclone-level statistical predictors derived from the forecast fields. For each sample, the input feature vector consisted of the cyclone location and a set of physically motivated intensity-related predictors. To calculate these statistics, we first select a forecast grid covering a \(60^\circ \times 60^\circ\) area centered on the cyclone location at the initial time. A lead-time-dependent spatial mask is then applied so that all predictors are calculated within the masked forecast field.

The mask is constructed on the \(241\times241\) cyclone-centred forecast grid. Let \((i,j)\) denote grid-cell indices and \((i_0,j_0)\) the grid cell corresponding to the cyclone centre. The radial distance used by the mask is measured in grid-cell units rather than geographic degrees:
\begin{equation}
d_{i,j}
=
\sqrt{(i-i_0)^2+(j-j_0)^2}.
\end{equation}

The mask weight \(M(d_{i,j})\in[0,1]\) is defined as
\begin{equation}
M(d_{i,j}) =
\begin{cases}
1.0,
& d_{i,j} \le R_{\mathrm{base}}, \\

\max\left(
0,\,
1.0-\left[k\left(d_{i,j}-R_{\mathrm{base}}\right)+b\right]
\right),
& R_{\mathrm{base}} < d_{i,j}
\le R_{\mathrm{base}}+R_{\mathrm{ext}}, \\

0.0,
& d_{i,j} > R_{\mathrm{base}}+R_{\mathrm{ext}},
\end{cases}
\end{equation}
where \(R_{\mathrm{base}}\) is the radius of the fully retained core region and \(R_{\mathrm{ext}}\) is the width of the outer transition region, both expressed in grid-cell units. The decay slope is \(k=1/R_{\mathrm{ext}}\), and a random intercept \(b\sim\mathcal{U}(0,k)\) is introduced to provide small stochastic variations in the transition region.

For forecast lead times from 6 to 120~h at 6-h intervals, the base radius is set to
\begin{equation}
R_{\mathrm{base}}\in\{6,12,\ldots,120\}
\end{equation}
grid cells, while the extension radius is fixed at
\begin{equation}
R_{\mathrm{ext}}=24
\end{equation}
grid cells. Because the forecast grid has a horizontal resolution of \(0.25^\circ\), these values correspond to base radii from \(1.5^\circ\) to \(30^\circ\), while the transition width corresponds to \(6^\circ\). Thus, the numerical values of \(R_{\mathrm{base}}\) and \(R_{\mathrm{ext}}\) refer to grid-cell distances in the implementation, not distances measured directly in degrees.

The probabilistic post-processing network was trained to calibrate tropical cyclone intensity forecasts. For each matched cyclone case and lead time \(\tau\), the network takes the cyclone-level feature vector described in Supplementary Section~\ref{app:intensity_module_details} as input and predicts Gaussian distributions for the future changes in maximum sustained wind and minimum sea-level pressure,
\begin{equation}
    \Delta \mathrm{MSW}_{\tau}
\sim
\mathcal N(\mu_{\mathrm{MSW}},\sigma_{\mathrm{MSW}}^2),\qquad
\Delta \mathrm{MSLP}_{\tau}
\sim
\mathcal N(\mu_{\mathrm{MSLP}},\sigma_{\mathrm{MSLP}}^2).
\end{equation}
The standard deviations are constrained to be positive using a softplus transformation.

Model parameters were optimized using the Gaussian continuous ranked probability score. For a Gaussian predictive distribution \(\mathcal N(\mu,\sigma^2)\) and observation \(y\), the closed-form CRPS is
\begin{equation}
\mathrm{CRPS}_{\mathcal N}(\mu,\sigma;y)
=
\sigma
\left[
z\left(2\Phi(z)-1\right)
+
2\phi(z)
-
\frac{1}{\sqrt{\pi}}
\right],\qquad
z=\frac{y-\mu}{\sigma},
\end{equation}
where \(\phi(\cdot)\) and \(\Phi(\cdot)\) denote the probability density function and cumulative distribution function of the standard normal distribution, respectively.

For a mini-batch of \(N\) cyclone samples, the post-processing loss was defined as the average Gaussian CRPS over the two intensity-change targets:
\begin{equation}
\mathcal L_{post}
=
\frac{1}{2N}
\sum_{i=1}^{N}
\left[
\mathrm{CRPS}_{\mathcal N}
\left(
\mu_{\mathrm{MSW},i},
\sigma_{\mathrm{MSW},i};
\Delta \mathrm{MSW}_{\tau,i}
\right)
+
\mathrm{CRPS}_{\mathcal N}
\left(
\mu_{\mathrm{MSLP},i},
\sigma_{\mathrm{MSLP},i};
\Delta \mathrm{MSLP}_{\tau,i}
\right)
\right].
\end{equation}
This objective jointly encourages accurate predictive means and calibrated predictive spreads, rather than only minimizing pointwise intensity errors. The full training configuration is summarized in Table~\ref{tab:postprocess_hparams}.

\begin{table}[!htbp]
    \caption{Training hyperparameters for the probabilistic post-processing network.}
    \label{tab:postprocess_hparams}
    \centering
    \normalsize
    \begin{tabular}{ll}
        \toprule
        Hyperparameter & Value \\
        \midrule
        Optimizer & AdamW \\
        LR decay schedule & Exponential \\
        Initial learning rate & \(1 \times 10^{-4}\) \\
        Batch size & 32 \\
        Number of epochs & 100 \\
        Weight decay & \(1 \times 10^{-4}\) \\
        Hidden dimension & 44 \\
        Number of hidden layers & 6 \\
        Activation function & GELU \\
        Dropout rate & 0.2 \\
        \bottomrule
    \end{tabular}
\end{table}
% \clearpage
\section{Extended Results}\label{app:dynamical_diagnostics}
 This section provides additional results supporting the main-text evaluation. We first report cohort-level source--target statistics and forecast coverage, followed by the complementary raw-protocol cyclone comparison. We then provide additional global probabilistic results, cyclone dynamical diagnostics and a computational-cost comparison.

\begin{table}[htbp]
\centering
\caption{
Cohort-level source--target contrasts across the 88 tropical cyclones supporting pathway-level comparison.
}
\label{tab:source_target_summary}
\begin{tabular}{lccc}
\toprule
Directional contrast & Cyclones & Fraction & 95\% Wilson CI \\
\midrule
Atmospheric-state pathway lower for track
& 69/88 & 78.41\% & 68.72--85.72\% \\
Learned-model pathway lower for intensity
& 58/88 & 65.91\% & 55.53--74.96\% \\
Both directional contrasts
& 37/88 & 42.05\% & 32.28--52.48\% \\
\bottomrule
\end{tabular}
\end{table}

\begin{figure}
    \centering
    \includegraphics[width=1.0\linewidth]{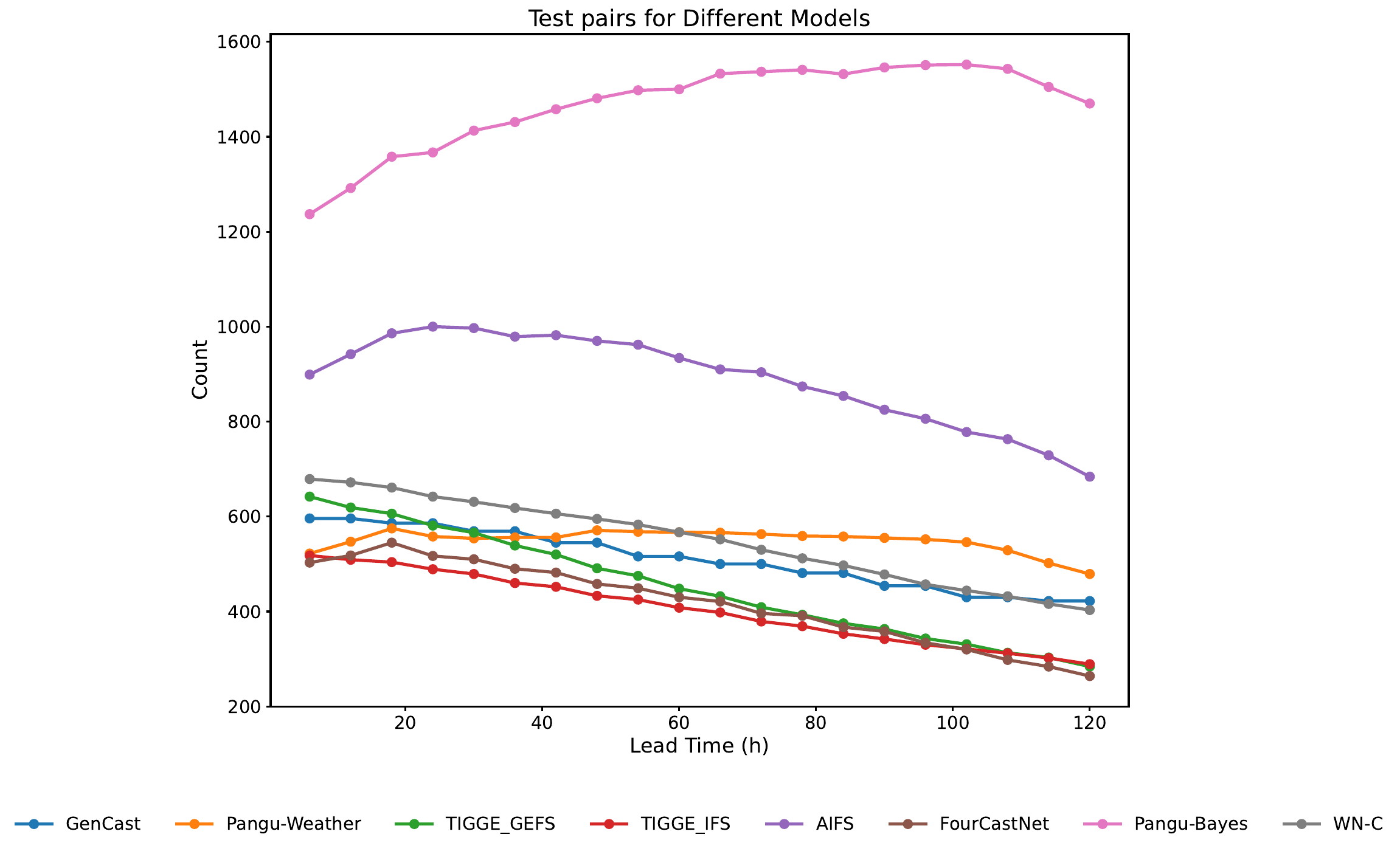}
    \caption{\textbf{Test pairs on the 2023 test set  by lead.} A pair is a unique (SID, t0, t0+L) observed on the 6-hour grid with t0 $\in\left \{ 00,12 \right \}$ UTC. A model covers a pair if it outputs any row for that key.}
    \label{cnt}
\end{figure}

\begin{figure}
    \centering
    \includegraphics[width=1.0\linewidth]{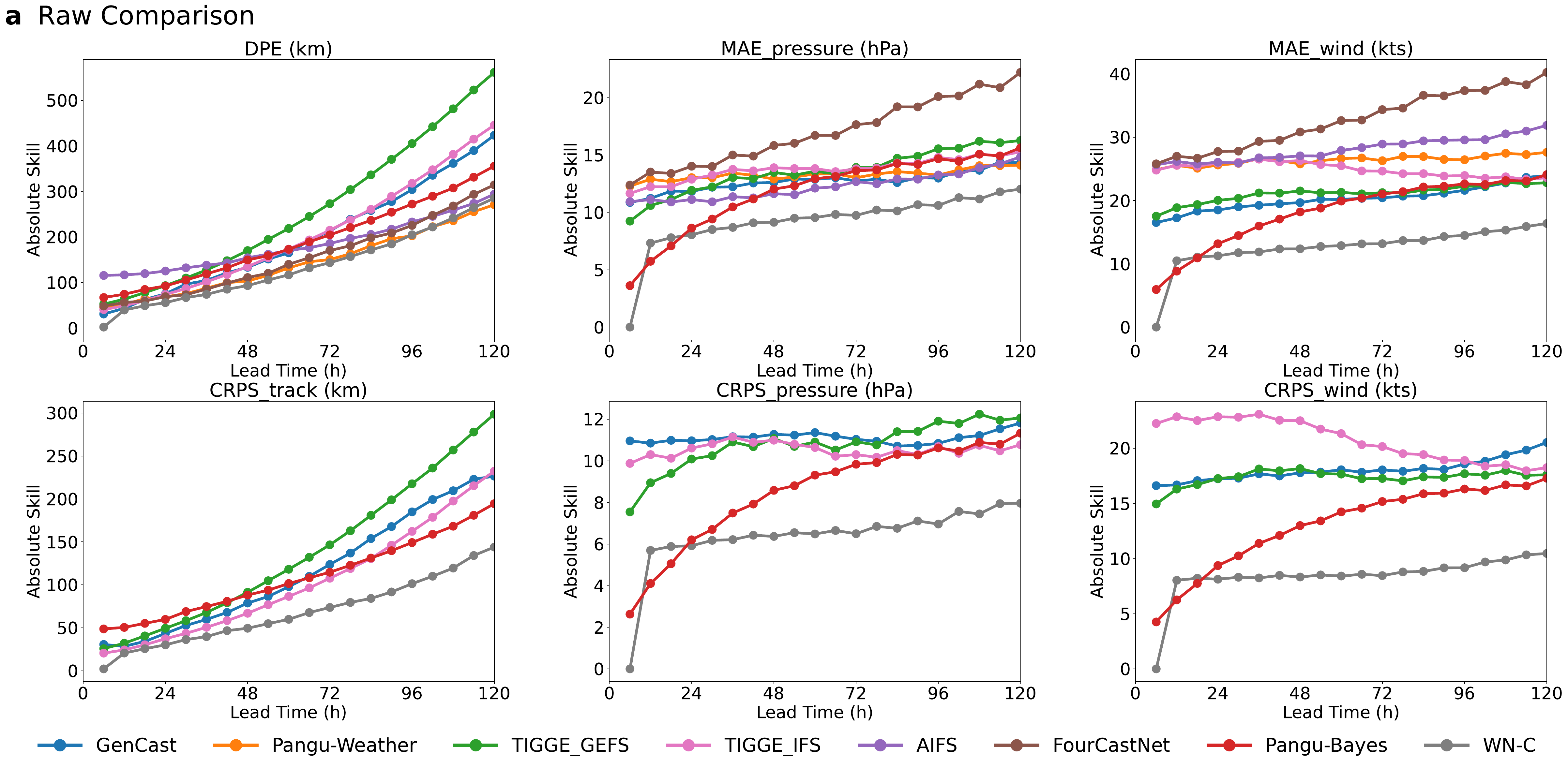}
    \caption{
    \textbf{Tropical cyclone forecast performance under the raw verification protocol.}
    Raw per-lead comparison of deterministic and probabilistic track and intensity scores from 6 to 120~h.
    No persistence filling is applied; each model is evaluated only on the forecast--verification pairs available for that model at each lead time.
    }
    \label{fig:raw_comparison}
\end{figure}

The number of available cyclone forecast--verification pairs decreases with lead time and differs across forecasting systems (Supplementary Fig.~\ref{cnt}). To complement the fair-protocol comparison in the main text, we therefore also report raw scores computed only from the forecasts actually produced by each model (Supplementary Fig.~\ref{fig:raw_comparison}). This comparison provides a sensitivity check on the influence of persistence filling and model-specific cyclone detection coverage.


Additional global probabilistic results are shown in Supplementary Fig.~\ref{z500_v850}. These variables complement those presented in the main text and show that the relative performance of Pangu-Bayes remains variable- and lead-time-dependent rather than indicating uniform superiority across the global benchmark.

\begin{figure}
    \centering
    \includegraphics[width=1.0\linewidth]{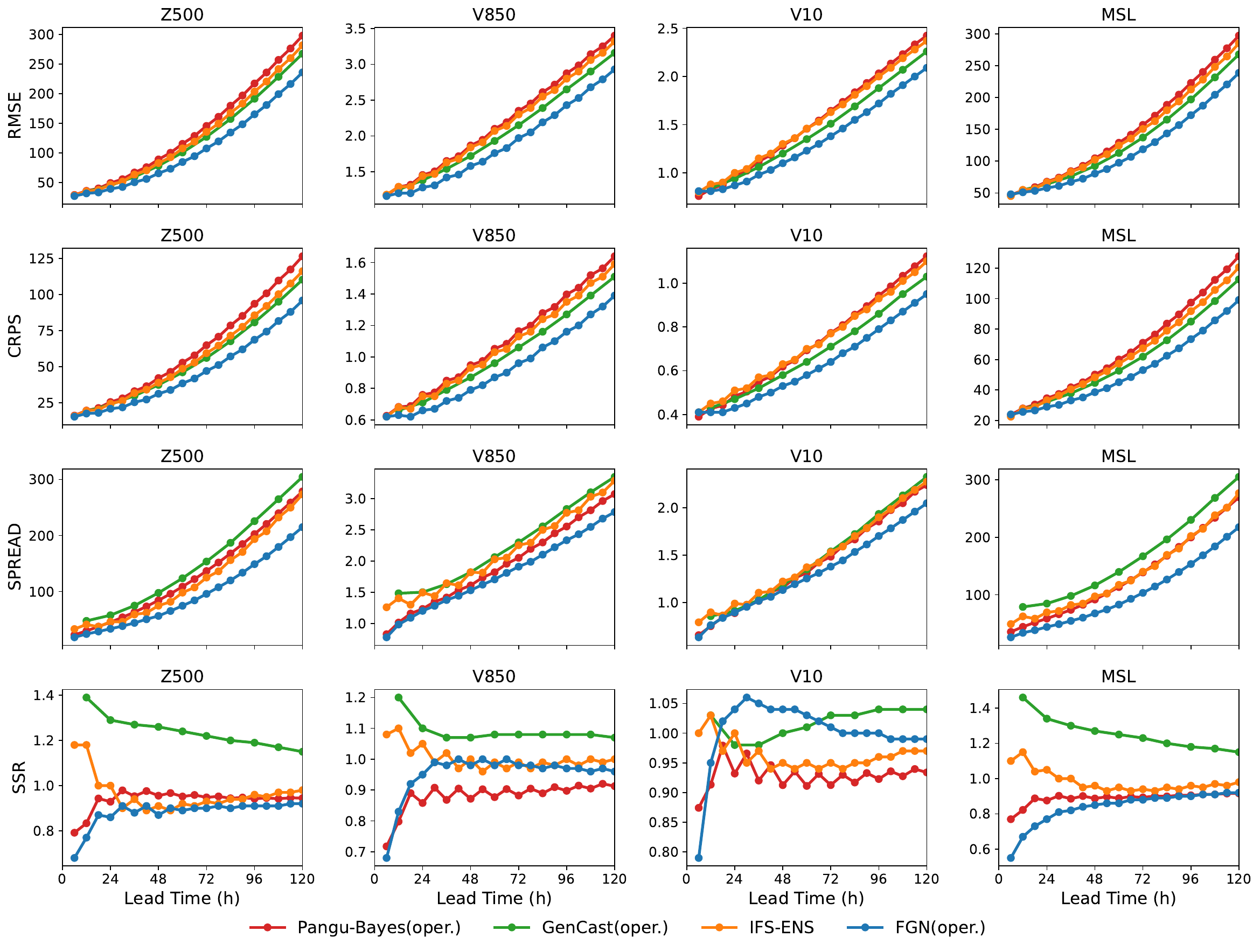}
    \caption{\textbf{Global probabilistic forecast performance.}
    Globally averaged and latitude-weighted RMSE, CRPS, ensemble spread and spread--skill ratio (SSR) for Pangu-Bayes(oper.), IFS-ENS, GenCast(oper.) and FGN(oper.) over the 2022 test year.
    Columns show 500-hPa geopotential (Z500), 850-hPa meridional wind (v850), 10-m meridional wind (v10) and mean sea level pressure (MSL); rows show RMSE, CRPS, ensemble spread and SSR.}
    \label{z500_v850}
\end{figure}

\begin{figure}
    \centering
    \includegraphics[
    width=\linewidth,
    height=0.85\textheight,
    keepaspectratio
]{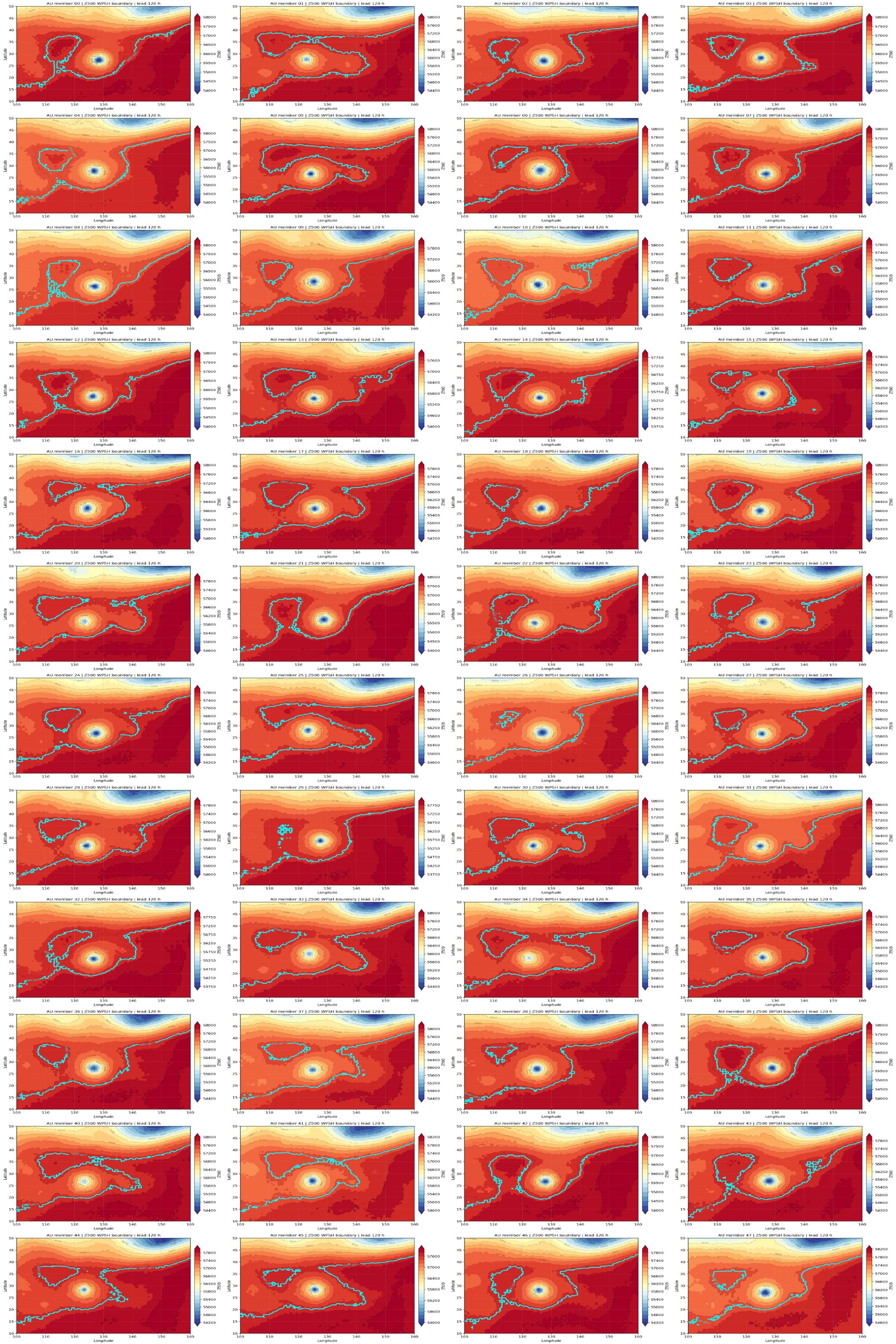}
    \caption{\textbf{Member-wise AU ensemble representation of the 500-hPa geopotential-height field and WPSH boundary at 120-h lead time of KHANUN, initialized
at 2023-07-30 12:00 UTC}}
    \label{WPSH}
\end{figure}
\begin{figure}
    \centering
\includegraphics[
    width=\linewidth,
    height=0.82\textheight,
    keepaspectratio
    ]{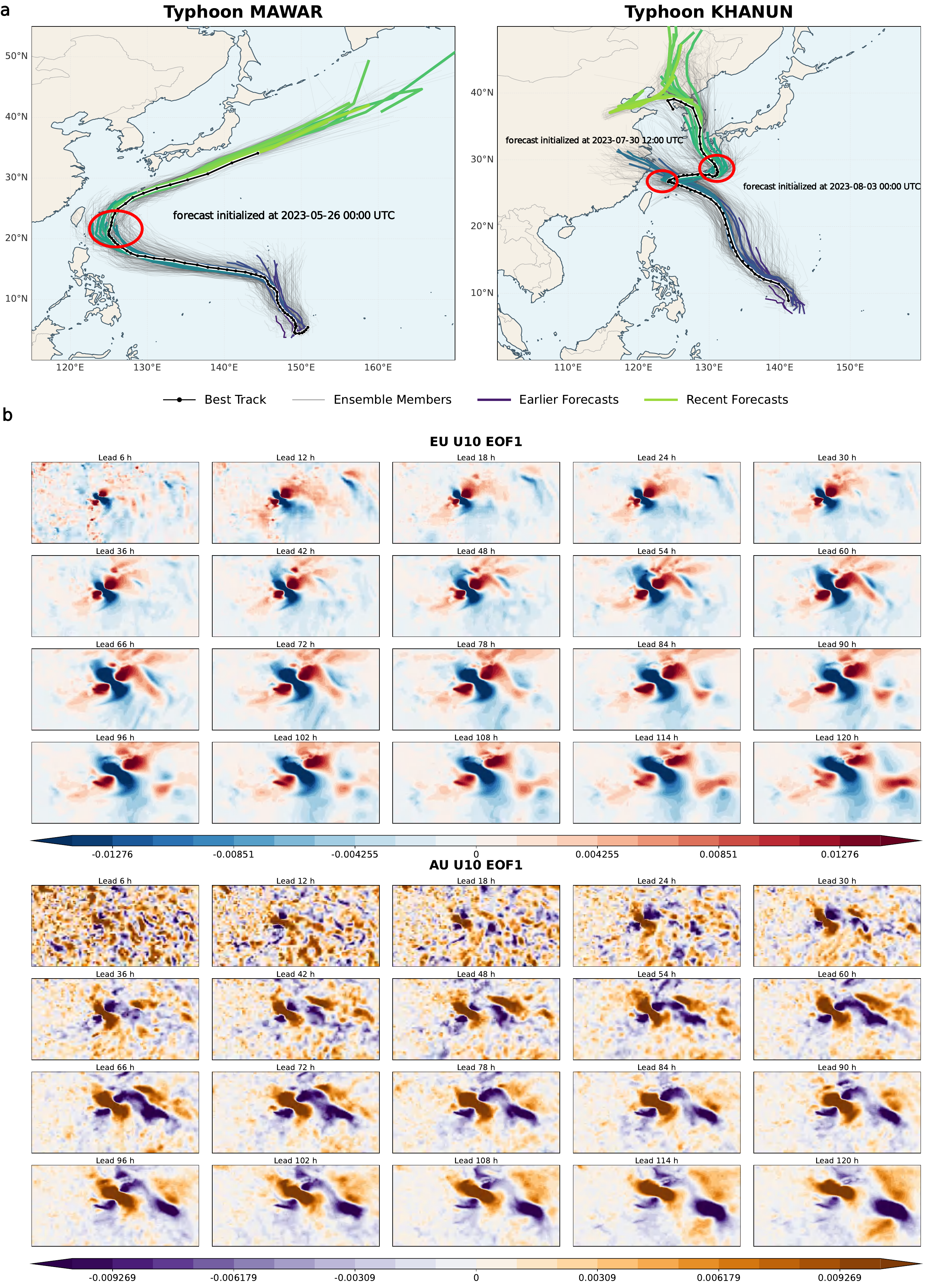}    \caption{
    \textbf{Ensemble track forecasts for Typhoons MAWAR and KHANUN and U10 EOF diagnostics.}
    \textbf{a} shows cyclone-track forecasts from selected initialization times, \textbf{b }shows the lead-time evolution of the leading U10 EOF mode for EU and AU ensembles of KHANUN (forecast initialized at 2023-08-03 00:00 UTC).
    }
    \label{fig:typhoon_kn}
\end{figure}
\begin{figure}[htbp]
    \centering
    \includegraphics[width=1.0\linewidth]{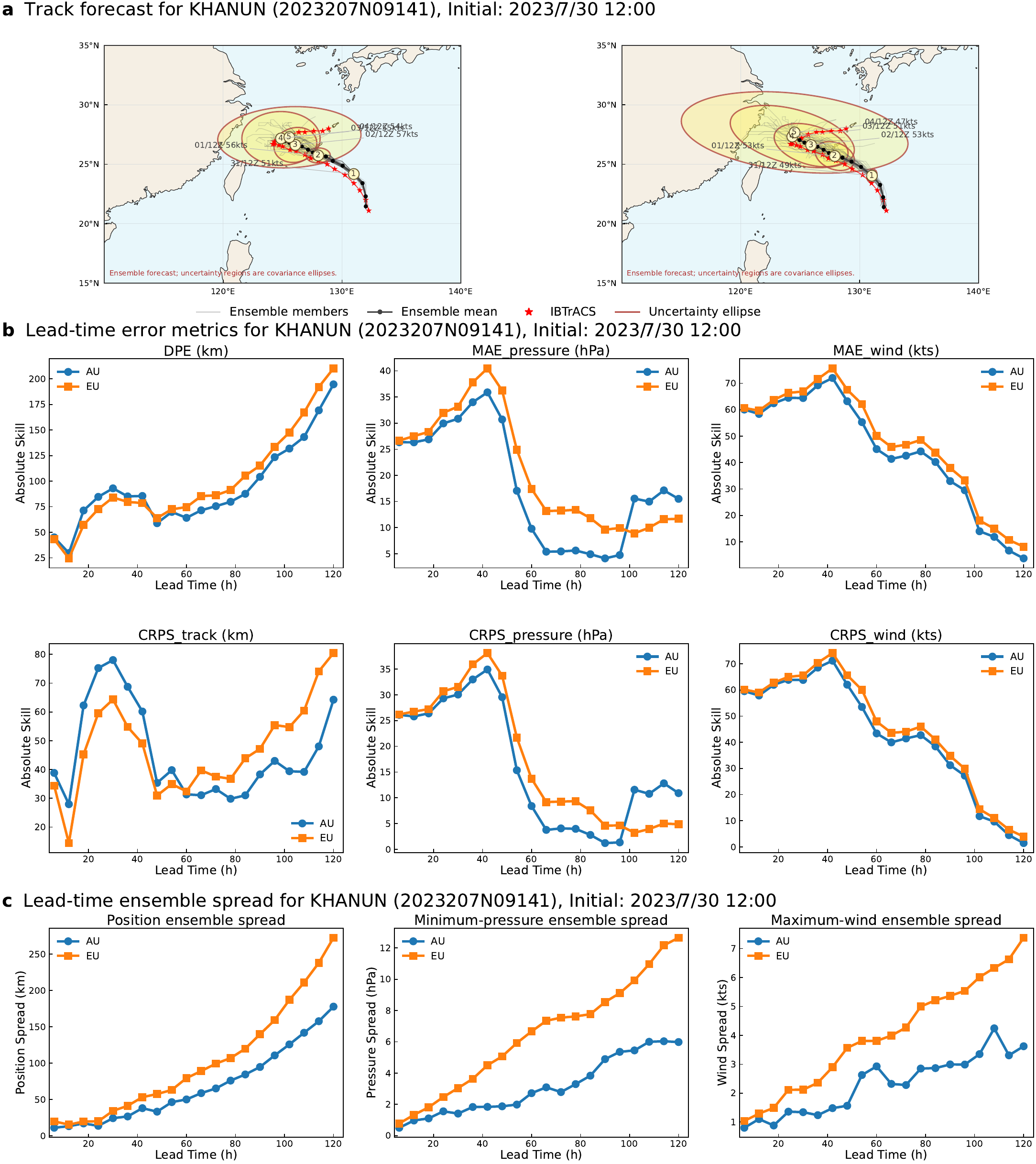}
        \caption{
            \textbf{Joint evolution of track and intensity uncertainty for Typhoon KHANUN, initialized at 12:00 UTC on 26 July 2023.}
            \textbf{a} compares AU and EU track ensembles against the IBTrACS best-track trajectory.
            \textbf{b} shows lead-time-dependent deterministic and probabilistic forecast errors for track, minimum sea-level pressure and maximum sustained wind. Blue circles and orange squares denote AU and EU, respectively.
            \textbf{c}, Lead-time evolution of ensemble spread, quantified using the sample variance across ensemble members.
            Position spread is defined as
            $\sqrt{\mathrm{Var}(x)+\mathrm{Var}(y)}$
            after converting member locations to local Cartesian coordinates, whereas pressure and wind spreads are their respective sample variances.
            Blue circles and orange squares denote the AU and EU component ensembles, respectively.
    }
    \label{730_track_intensity_distribution}
\end{figure}
\begin{figure}[htbp]
    \centering
    \includegraphics[width=1.0\linewidth]{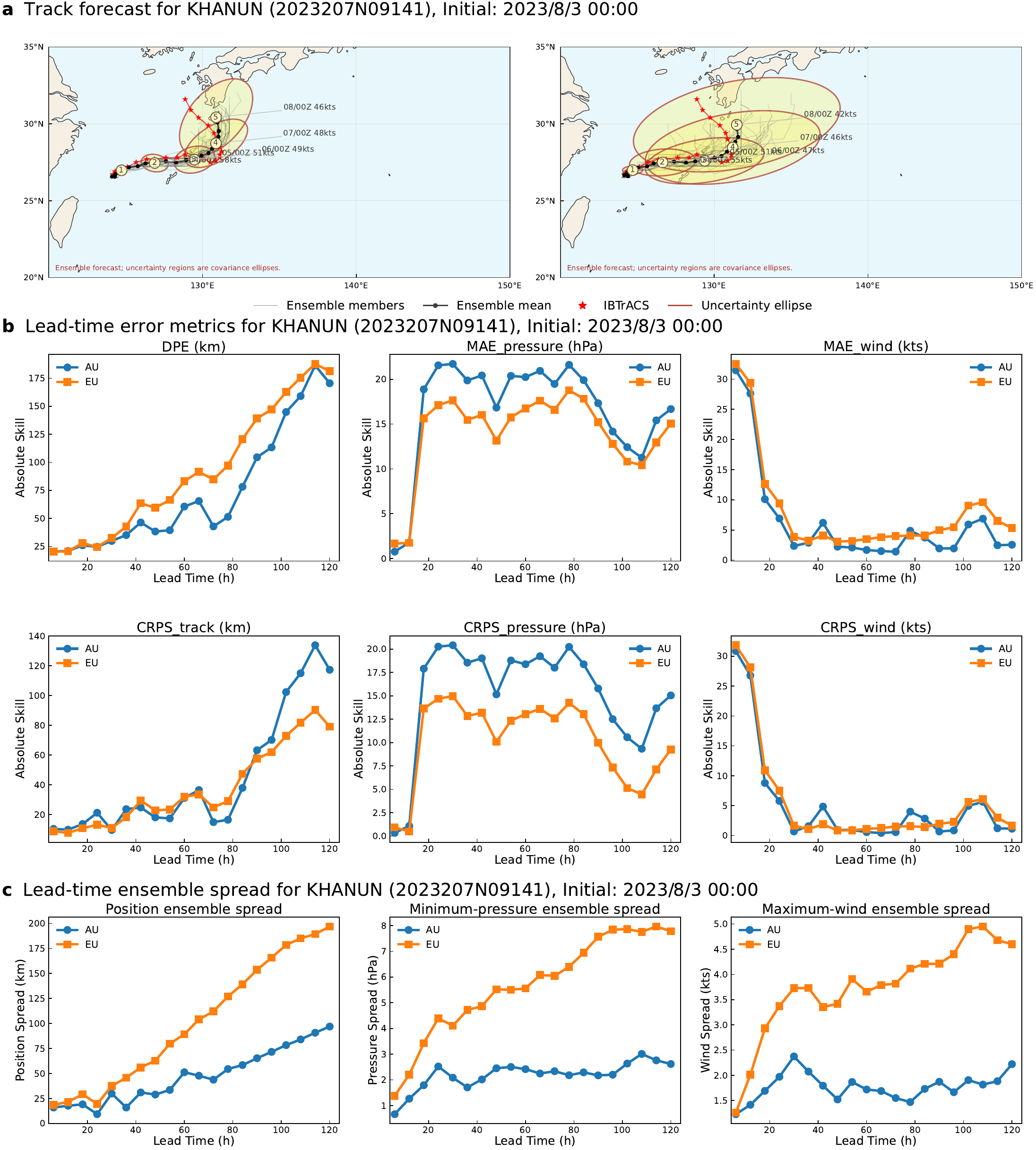}
        \caption{
            \textbf{Joint evolution of track and intensity uncertainty for Typhoon KHANUN, initialized at 00:00 UTC on 26 August 2023.}
            \textbf{a} compares AU and EU track ensembles against the IBTrACS best-track trajectory.
            \textbf{b} shows lead-time-dependent deterministic and probabilistic forecast errors for track, minimum sea-level pressure and maximum sustained wind. Blue circles and orange squares denote AU and EU, respectively.
            \textbf{c}, Lead-time evolution of ensemble spread, quantified using the sample variance across ensemble members.
            Position spread is defined as
            $\sqrt{\mathrm{Var}(x)+\mathrm{Var}(y)}$
            after converting member locations to local Cartesian coordinates, whereas pressure and wind spreads are their respective sample variances.
            Blue circles and orange squares denote the AU and EU component ensembles, respectively.
    }
    \label{803_track_intensity_distribution}
\end{figure}
\begin{table}[htbp]
\centering
\caption{Computational cost and accessibility of probabilistic AI weather forecasting systems.}
\label{tab:compute_comparison}
\setlength{\tabcolsep}{1.5pt}
\renewcommand{\arraystretch}{1.45}
\scriptsize
\begin{tabular}{lcccccc}
\toprule
Model & Training & \makecell[c]{Inference\\(60-steps/member)} & Device & \makecell[c]{BF16 peak\\(TFLOPS/device)} & \makecell[c]{Train cost\\vs PB} & \makecell[c]{Infer cost\\vs PB} \\
\midrule
FGN     & 1960 TPU-days & 1.00 min & 163$^{a}$$\times$(TPUv5p+v6e) & 688.5$^{b}$ & $\times$36.04 & $\times$4.09 \\
GenCast & 160 TPU-days  & 16.00 min & 32$\times$TPUv5p             & 459         & $\times$1.96  & $\times$43.58 \\
Pangu-Bayes & 120 GPU-days & 0.54 min & 8$\times$NVIDIA A800         & 312         & $\times$1.00  & $\times$1.00 \\
\bottomrule
\end{tabular}

\vspace{2pt}
\begin{minipage}{0.98\linewidth}
\footnotesize
$^{a}$ For FGN, the reported cost of 490 TPUv5p/v6e-days per model seed over 3 wall-clock days gives an inferred average of \(490/3 \approx 163\) devices per seed.
This is an inferred average rather than an explicitly reported fixed device count.
$^{b}$ FGN BF16 peak throughput is approximated as the arithmetic mean of TPUv5p (459 TFLOPS) and TPUv6e (918 TFLOPS).
Training-cost ratios are normalized by approximate peak BF16 throughput relative to Pangu-Bayes; inference ratios are computed from per-member 60-steps inference time.
\end{minipage}
\end{table}
% We next compare the computational cost of Pangu-Bayes with recent probabilistic AI weather forecasting systems.
% As summarized in Table~\ref{tab:compute_comparison}, Pangu-Bayes uses substantially fewer training resources than FGN and has a lower per-member inference cost than both FGN and GenCast.
% This efficiency reflects its modelling strategy.
% Like FGN, Pangu-Bayes avoids the multi-step sampling procedure used by fully generative systems such as GenCast and instead generates ensemble members through direct forecast-model evaluations.
% Unlike FGN, however, Pangu-Bayes does not require independently trained large model ensembles; it builds on a strong deterministic backbone and introduces uncertainty through state-dependent input perturbations and Bayesian parameter sampling.
% This design enables competitive probabilistic skill with lower training and inference resources, while preserving a source-resolved interpretation of forecast uncertainty.

% To support reproducibility and reuse, we will release the training and inference code, trained checkpoints and evaluation scripts, enabling the probabilistic forecasts, tropical cyclone diagnostics and uncertainty-decomposition analyses reported here to be reproduced and extended.
\clearpage
\section{Derivation and Proof Details}\label{app:local_uncertainty_propagation}
\noindent\textbf{Tangent linearized error dynamics.}
We consider the deterministic backbone model \(\mathcal M\), whose state evolves according to the second-order autoregressive relation
\begin{equation}
x_{t+1}=\mathcal M(x_t,x_{t-1},\theta).
\end{equation}
Here \(x_t\) denotes the atmospheric state at time \(t\), and \(\theta\) denotes the model parameters. Since the model takes two consecutive states as input, perturbations in both \(x_t\) and \(x_{t-1}\) can contribute to the forecast error at the next time step.

To study the propagation of forecast uncertainty, we linearize the dynamics around a reference trajectory generated by the posterior-mean parameter
\begin{equation}
\bar{\theta}=\mathbb E_{q(\theta\mid\mathcal D)}[\theta].
\end{equation}
The corresponding parameter perturbation is defined as
\begin{equation}
\Delta\theta:=\theta-\bar{\theta},
\qquad
\mathbb E[\Delta\theta]=0,
\qquad
\mathrm{Cov}(\Delta\theta)=\Sigma_\theta .
\end{equation}
We also introduce an additive state perturbation \(\eta_t\) at forecast step \(t\). In the present formulation, \(\eta_t\) is injected into the current input state \(x_t\), while the accumulated influence of previous perturbations is already represented by the propagated error in the previous state.

The reference trajectory is defined by
\begin{equation}
x^*_{t+1}
=
\mathcal M(x_t^*,x_{t-1}^*,\bar{\theta}),
\end{equation}
whereas the perturbed trajectory is written as
\begin{equation}
\tilde{x}_{t+1}
=
\mathcal M(\tilde{x}_t+\eta_t,\tilde{x}_{t-1},\bar{\theta}+\Delta\theta).
\end{equation}
The forecast error is
\begin{equation}
e_t=\tilde{x}_t-x_t^*,
\qquad
e_{t-1}=\tilde{x}_{t-1}-x_{t-1}^*.
\end{equation}
Therefore, the pertur
bed inputs can be expressed relative to the reference trajectory as
\begin{equation}
\tilde{x}_t+\eta_t
=
x_t^*+e_t+\eta_t,
\qquad
\tilde{x}_{t-1}
=
x_{t-1}^*+e_{t-1}.
\end{equation}

Substituting these relations into the definition of \(e_{t+1}\), we obtain
\begin{align}
e_{t+1}
&=
\tilde{x}_{t+1}-x_{t+1}^* \\
&=
\mathcal M(\tilde{x}_t+\eta_t,\tilde{x}_{t-1},\bar{\theta}+\Delta\theta)
-
\mathcal M(x_t^*,x_{t-1}^*,\bar{\theta}) \\
&=
\mathcal M(x_t^*+e_t+\eta_t,\,
x_{t-1}^*+e_{t-1},\,
\bar{\theta}+\Delta\theta)
-
\mathcal M(x_t^*,x_{t-1}^*,\bar{\theta}).
\end{align}

Assume that \(\mathcal M\) is locally differentiable with respect to both state arguments and the parameter. Applying a first-order Taylor expansion around
$
(x_t^*,x_{t-1}^*,\bar{\theta})
$
gives
\begin{align}
\mathcal M(x_t^*+e_t+\eta_t,\,
x_{t-1}^*+e_{t-1},\,
\bar{\theta}+\Delta\theta)
&=
\mathcal M(x_t^*,x_{t-1}^*,\bar{\theta})
\nonumber\\
&\quad+
A_t(e_t+\eta_t)
+
B_t e_{t-1}
+
G_t\Delta\theta
\nonumber\\
&\quad+
\mathcal O\big(\|(e_t+\eta_t,\,e_{t-1},\,\Delta\theta)\|^2\big),
\end{align}
where the local Jacobian operators are defined as
\begin{equation}
A_t:=
\left.
\frac{\partial\mathcal M}{\partial x_t}
\right|_{(x_t^*,x_{t-1}^*,\bar{\theta})},
B_t:=
\left.
\frac{\partial\mathcal M}{\partial x_{t-1}}
\right|_{(x_t^*,x_{t-1}^*,\bar{\theta})},
G_t:=
\left.
\frac{\partial\mathcal M}{\partial\theta}
\right|_{(x_t^*,x_{t-1}^*,\bar{\theta})}.
\end{equation}

Subtracting the reference forecast
\(\mathcal M(x_t^*,x_{t-1}^*,\bar{\theta})\) yields
\begin{align}
e_{t+1}
&=
A_t(e_t+\eta_t)
+
B_t e_{t-1}
+
G_t\Delta\theta
+
\mathcal O\big(\|(e_t+\eta_t,\,e_{t-1},\,\Delta\theta)\|^2\big).
\end{align}
Equivalently,
\begin{align}
e_{t+1}
&=
A_t e_t
+
B_t e_{t-1}
+
A_t\eta_t
+
G_t\Delta\theta
+
\mathcal O\big(\|(e_t+\eta_t,\,e_{t-1},\,\Delta\theta)\|^2\big).
\end{align}
Neglecting higher-order terms gives the tangent linearized error dynamics
\begin{equation}
e_{t+1}
\approx
A_t e_t
+
B_t e_{t-1}
+
A_t\eta_t
+
G_t\Delta\theta .
\end{equation}

This expression separates the next-step forecast error into four contributions. The terms \(A_t e_t\) and \(B_t e_{t-1}\) describe the propagation of pre-existing errors in the current and previous states. The term \(A_t\eta_t\) represents the contribution of the stochastic state perturbation injected at step \(t\). The term \(G_t\Delta\theta\) captures the error induced by uncertainty in the model parameters. Thus, \(A_t\) and \(B_t\) act as local tangent propagators in state space, whereas \(G_t\) maps parameter-space uncertainty into state-space forecast error.

\noindent\textbf{Augmented state representation for error evolution.}
The tangent linearized dynamics derived above show that the next-step error depends on both the current error \(e_t\) and the previous error \(e_{t-1}\):
\begin{equation}
e_{t+1}
\approx
A_t e_t
+
B_t e_{t-1}
+
A_t\eta_t
+
G_t\Delta\theta .
\end{equation}
Thus, the error process is second-order in \(e_t\) and cannot be represented as a first-order Markov process using \(e_t\) alone. To obtain a first-order representation, we augment the error state by stacking the two consecutive errors required by the backbone model:
\begin{equation}
E_t=
\begin{bmatrix}
e_t\\
e_{t-1}
\end{bmatrix},
\qquad
E_0=
\begin{bmatrix}
e_0\\
e_{-1}
\end{bmatrix},
\qquad
P_0=\mathrm{Cov}(E_0).
\end{equation}
Here \(E_0\) is treated as a random initial error, representing uncertainty in the two input states required by the second-order autoregressive model. If the initial states are assumed to be deterministic, then \(P_0=0\).

By definition,
\begin{equation}
E_{t+1}
=
\begin{bmatrix}
e_{t+1}\\
e_t
\end{bmatrix}.
\end{equation}
Substituting the linearized expression for \(e_{t+1}\) gives
\begin{align}
E_{t+1}
&=
\begin{bmatrix}
A_t e_t + B_t e_{t-1} + A_t\eta_t + G_t\Delta\theta\\
e_t
\end{bmatrix} \\
&=
\begin{bmatrix}
A_t & B_t\\
I & 0
\end{bmatrix}
\begin{bmatrix}
e_t\\
e_{t-1}
\end{bmatrix}
+
\begin{bmatrix}
A_t\eta_t\\
0
\end{bmatrix}
+
\begin{bmatrix}
G_t\Delta\theta\\
0
\end{bmatrix}.
\end{align}
Therefore, the second-order error dynamics can be written as the following first-order linear system:
\begin{equation}
E_{t+1}
=
\Phi_t E_t
+
\begin{bmatrix}
A_t\eta_t\\
0
\end{bmatrix}
+
\begin{bmatrix}
G_t\Delta\theta\\
0
\end{bmatrix},
\qquad
\Phi_t=
\begin{bmatrix}
A_t & B_t\\
I & 0
\end{bmatrix}.
\end{equation}
The matrix \(\Phi_t\) is the local companion-form tangent propagator. It transports the augmented error state along the reference trajectory and determines how perturbations are amplified, damped, or redistributed by the local dynamics.

We now iterate this first-order system forward in time. For the first step,
\begin{equation}
E_1
=
\Phi_0 E_0
+
\begin{bmatrix}
A_0\eta_0\\
0
\end{bmatrix}
+
\begin{bmatrix}
G_0\Delta\theta\\
0
\end{bmatrix}.
\end{equation}
For the second step,
\begin{align}
E_2
&=
\Phi_1 E_1
+
\begin{bmatrix}
A_1\eta_1\\
0
\end{bmatrix}
+
\begin{bmatrix}
G_1\Delta\theta\\
0
\end{bmatrix} \\
&=
\Phi_1\Phi_0 E_0
+
\Phi_1
\begin{bmatrix}
A_0\eta_0\\
0
\end{bmatrix}
+
\begin{bmatrix}
A_1\eta_1\\
0
\end{bmatrix}
+
\left(
\Phi_1
\begin{bmatrix}
G_0\\
0
\end{bmatrix}
+
\begin{bmatrix}
G_1\\
0
\end{bmatrix}
\right)
\Delta\theta .
\end{align}
This expansion shows how perturbations injected at earlier times are propagated by the subsequent tangent operators. Repeating the same recursion to lead time \(T\) gives
\begin{equation}
E_T
=
\Phi_{T:0}E_0
+
\sum_{k=0}^{T-1}
\Phi_{T:k+1}
\begin{bmatrix}
A_k\eta_k\\
0
\end{bmatrix}
+
\left(
\sum_{k=0}^{T-1}
\Phi_{T:k+1}
\begin{bmatrix}
G_k\\
0
\end{bmatrix}
\right)\Delta\theta,
\end{equation}
where the time-ordered tangent propagator is defined as
\begin{equation}
\Phi_{T:s}:=
\begin{cases}
\Phi_{T-1}\Phi_{T-2}\cdots\Phi_s, & T>s,\\
I, & T=s.
\end{cases}
\end{equation}
In particular, when \(k=T-1\), \(\Phi_{T:k+1}=\Phi_{T:T}=I\), so perturbations injected at the final step are not further propagated.

This representation separates the lead-time error into three contributions: the propagated initial error, the accumulated stochastic state perturbations, and the parameter-induced error. In the parameter term, the same parameter perturbation \(\Delta\theta\) is kept fixed throughout the rollout, corresponding to the case where each Bayesian ensemble member is generated using one draw from the posterior distribution. The expression above provides the starting point for the covariance calculation in the next section, where these three error pathways are converted into a decomposition of total predictive uncertainty.

\noindent\textbf{Decomposition of forecast uncertainty.}
We characterize the total forecast uncertainty by the covariance of the augmented error state,
\begin{equation}
P_T := \mathrm{Cov}(E_T).
\end{equation}
The following result gives the covariance decomposition induced by initial-condition uncertainty, stochastic state perturbations, and parameter uncertainty.

\begin{theorem}[Decomposition of predictive uncertainty]\label{thm1}
Let the augmented error state at lead time \(T\) be given by
\begin{equation}
E_T
=
\Phi_{T:0}E_0
+
\sum_{k=0}^{T-1}
\Phi_{T:k+1}
\begin{bmatrix}
A_k\eta_k\\
0
\end{bmatrix}
+
\left(
\sum_{k=0}^{T-1}
\Phi_{T:k+1}
\begin{bmatrix}
G_k\\
0
\end{bmatrix}
\right)
\Delta\theta .
\end{equation}
Assume that the initial error \(E_0\), the state perturbations \(\eta_t\), and the parameter perturbation \(\Delta\theta\) are centered and mutually uncorrelated, namely
\begin{equation}
\mathbb E[E_0]=0,\qquad
\mathbb E[\eta_t]=0,\qquad
\mathbb E[\Delta\theta]=0,
\end{equation}
and
\begin{equation}
\mathrm{Cov}(E_0,\eta_t)=0,\qquad
\mathrm{Cov}(E_0,\Delta\theta)=0,\qquad
\mathrm{Cov}(\eta_t,\Delta\theta)=0.
\end{equation}
Then \(P_T=\mathrm{Cov}(E_T)=\mathbb E[E_T E_T^\top]\), and the total predictive uncertainty decomposes as
\begin{equation}
P_T
=
\underbrace{
\Phi_{T:0}P_0\Phi_{T:0}^\top
+
\sum_{k=0}^{T-1}
\sum_{j=0}^{T-1}
\Phi_{T:k+1}\mathbb K_{k,j}\Phi_{T:j+1}^{\top}
}_{\text{Aleatoric uncertainty}}
+
\underbrace{
\left(
\sum_{k=0}^{T-1}
\Phi_{T:k+1}
\begin{bmatrix}
G_k\\
0
\end{bmatrix}
\right)
\Sigma_\theta
\left(
\sum_{k=0}^{T-1}
\Phi_{T:k+1}
\begin{bmatrix}
G_k\\
0
\end{bmatrix}
\right)^{\top}
}_{\text{Epistemic uncertainty}},
\end{equation}
where
\begin{equation}
P_0:=\mathrm{Cov}(E_0),\qquad
\Sigma_\theta:=\mathrm{Cov}(\Delta\theta),
\end{equation}
and
\begin{equation}
\mathbb K_{k,j}
=
\begin{bmatrix}
A_k C_{k,j} A_j^\top & 0\\
0 & 0
\end{bmatrix},
\qquad
C_{k,j}:=\mathrm{Cov}(\eta_k,\eta_j).
\end{equation}
\end{theorem}

\begin{proof}
Starting from the iterated augmented error equation,
\begin{equation}
E_T
=
\Phi_{T:0}E_0
+
\sum_{k=0}^{T-1}
\Phi_{T:k+1}
\begin{bmatrix}
A_k\eta_k\\
0
\end{bmatrix}
+
\left(
\sum_{k=0}^{T-1}
\Phi_{T:k+1}
\begin{bmatrix}
G_k\\
0
\end{bmatrix}
\right)
\Delta\theta ,
\end{equation}
we compute the covariance
\begin{equation}
P_T=\mathrm{Cov}(E_T).
\end{equation}
Since \(E_0\), \(\eta_k\), and \(\Delta\theta\) are centered, \(E_T\) is also centered:
\begin{equation}
\mathbb E[E_T]=0.
\end{equation}
Therefore,
\begin{equation}
P_T=\mathbb E[E_T E_T^\top].
\end{equation}

Substituting the expression for \(E_T\) yields
\begin{align}
P_T
&=
\mathbb E\Bigg[
\bigg(
\Phi_{T:0}E_0
+
\sum_{k=0}^{T-1}
\Phi_{T:k+1}
\begin{bmatrix}
A_k\eta_k\\
0
\end{bmatrix}
+
\left(
\sum_{k=0}^{T-1}
\Phi_{T:k+1}
\begin{bmatrix}
G_k\\
0
\end{bmatrix}
\right)
\Delta\theta
\bigg)
\nonumber\\
&\qquad\qquad\qquad \times
\bigg(
\Phi_{T:0}E_0
+
\sum_{j=0}^{T-1}
\Phi_{T:j+1}
\begin{bmatrix}
A_j\eta_j\\
0
\end{bmatrix}
+
\left(
\sum_{j=0}^{T-1}
\Phi_{T:j+1}
\begin{bmatrix}
G_j\\
0
\end{bmatrix}
\right)
\Delta\theta
\bigg)^\top
\Bigg].
\end{align}

Expanding the product gives three self-covariance terms and six cross-covariance terms:
\begin{align}
P_T
&=
\mathbb E\!\left[
\Phi_{T:0}E_0E_0^\top\Phi_{T:0}^\top
\right]
\nonumber\\
&\quad+
\mathbb E\!\left[
\sum_{k=0}^{T-1}
\sum_{j=0}^{T-1}
\Phi_{T:k+1}
\begin{bmatrix}
A_k\eta_k\\
0
\end{bmatrix}
\begin{bmatrix}
A_j\eta_j\\
0
\end{bmatrix}^{\!\top}
\Phi_{T:j+1}^{\top}
\right]
\nonumber\\
&\quad+
\mathbb E\!\left[
\left(
\sum_{k=0}^{T-1}
\Phi_{T:k+1}
\begin{bmatrix}
G_k\\
0
\end{bmatrix}
\right)
\Delta\theta\Delta\theta^\top
\left(
\sum_{j=0}^{T-1}
\Phi_{T:j+1}
\begin{bmatrix}
G_j\\
0
\end{bmatrix}
\right)^{\!\top}
\right]
\nonumber\\
&\quad+
\mathbb E\!\left[
\Phi_{T:0}E_0
\left(
\sum_{j=0}^{T-1}
\Phi_{T:j+1}
\begin{bmatrix}
A_j\eta_j\\
0
\end{bmatrix}
\right)^{\!\top}
\right]
\nonumber\\
&\quad+
\mathbb E\!\left[
\left(
\sum_{k=0}^{T-1}
\Phi_{T:k+1}
\begin{bmatrix}
A_k\eta_k\\
0
\end{bmatrix}
\right)
E_0^\top\Phi_{T:0}^{\top}
\right]
\nonumber\\
&\quad+
\mathbb E\!\left[
\Phi_{T:0}E_0\Delta\theta^\top
\left(
\sum_{j=0}^{T-1}
\Phi_{T:j+1}
\begin{bmatrix}
G_j\\
0
\end{bmatrix}
\right)^{\!\top}
\right]
\nonumber\\
&\quad+
\mathbb E\!\left[
\left(
\sum_{k=0}^{T-1}
\Phi_{T:k+1}
\begin{bmatrix}
G_k\\
0
\end{bmatrix}
\right)
\Delta\theta E_0^\top\Phi_{T:0}^{\top}
\right]
\nonumber\\
&\quad+
\mathbb E\!\left[
\left(
\sum_{k=0}^{T-1}
\Phi_{T:k+1}
\begin{bmatrix}
A_k\eta_k\\
0
\end{bmatrix}
\right)
\Delta\theta^\top
\left(
\sum_{j=0}^{T-1}
\Phi_{T:j+1}
\begin{bmatrix}
G_j\\
0
\end{bmatrix}
\right)^{\!\top}
\right]
\nonumber\\
&\quad+
\mathbb E\!\left[
\left(
\sum_{k=0}^{T-1}
\Phi_{T:k+1}
\begin{bmatrix}
G_k\\
0
\end{bmatrix}
\right)
\Delta\theta
\left(
\sum_{j=0}^{T-1}
\Phi_{T:j+1}
\begin{bmatrix}
A_j\eta_j\\
0
\end{bmatrix}
\right)^{\!\top}
\right].
\end{align}

By the mutual uncorrelatedness assumptions,
\begin{equation}
\mathrm{Cov}(E_0,\eta_k)=0,\qquad
\mathrm{Cov}(E_0,\Delta\theta)=0,\qquad
\mathrm{Cov}(\eta_k,\Delta\theta)=0,
\end{equation}
all six cross-covariance terms vanish. Thus,
\begin{align}
P_T
&=
\Phi_{T:0}\mathbb E[E_0E_0^\top]\Phi_{T:0}^\top
\nonumber\\
&\quad+
\sum_{k=0}^{T-1}
\sum_{j=0}^{T-1}
\Phi_{T:k+1}
\mathbb E\!\left[
\begin{bmatrix}
A_k\eta_k\\
0
\end{bmatrix}
\begin{bmatrix}
A_j\eta_j\\
0
\end{bmatrix}^{\!\top}
\right]
\Phi_{T:j+1}^{\top}
\nonumber\\
&\quad+
\left(
\sum_{k=0}^{T-1}
\Phi_{T:k+1}
\begin{bmatrix}
G_k\\
0
\end{bmatrix}
\right)
\mathbb E[\Delta\theta\Delta\theta^\top]
\left(
\sum_{j=0}^{T-1}
\Phi_{T:j+1}
\begin{bmatrix}
G_j\\
0
\end{bmatrix}
\right)^{\!\top}.
\end{align}

Using
\begin{equation}
P_0=\mathrm{Cov}(E_0)=\mathbb E[E_0E_0^\top],
\qquad
\Sigma_\theta=\mathrm{Cov}(\Delta\theta)=\mathbb E[\Delta\theta\Delta\theta^\top],
\end{equation}
and
\begin{equation}
C_{k,j}
=
\mathrm{Cov}(\eta_k,\eta_j)
=
\mathbb E[\eta_k\eta_j^\top],
\end{equation}
we obtain
\begin{align}
\mathbb E\!\left[
\begin{bmatrix}
A_k\eta_k\\
0
\end{bmatrix}
\begin{bmatrix}
A_j\eta_j\\
0
\end{bmatrix}^{\!\top}
\right]
&=
\begin{bmatrix}
A_k\mathbb E[\eta_k\eta_j^\top]A_j^\top & 0\\
0 & 0
\end{bmatrix}
\nonumber\\
&=
\begin{bmatrix}
A_k C_{k,j}A_j^\top & 0\\
0 & 0
\end{bmatrix}
=
\mathbb K_{k,j}.
\end{align}

Substituting these identities gives
\begin{align}
P_T
&=
\Phi_{T:0}P_0\Phi_{T:0}^\top
+
\sum_{k=0}^{T-1}
\sum_{j=0}^{T-1}
\Phi_{T:k+1}\mathbb K_{k,j}\Phi_{T:j+1}^{\top}
\nonumber\\
&\quad+
\left(
\sum_{k=0}^{T-1}
\Phi_{T:k+1}
\begin{bmatrix}
G_k\\
0
\end{bmatrix}
\right)
\Sigma_\theta
\left(
\sum_{j=0}^{T-1}
\Phi_{T:j+1}
\begin{bmatrix}
G_j\\
0
\end{bmatrix}
\right)^{\!\top}.
\end{align}
Because \(j\) is a dummy summation index in the last factor, the epistemic term can equivalently be written as
\begin{equation}
\left(
\sum_{k=0}^{T-1}
\Phi_{T:k+1}
\begin{bmatrix}
G_k\\
0
\end{bmatrix}
\right)
\Sigma_\theta
\left(
\sum_{k=0}^{T-1}
\Phi_{T:k+1}
\begin{bmatrix}
G_k\\
0
\end{bmatrix}
\right)^{\!\top}.
\end{equation}
This proves the stated decomposition.
\end{proof}

The first component in Theorem~\ref{thm1} represents state-originated uncertainty. It contains the propagation of initial-condition error through the tangent linear dynamics and the accumulated contribution of stochastic state perturbations \(\eta_t\), modulated by the local state-sensitivity operators \(A_t\). The second component represents epistemic uncertainty. It is governed by the parameter-sensitivity operators \(G_t\) and describes how uncertainty in the neural-network parameters is mapped into state-space error and propagated along the forecast trajectory through the operators \(\Phi_{T:k+1}\).

\section{Visualization}
\subsection{Cyclone Track}
\begin{figure}[htbp]
    \centering
    \includegraphics[
    width=\linewidth,
    height=0.82\textheight,
    keepaspectratio
]{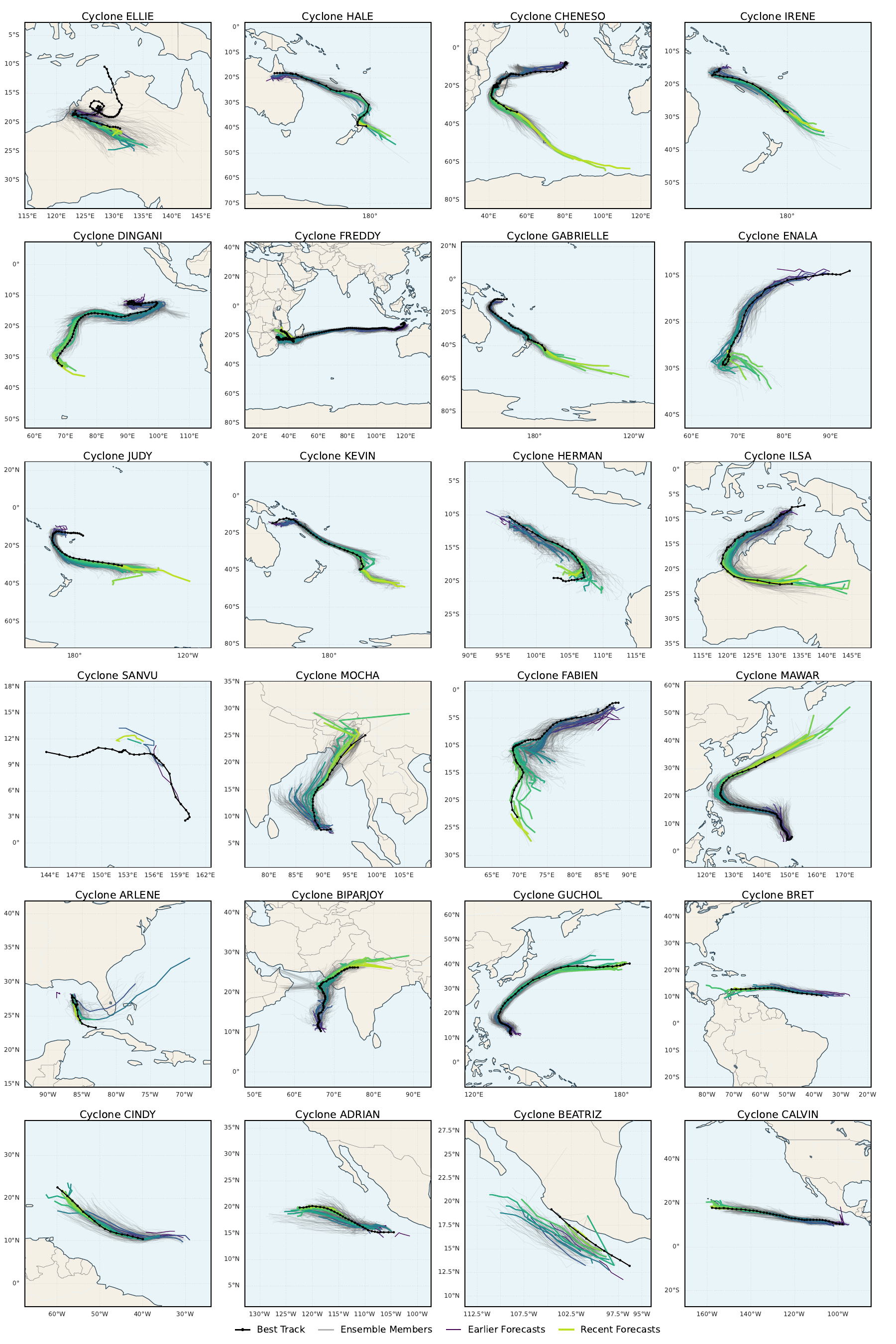}
    \caption{Comparison of ensemble track forecasts for multiple tropical cyclones across all initialization times}
    \label{All_Typhoons_4col6row_Part1}
\end{figure}

\begin{figure}[htbp]
    \centering
 \includegraphics[
        width=\linewidth,
        height=0.78\textheight,
        keepaspectratio
    ]{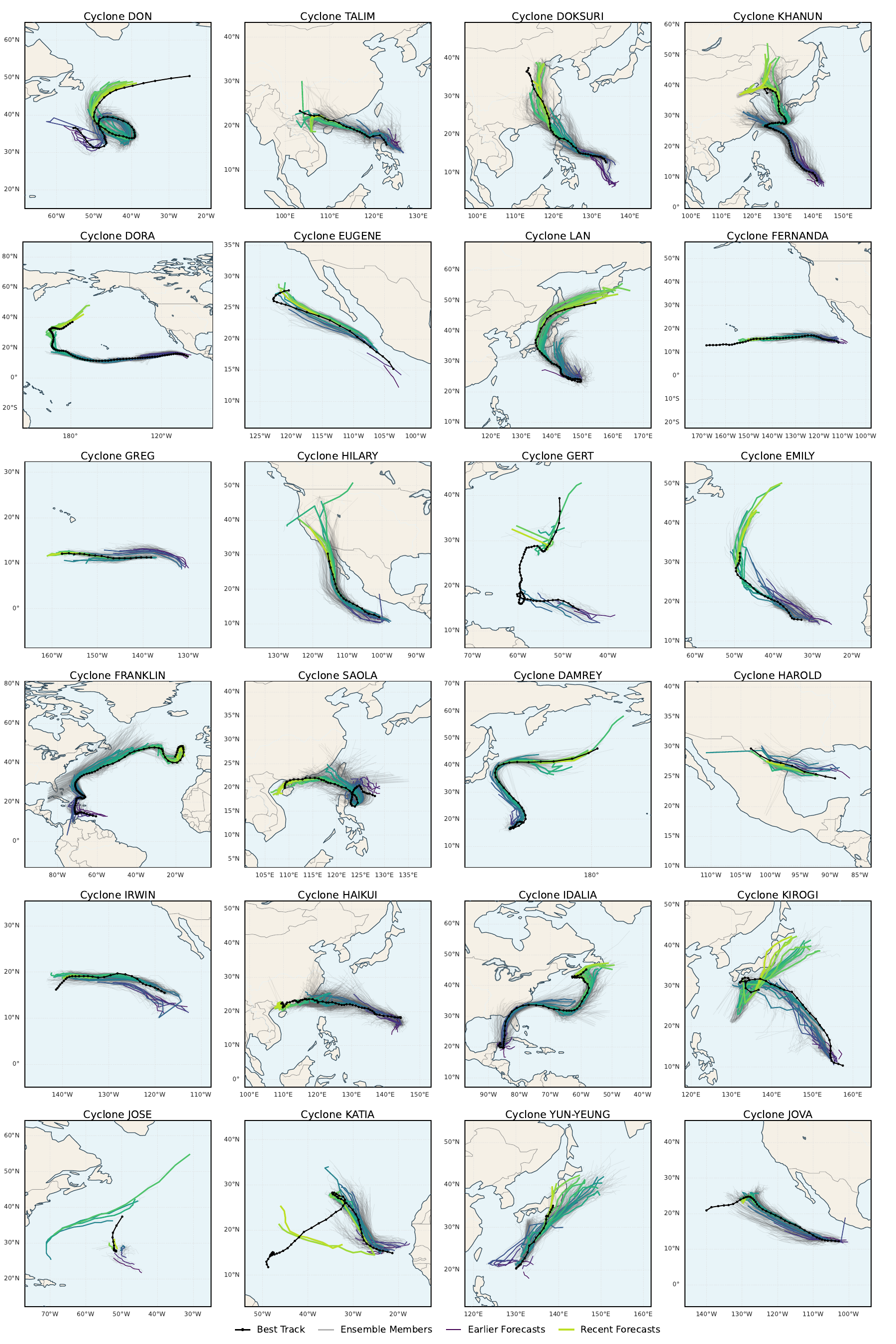}
    \caption{Comparison of ensemble track forecasts for multiple tropical cyclones across all initialization times}
    \label{All_Typhoons_4col6row_Part2}
\end{figure}

\begin{figure}[htbp]
    \centering
    \includegraphics[
        width=\linewidth,
        height=0.78\textheight,
        keepaspectratio
    ]{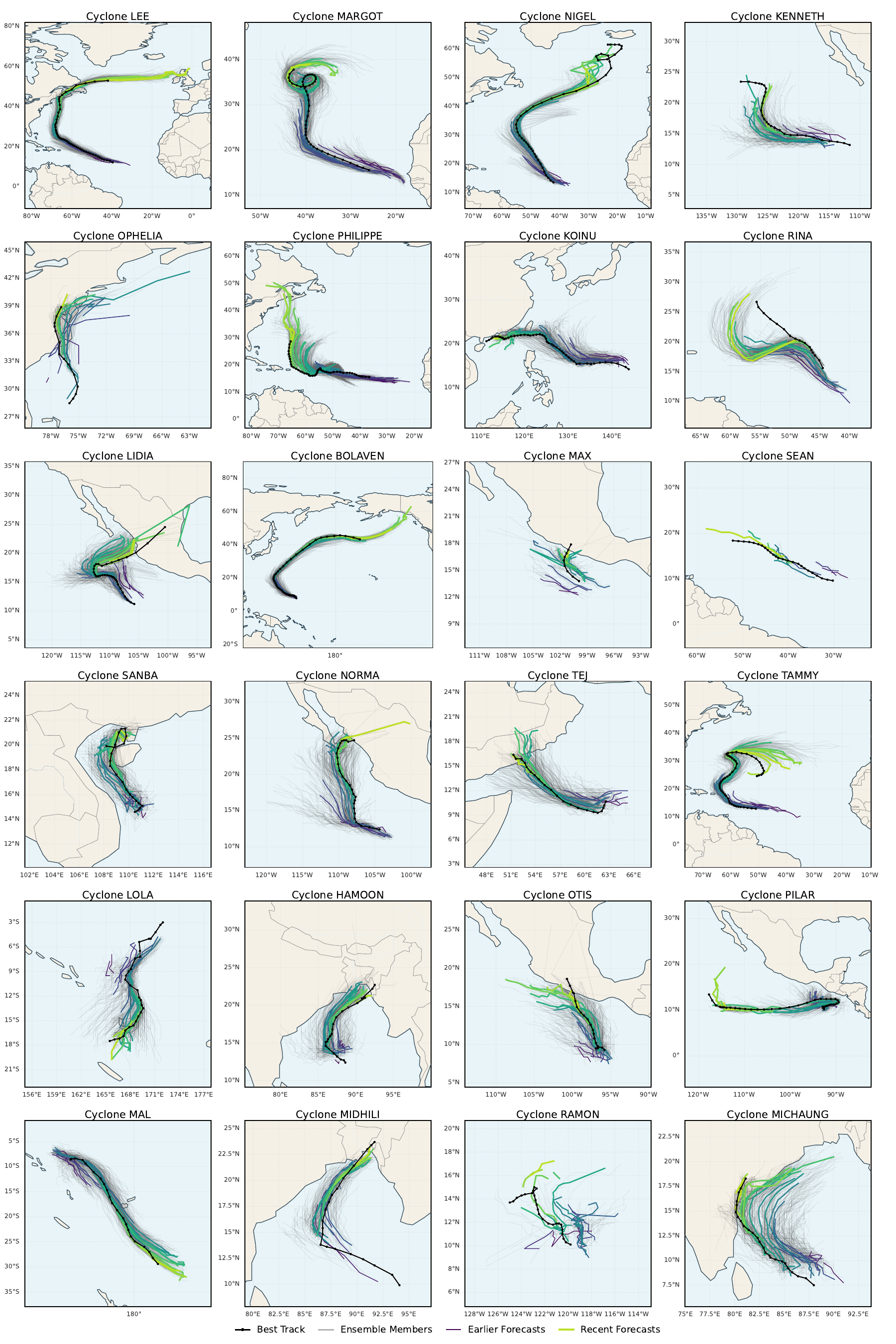}    \caption{Comparison of ensemble track forecasts for multiple tropical cyclones across all initialization times}
    \label{All_Typhoons_4col6row_Part3}
\end{figure}

\begin{figure}[htbp]
    \centering
    \includegraphics[width=1.0\linewidth]{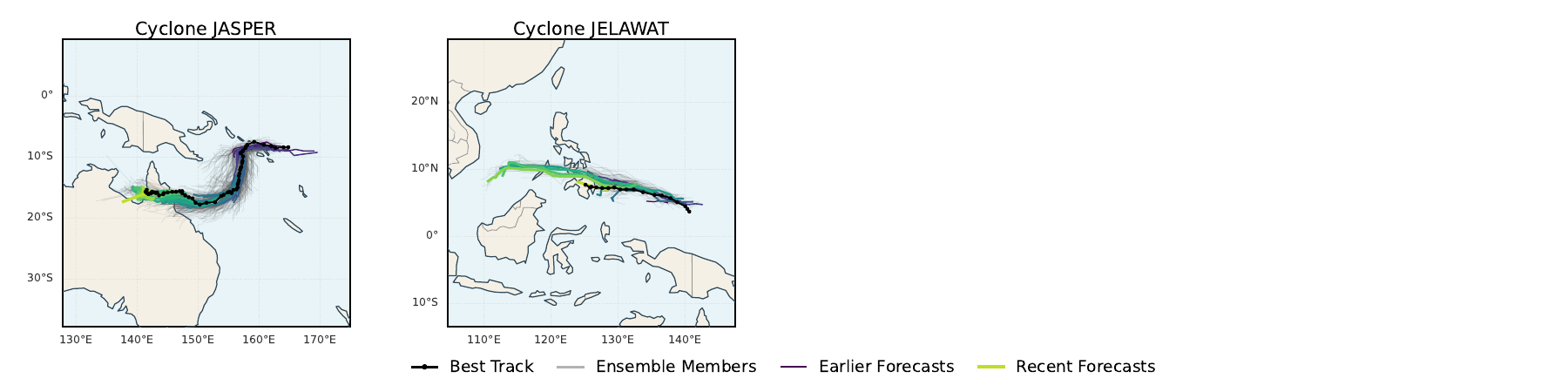}
    \caption{Comparison of ensemble track forecasts for multiple tropical cyclones across all initialization times}
    \label{All_Typhoons_4col6row_Part4}
\end{figure}

To complement the aggregate source--target statistics in the main text, Supplementary Figs.~\ref{All_Typhoons_4col6row_Part1}--\ref{All_Typhoons_4col6row_Part4} visualize atmospheric-state and learned-model ensemble tracks across the 2023 tropical-cyclone cohort and multiple initialization times. These examples provide a qualitative view of how the two uncertainty pathways generate different trajectory distributions across storms; the cohort-level conclusions themselves are based on the quantitative comparisons reported in the main text and Supplementary Table~\ref{tab:source_target_summary}.

\end{document}